\documentclass[10pt]{article} %
\usepackage[preprint]{tmlr}
\usepackage{graphicx}
\usepackage[utf8]{inputenc} %
\usepackage[T1]{fontenc}    %

\usepackage{url}            %
\usepackage{booktabs}       %
\usepackage{amsfonts}       %
\usepackage{nicefrac}       %
\usepackage{wrapfig}
\usepackage{microtype}      %
\usepackage{xcolor}         %
\definecolor{darkblue}{RGB}{0,0,128}
\definecolor{darkgreen}{RGB}{0, 200, 0}
\definecolor{forestlikegreen}{RGB}{0, 160, 0}
\definecolor{darkred}{RGB}{128, 0, 0}
\definecolor{black}{RGB}{0, 0, 0}
\definecolor{errorcolor}{HTML}{8b0000}
\definecolor{viridisgreen}{HTML}{55C667}
\definecolor{observedcolor}{HTML}{eb5760}
\definecolor{networkcolor}{HTML}{2A0593}

\usepackage{amsmath}
\usepackage[title]{appendix}
\usepackage{accents}
\newcommand{\observed}[1]{\protect\accentset{\text{o}}{#1}}%

\usepackage{array}

\newcolumntype{L}{>{\scriptstyle}l}
\newcolumntype{C}{>{\scriptstyle}c}
\newcolumntype{R}{>{\scriptstyle}r}
\newenvironment{mysubarray}{%
  \scriptstyle
  \setlength\arraycolsep{0pt}%
  \setlength\extrarowheight{-1ex}
  \renewcommand\arraystretch{0}
  \begin{array}{RCL}}{\end{array}}

\usepackage{tikz}
\usepackage{tkz-euclide}
\usetikzlibrary{positioning, shapes, arrows, fit, shapes.multipart}
\usepackage{adjustbox}
\usepackage{algorithm}
\usepackage{algpseudocode}

\usepackage{chngcntr}

\usepackage{textcomp}%
\usepackage{manyfoot}%
\usepackage{lmodern}

\usepackage{comment}
\usepackage{enumitem}

\usepackage{subcaption}

\usepackage{placeins}

\usepackage{chngcntr}

\usepackage{booktabs} 
\usepackage{multirow}
\usepackage{makecell}

\newcommand{\numberGaussianMeans}{1}
\newcommand{\numberCS}{2}
\newcommand{\numberDDM}{3}
\newcommand{\numberCovid}{4}
\newcommand{\numberGIN}{5}
\newcommand{\numberGaussianMeansCov}{6}

\newcommand{\x}{\boldsymbol{x}}
\newcommand{\z}{\boldsymbol{z}}

\newcommand{\mub}{\boldsymbol{\mu}}
\newcommand{\xib}{\boldsymbol{\xi}}

\newcommand{\Sigmab}{\boldsymbol{\Sigma}}
\newcommand{\thetab}{\boldsymbol{\theta}}
\newcommand{\phib}{\boldsymbol{\phi}}
\newcommand{\psib}{\boldsymbol{\psi}}
\newcommand{\Psib}{\boldsymbol{\Psi}}

\newcommand{\0}{\boldsymbol{0}}

\newcommand{\given}{\,|\,}

\newcommand{\priorm}{p(\thetab \given \mathcal{M})}
\newcommand{\likm}{p(\x \given \thetab, \mathcal{M})}

\newcommand{\postm}{p(\thetab \given \x, \mathcal{M})}
\newcommand{\NIW}{\text{N-}\mathcal{W}^{-1}}

\newcommand{\prior}{p(\thetab)}

\newcommand{\jointm}{p(\thetab, \x \given \mathcal{M})}

\newcommand{\noised}{p(\xib \given \thetab)}
\newcommand{\model}{g(\thetab, \xib)}
\newcommand{\diff}{\mathrm{d}}

\newcommand{\M}{\mathcal{M}}

\newcommand{\colsquare}[1]{\fcolorbox{#1}{#1}{\rule{0pt}{3pt}\rule{3pt}{0pt}}\,}
\DeclareMathOperator*{\argmin}{argmin}

\usepackage{tcolorbox}

\usepackage{hyperref}
\hypersetup{
	colorlinks=true,
	linkcolor=darkblue,
	filecolor=darkblue,
	urlcolor=darkblue,
	citecolor=darkblue,
	pdftitle={Detecting Model Misspecification in Amortized Bayesian Inference with Neural Networks: An Extended Investigation}
}

\title{Detecting Model Misspecification in Amortized Bayesian Inference with Neural Networks: An Extended Investigation}

\author{\noindent
    \name Marvin Schmitt \email mail.marvinschmitt@gmail.com \\
    \addr Cluster of Excellence SimTech, University of Stuttgart, Germany
      \AND
      \name Paul-Christian Bürkner \email paul.buerkner@gmail.com \\
      \addr Department of Statistics, TU Dortmund University, Germany
      \AND
      \name Ullrich Köthe \email ullrich.koethe@iwr.uni-heidelberg.de\\
      \addr Visual Learning Lab, Heidelberg University, Germany
      \AND
      \name Stefan T. Radev \email radevs@rpi.edu \\
      \addr  Cognitive Science, Rensselaer Polytechnic Institute, United States}

\begin{document}

\maketitle
\vspace*{-5mm}
\begin{tcolorbox}[colback=white, colframe=darkblue]
This article is an extended version of the conference paper \textit{Detecting Model Misspecification in Amortized Bayesian Inference with Neural Networks} \citep[][\href{https://link.springer.com/chapter/10.1007/978-3-031-54605-1_35}{Link}]{schmitt2023mmsgcpr}.
\end{tcolorbox}
\vspace*{5mm}

\begin{abstract}
Recent advances in probabilistic deep learning enable efficient amortized Bayesian inference in settings where the likelihood function is only implicitly defined by a simulation program (simulation-based inference; SBI). But how faithful is such inference if the simulation represents reality somewhat inaccurately, that is, if the true system behavior at test time deviates from the one seen during training? We conceptualize the types of such model misspecification arising in SBI and systematically investigate how the performance of neural posterior approximators gradually deteriorates as a consequence, making inference results less and less trustworthy. To notify users about this problem, we propose a new misspecification measure that can be trained in an unsupervised fashion (i.e., without training data from the true distribution) and reliably detects model misspecification at test time. Our experiments clearly demonstrate the utility of our new measure both on toy examples with an analytical ground-truth and on representative scientific tasks in cell biology, cognitive decision making, disease outbreak dynamics, and computer vision. We show how the proposed misspecification test warns users about suspicious outputs, raises an alarm when predictions are not trustworthy, and guides model designers in their search for better simulators.
\end{abstract}

\section{Introduction}
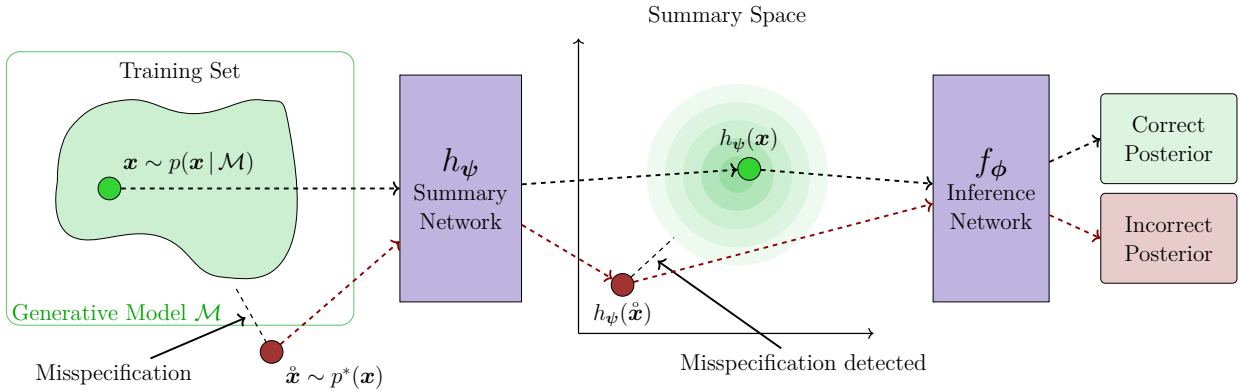
\begin{figure*}[t]
    \centering
    \begin{adjustbox}{width=1.0\textwidth}
    \begin{tikzpicture}[
every text node part/.style={align=center, font={\Large}},
dot/.style={draw, circle, minimum width=0.5cm},
network-box/.style={draw, rectangle, fill=networkcolor!30, minimum height=5cm, minimum width = 2.5cm, inner sep=0.3cm},
posterior-box/.style={draw, rectangle, rounded corners = .10cm, minimum width = 3cm, inner sep=0.5cm},
arrow/.style = {->, very thick}
]

    \node[draw=none, label={Training Set}] (typical-generative-set) {
    \begin{tikzpicture}[scale=0.6]
    \filldraw[draw=black,fill=viridisgreen!30]  plot[smooth, tension=.8, fill=viridisgreen!30] coordinates {(-3.5,0.5) (-3,2.5) (-1,3.5) (1.5,3) (4,3.5) (5,2.5) (5,-2) (2.5,-3) (0.2,-1.5) (-3,-1.5) (-3.5,0.5)};
    \end{tikzpicture}
    };
    \node[dot, fill=darkgreen!80, left=1.3, label=above right:{$\x\sim p(\x\given\mathcal{M})$}] (x-star) at (typical-generative-set) {};
    
    \node[draw, color=viridisgreen,
    fit={(typical-generative-set)}, 
    label={[anchor=south west]south west:\color{forestlikegreen}Generative Model $\mathcal{M}$}, 
    rounded corners=.30cm, inner sep=0.8cm
    ] (generative-model) {};

    \node[dot, fill=errorcolor!80, below right = 0.4cm and -2cm of generative-model, label=below right:{$\observed{\x}\sim p^*(\x)$}] (x-obs) {};

    \node[network-box, right = of generative-model] (summary-network) {$\huge{h_{\psib}}$\\Summary\\Network};
    
    \node[draw=none, right = of summary-network, label={Summary Space}] (kde) {
        \begin{tikzpicture}
            \draw[thick,->] (-3.5, -3.5) -- (-3.5, 3);
            \draw[thick,->] (-3.5, -3.5) -- (3, -3.5);
            \foreach \x/\alpha in {2/10, 1.6/20, 1.2/30, 0.8/45, 0.4/60}
                \fill[viridisgreen!\alpha] (0, 0) circle (\x);
        \end{tikzpicture}
    };
    
    \node[dot, fill=darkgreen!80,
    above right = 0.25cm and 0.3cm of kde,
    label=above:{$h_{\psib}(\x)$}
    ] (s-star) at (kde) {};
    \node[dot, 
    fill=errorcolor!80, 
    below left = -1.5cm and -1.4cm of kde,
    label=below:{$h_{\psib}(\observed{\x})$}
    ] (s-obs) {};
    
    \node[network-box, right = of kde] (inference-network) {$\huge{f_{\phib}}$\\Inference\\Network};

    \matrix[right = of inference-network, row sep = 0.2cm] {
    \node[posterior-box, fill=viridisgreen!20]  (correct-posterior) {Correct\\Posterior}; \\ 
    \node[posterior-box, fill=errorcolor!20] (incorrect-posterior) {Incorrect\\Posterior}; \\
    };

    \draw [dashed, thick] (typical-generative-set) -- (x-obs) node [sloped,midway](M){};;
    
    \node[below left = of M] (simulation-gap) {Misspecification};
    \draw [arrow] (simulation-gap)  -- (M);
    
    \node[below left=-2.6cm and -2.6cm of kde] (kde-outer) {};
    \draw [dashed, thick] (s-obs) -- (kde-outer) node [sloped,midway](M2){};;
    
    \node[below right = 2cm and 0.3cm of M2] (simulation-gap-detected) {Misspecification detected};
    \draw [arrow] (simulation-gap-detected)  -- (M2);

    \draw [arrow, dashed] (x-star) -- (summary-network);
    \draw [arrow, dashed, color=errorcolor] (x-obs) -- (summary-network);
    
    \draw [arrow, dashed] (summary-network) -- (s-star);
    \draw [arrow, color=errorcolor, dashed] (summary-network) -- (s-obs);

    \draw [arrow, dashed] (s-star) -- (inference-network);
    \draw [arrow, dashed, color=errorcolor] (s-obs) -- (inference-network);
    
    \draw [arrow, dashed] (inference-network) -- (correct-posterior.west);
    \draw [arrow, color=errorcolor, dashed] (inference-network) -- (incorrect-posterior.west);
\end{tikzpicture}
    \end{adjustbox}
    \caption{Conceptual overview of our neural approach.
    The summary network $h_\psi$ maps observations $\x$ to summary statistics $h_\psi(\x)$, and the inference network $f_\phi$ estimates the posterior $p(\thetab\given\x,\mathcal{M})$ from the summary statistics.
    The generative model $\mathcal{M}$ creates training data $\x$ in the green region, and the networks learn to map these data to well-defined summary statistics and posteriors (green regions/dot/box).
    If the generative model $\mathcal{M}$ is misspecificed, real observations $\observed{\x}$ fall outside the training region and are therefore mapped to outlying summary statistics and potentially incorrect posteriors (red dots/box).
    Since our learning approach enforces a known inlier summary distribution (e.g., Gaussian), misspecification can be detected by a distribution mismatch in summary space, as signaled by a high maximum mean discrepancy score \citep{Gretton2012}.}
    \label{fig:conceptual}
\end{figure*}

Computer simulations play a fundamental role in many fields of science and engineering. 
However, the associated {\em inverse} problems of finding simulation parameters that accurately reproduce or predict real-world behavior are generally difficult and often analytically intractable.
Here, we consider \emph{simulation-based inference} \citep[SBI;][]{frontier} as a general approach to address this problem within a Bayesian inference framework.
That is, given an assumed generative model $\mathcal{M}$ (as represented by the simulation program, see Section~\ref{sec:defining-model-misspecification} for details) and observations $\x$ (real or simulated outcomes), we estimate the posterior distribution $p(\thetab\given\x,\mathcal{M})$ of the simulation parameters $\thetab$ that would reproduce the observed $\x$.
The recent introduction of efficient neural network approximators for this task has inspired a rapidly growing literature on SBI solutions for various application domains \citep[e.g.,][]{butter2022machine,lueckmann2021benchmarking,gonccalves2020training,bayesflow_agent,bayesflow_qcd,von_krause_mental_2022,ghaderi-kangavari_general_2023}.
These empirical successes call for a systematic investigation of the trustworthiness of SBI, see \autoref{fig:conceptual}.

\newpage
\lhead{Detecting Model Misspecification in Amortized Bayesian Inference}
\rhead{Schmitt et al.}
We conduct an extensive analysis of neural posterior estimation (NPE) and sequential neural posterior estimation (SNPE), two deep learning approaches to approximate the posterior distribution $p(\thetab\given\x,\mathcal{M})$.
In particular, we study their accuracy under model misspecification, where the generative model $\mathcal{M}^*$ at test time (the ``true data generating process'') deviates from the one assumed during training (i.e., $\mathcal{M}^*\ne\mathcal{M}$), a situation commonly known as {\em simulation gap}.
As a consequence of a simulation gap, the observed data of interest might lie outside of the simulated data from the training phase of SBI.
Paralleling the notion of ``out-of-distribution'' in anomaly detection and representation learning \citep{yang2021generalized}, simulation gaps may lead to ``out-of-simulation'' samples, and ultimately to silent errors in posterior estimates.

\begin{figure}[t]
\centering
    \begin{minipage}{.40\linewidth}
        \includegraphics[width=\linewidth]{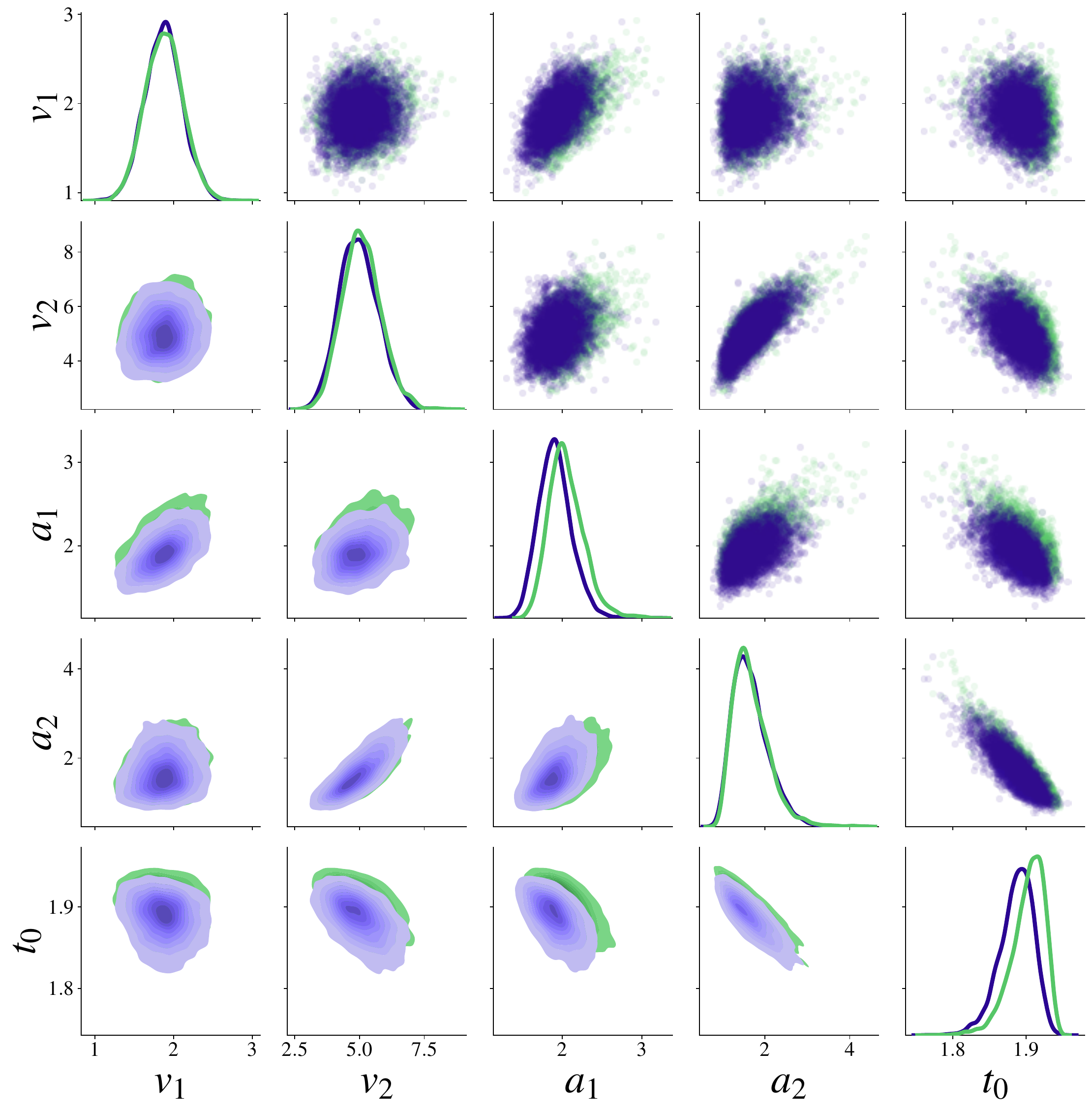}
    \end{minipage}
    \hspace*{2cm}
    \begin{minipage}{.40\linewidth}
        \includegraphics[width=\linewidth]{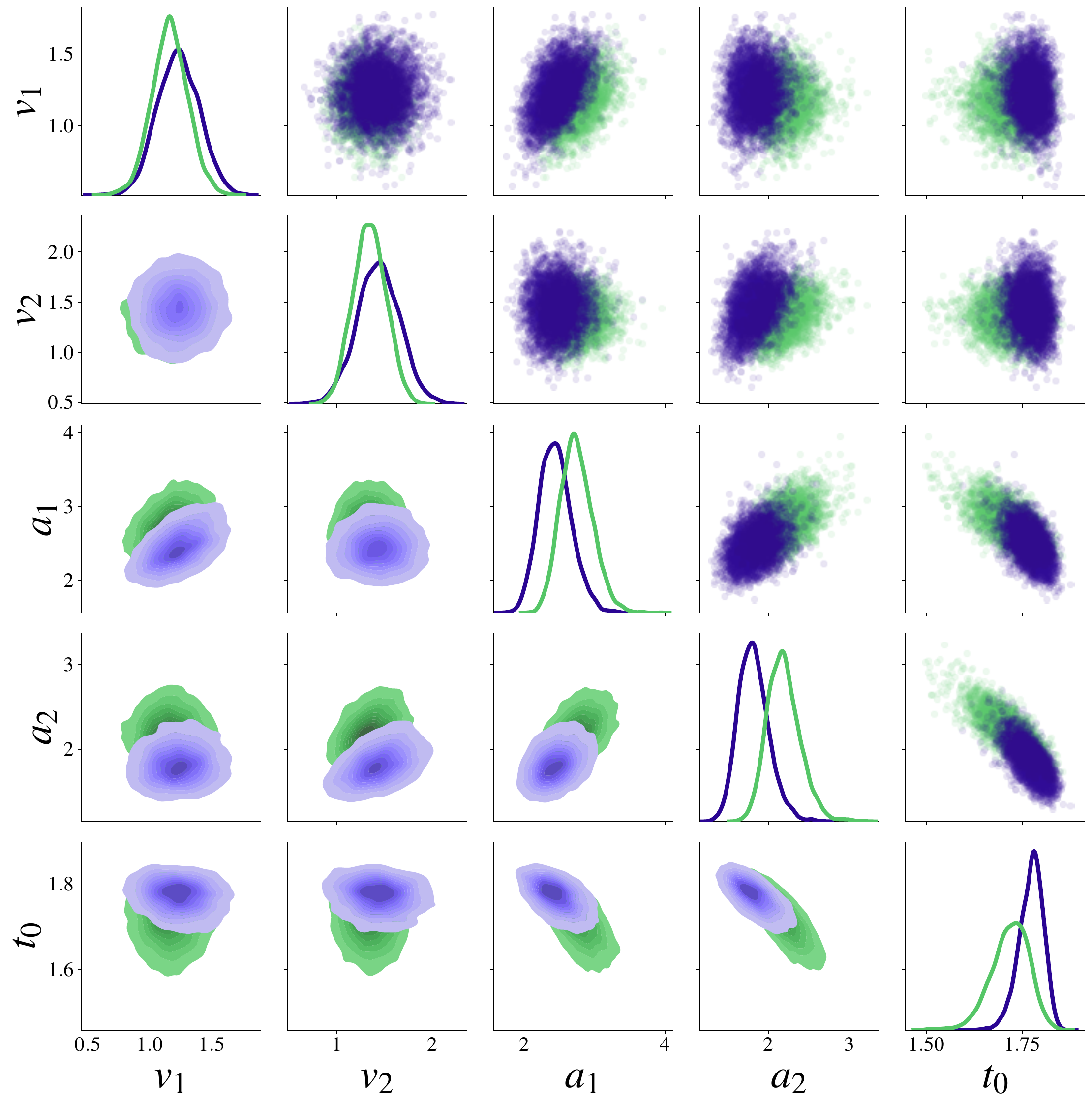}   
    \end{minipage}\\
    \begin{minipage}{.40\linewidth}
        \includegraphics[width=\linewidth]{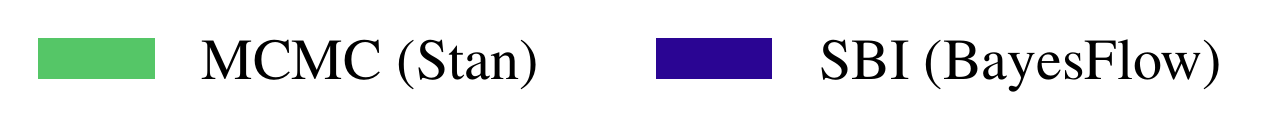}
    \end{minipage}\\
    \begin{minipage}{.45\linewidth}
        \begin{subfigure}[t]{\linewidth}
            \caption{Well-specified: Similar posteriors.}
            \label{fig:exp:ddm:stan-bf:clean}
        \end{subfigure}
    \end{minipage}
    \hfill
    \begin{minipage}{.45\linewidth}
        \begin{subfigure}[t]{\linewidth}
            \caption{Misspecified: Dissimilar posteriors.}
            \label{fig:exp:ddm:stan-bf:slow}
        \end{subfigure}
    \end{minipage}
    \caption{Preview of \textbf{Experiment \numberDDM} on reaction time modeling in psychological experiments. Posteriors obtained via non-amortized MCMC (HMC in Stan) and amortized simulation-based inference (NPE in BayesFlow) are very similar when the model is well-specified (left).
    However, a simulation gap (here: not accounting for occasional slow responses due to mind wandering) leads to considerable disagreement between these methods (right).
    }
    \label{fig:exp:ddm:stan-bf}
\end{figure}

Our experiments illustrate the difference between the two situations (see \autoref{fig:exp:ddm:stan-bf} for a quick preview):
When the simulation model is well-specified, the posterior estimates of a neural posterior estimator and a gold-standard MCMC sampler \citep[HMC-MCMC;][]{Stan2022} are essentially equal, whereas both approaches disagree considerably under model misspecification.
In order to avoid drawing incorrect conclusions from misspecified models and incorrect posteriors, it is of crucial importance to detect whether $\mathcal{M}^*\ne\mathcal{M}$ and to quantify the severity of the mismatch.
However, this is difficult in practice because the true data generating process $\mathcal{M}^*$ is generally unknown (except in controlled validation settings).

In this work, we propose a new misspecification measure that can be trained in an unsupervised fashion (i.e., without knowledge of $\mathcal{M}^*$ or training data from the true data distribution) and reliably quantifies by how much $\mathcal{M}^*$ deviates from $\mathcal{M}$ at test time.
Our experiments clearly demonstrate the power of our new measure both on toy examples with an analytical ground-truth, and on representative scientific tasks in cell biology, cognitive decision making, and disease outbreak dynamics.
We show how simulation-based posterior inference gradually deteriorates as the simulation gap widens and how the proposed misspecification test warns users about suspicious outputs, raises an alarm when predictions are not trustworthy, and guides model designers in their search for better simulators.

Our investigations complement existing work on deep amortized SBI, whose main focus has been on network architectures and training algorithms for high accuracy in the well-specified case $\mathcal{M}^*=\mathcal{M}$ \citep{ramesh2022gatsbi,pacchiardi2022likelihood,contrastive,bayesflow,apt,bayes_lstm,papamakarios2016fast}, with a steadily growing interest in inference algorithms that are robust to certain types of model misspecification \citep{huang2023,ward_robust_2022, dellaporta_robust_2022,leclercq_simulation-based_2022,kelly2023misspecification, gloeckler2023adversarial}.
In particular, our paper makes the following key contributions:
\begin{enumerate}[label=(\roman*), parsep=6pt, topsep=6pt]
    \item We systematically conceptualize different sources of model misspecification in amortized Bayesian inference with neural networks and propose a new detection criterion that is widely applicable to different model structures, inputs, and outputs.
    \item We incorporate this criterion into existing neural posterior estimation methods, with hand-crafted and learned summary statistics, with sequential or amortized inference regimes, and we extend the associated learning algorithms in a largely non-intrusive manner.
    \item We conduct a systematic empirical evaluation of our detection criterion, the influence of the summary space dimension, and the relationship between summary outliers and posterior distortion under various types and strengths of model misspecification.
\end{enumerate}

\section{Related Work}\label{sec:related-work}
Model misspecification has been studied both in the context of standard Bayesian inference and generalizations thereof \citep{knoblauch2019generalized,schmon_generalized_2021}.
To alleviate model misspecification in generalized Bayesian inference, researchers have investigated probabilistic classifiers \citep{mms_genbayes}, second-order PAC-Bayes bounds \citep{masegosa2020learning}, scoring rules \citep{giummole2019objective}, priors over a class of predictive models \citep{loaiza2021focused}, or Stein discrepancy as a loss function \citep{matsubara_robust_2022}.
Notably, these approaches deviate from the standard Bayesian formulation and investigate alternative schemes for belief updating and learning (e.g., replacing the likelihood function with a generic loss function).
In contrast, our method remains grounded in the standard Bayesian framework embodying an implicit likelihood principle \citep{berger_likelihood_1988,frontier}.
Differently, power scaling methods incorporate a modified likelihood (raised to a power $0 < \alpha < 1)$ in order to prevent potentially overconfident Bayesian updating \citep{bayesian_miss,holmes2017assigning}.
However, the SBI setting assumes that the likelihood function is not available in closed-form, which makes an explicit modification of the implicitly defined likelihood less obvious.

Neural approaches to amortized SBI can be categorized as either targeting the posterior \citep{bayesflow,apt}, the likelihood \citep{snle,ratios}, or both \citep{snpla,radev2023jana}.
These methods employ simulations for training neural approximators which can either generate samples from the posterior directly \citep{bayesflow,apt,snpla} or in tandem with Markov chain Monte Carlo (MCMC) sampling algorithms \citep{snle,ratios}.
Since the behavior of these methods depends on the fidelity of the simulations used as training data, we hypothesize that their estimation quality will be, in general, unpredictable, when faced with atypical real-world data.
Indeed, the critical impact of model misspecification in neural SBI has been commonly acknowledged in the scientific research community \citep{cannon_investigating_2022, alquier_concentration_2019,zhang_convergence_2020,frazier_model_2020,frazier_robust_2021,pacchiardi_score_2022, ward_robust_2022, kelly2023misspecification, gloeckler2023adversarial, huang2024learning}.

Recent approaches to detect model misspecification in simulation-based inference are usually based on the obtained approximate posterior distribution \citep{dellaporta_robust_2022,leclercq_simulation-based_2022}.
However, we explicitly show in \textbf{Experiment \numberGaussianMeans} and \textbf{Experiment \numberDDM} that the approximate posteriors in simulation-based inference tend to show pathological behavior under misspecified models. 
Posteriors from misspecified models may erroneously look legitimate, rendering diagnostic methods on their basis unreliable.
Moreover, the same applies for approaches based on the \textit{posterior predictive distribution} \citep{burkner_approximate_2020,gabry_visualization_2019,vehtari_survey_2012} since these also rely on the fidelity of the posterior distribution and can therefore only serve as an indirect measure of misspecification.

A few novel techniques aim to \emph{mitigate} model misspecification in simulation-based inference to achieve robust inference.
\citet{ward_robust_2022} explore a way to alleviate model misspecification with two neural approximators and subsequent MCMC.
While both approaches are appealing in theory, the computational burden of MCMC sampling contradicts the idea of amortized inference and prohibits their use in complex applications with learned summary statistics and large amounts of data.
In fact, \citet{von_krause_mental_2022} used amortized neural SBI on more than a million data sets of multiple observations each and demonstrated that an alternative inference method involving non-amortized MCMC would have taken years of sampling.
\citet{huang2024learning} propose a method to learn summary statistics that are robust to unmodeled data contamination for sequential (i.e., non-amortized) neural posterior estimation.
Finally, \citet{gloeckler2023adversarial} propose to penalize the Fisher information of neural posterior estimators to increase their robustness to adversarial attacks (which can be considered a form of subtle model misspecification).  

For robust non-amortized ABC samplers, the possibility of utilizing hand-crafted summary statistics as an important element of misspecification analysis has already been explored \citep{frazier_model_2020,frazier_robust_2021}.
Our work parallels these ideas and extends them to the case of \textit{learnable summary statistics} in amortized SBI on potentially massive data sets, where ABC becomes infeasible.
However, we show in \textbf{Experiment \numberCS} that our method also works with hand-crafted summary statistics.

Finally, from the perspective of deep anomaly detection, our approach for learning informative summary statistics can be viewed as a special case of \emph{generic normality feature learning} \citep{pang_deep_2022}. 
Standard learned summary statistics are optimized with a generic feature learning objective which is not primarily designed for anomaly detection \citep{bayesflow, chen2020neural}.
However, since learned summary statistics are also optimized to be maximally informative for posterior inference, they will likely capture underlying data regularities \citep{pang_deep_2022} and serve as a natural target for detecting model misspecification.

\newpage
\section{Method}\label{sec:methods}

For simulation-based Bayesian inference, we define a generative model as a triple $\mathcal{M}=\big(\model,\noised,\prior\big)$.
A generative model $\M$ generates data $\x \in \mathcal{X}$ according to the system
\begin{equation}
    \x = \model \quad\textrm{with}\quad \xib \sim \noised,\; \thetab \sim \prior,
\end{equation}
where $g$ denotes a (randomized) simulator, $\xib \in \Xi$ is a source of randomness (i.e., noise) with density function $\noised$, and $\prior$ encodes prior knowledge about plausible simulation parameters $\thetab \in \Theta$.
Throughout the paper, we use the decorated symbol $\observed{\x}$ to mark data that was in fact \emph{observed} in the real world and not merely simulated by the assumed model $\M$.
The parameters $\thetab$ consist of hidden properties whose role in $g$ we explicitly understand and model, and $\xib$ takes care of nuisance effects that we only treat statistically.
The abstract spaces $\mathcal{X}, \Xi$, and $\Theta$ denote the domain of possible output data (possible worlds), the scope of noise, and the set of admissible model parameters, respectively.
The distinction between hidden properties $\thetab$ and noise $\xib$ is not entirely clear-cut, but depends on our modeling goals and may vary across applications.

Our generative model formulation is equivalent to the standard factorization of the Bayesian joint distribution into likelihood and prior, $\jointm = \likm\,\priorm$, where $\mathcal{M}$ expresses the prior knowledge and assumptions embodied in the model.
The likelihood is obtained by marginalizing the joint distribution $p(\xib, \x \given \thetab, \mathcal{M})$ over all possible values of the nuisance parameters $\xib$, that is, over all possible execution paths of the simulation program, for fixed $\thetab$:
\begin{equation}
    \likm = \int_{\Xi} p(\xib, \x \given \thetab, \mathcal{M})\,\diff \xib.
\end{equation}
This integral is typically intractable \citep{frontier}, but we assume that it exists and is non-degenerate, that is, it defines a proper density over the constrained manifold $\left(g(\thetab, \xib), \xib\right)$, and this density can be learned. 
A major challenge in Bayesian inference is approximating the posterior distribution $p(\thetab\given\x,\M)\propto p(\x\given\thetab,\M)\,p(\thetab\given\M)$.
Below, we focus on \emph{amortized} posterior approximation with neural networks, which aims to achieve zero-shot posterior sampling for any input data $\x$ compatible with the reference model $\mathcal{M}$.\footnote{We demonstrate in \textbf{Experiment \numberGaussianMeans} that model misspecification also affects the performance of non-amortized sequential neural posterior estimation, and our method can also detect such misspecification when deployed with SNPE.}

\subsection{Neural Posterior Estimation}

Neural Posterior Estimation (NPE) with learned summary statistics $h_{\psib}(\x)$ involves a posterior network and a summary (aka embedding) network.
The summary network transforms input data $\x$ of variable size and structure (e.g., multiple IID measurements) to a fixed-length representation $\z = h_{\psib}(\x)$.
The inference network $f_{\phib}$ generates random draws from an approximate posterior $q_{\phib}$ via a conditional generative neural network.
In the following, we focus on normalizing flows realized as conditional invertible neural networks \citep{Ardizzone2019}, but most of our discussion applies to any generative backbone architecture that relies on (explicit or implicit) data compression for flexible inference.
Ideally, the two networks jointly minimize the expected forward Kullback-Leibler (KL) divergence between analytic and approximate posterior
\begin{align}
  (\psib^*, \phib^*) &=
  \argmin_{\psib, \phib}  
    \mathbb{E}_{p^*(\x)}\Big[\mathbb{KL}\left[\postm \,||\, q_{\phib}\big(\thetab\given h_{\psib}(\x), \mathcal{M}\big) \right] \Big] \label{eq:kl-npe} \\
    &= \argmin_{\psib, \phib}\mathbb{E}_{p^*(\x)}\Big[ \mathbb{E}_{p(\thetab \given \x, \mathcal{M})}\left[-\log q_{\phib}\big(\thetab\given h_{\psib}(\x), \mathcal{M}\big) \right] \Big] \label{eq:kl_bf_wannabe},
\end{align}
where the expectation runs over the unknown data-generating distribution $p^*(\x)$ and Eq.~\ref{eq:kl_bf_wannabe} follows from Eq.~\ref{eq:kl-npe} since the analytic posterior $\postm$ and its entropy do not depend on the trainable neural network parameters ($\psib$, $\phib$).
In theory, minimizing the above criterion ensures that the networks learn to perform correct \textit{amortized} posterior estimation over any observable configuration $\x \sim p^*(\x)$ \citep{apt,bayesflow}.
In practice, however, Eq.~\ref{eq:kl_bf_wannabe} is hardly a feasible target, since we may not have enough real data to approximate the outer expectation and the inner expectation runs over the analytic posterior which we assume is intractable in the first place. 

To render Eq.~\ref{eq:kl_bf_wannabe} tractable for NPE, we replace the unknown $p^*(\x)$ in the outer expectation with the \textit{model-implied distribution} (aka \textit{evidence}, \textit{marginal likelihood}, or \textit{prior predictive distribution})
\begin{equation}
    p(\x \given \M) = \int p(\x \given \thetab, \mathcal{M})\,p(\thetab \given \M)\,\diff\x.
\end{equation}
The marginal likelihood is a key quantity in Bayesian inference \citep{mackay2003information, lotfi2022bayesian} and, informally speaking, determines the \textit{generative scope} of model $\mathcal{M}$ as an average over the \textit{joint model} $p(\thetab, \x \given \M) = p(\x\given\thetab,\M)\,p(\thetab\given\M)$.
Thus, we can now rephrase Eq.~\ref{eq:kl_bf_wannabe} solely in terms of model-implied quantities:
\begin{align}
  (\psib^*, \phib^*) &= 
  \argmin_{\psib, \phib}\mathbb{E}_{p(\x \given \M)}\Big[ \mathbb{E}_{p(\thetab \given \x, \mathcal{M})}\left[-\log q_{\phib}\big(\thetab\given h_{\psib}(\x), \mathcal{M}\big) \right] \Big]\\
  &= \argmin_{\psib, \phib}  
    \mathbb{E}_{\jointm}\Big[-\log q_{\phib}\big(\thetab\given h_{\psib}(\x), \mathcal{M}\big)\Big]. \label{eq:kl_bf}
\end{align}
This ``trick'' allows us to approximate the expectation in Eq.~\ref{eq:kl_bf} via an ``online'' stream or an ``offline'' data set of simulations from the generative model $\M$ and repeat the process until convergence to an amortized posterior approximator that is ``competent'' over $p(\x \given \M)$.
However, in doing so, we have tacitly assumed a \textit{closed-world} setting in which sampling from the prior predictive distribution $p(\x \given \M)$ is equivalent to sampling from the \textit{true} data distribution $p^*(\x)$.
This critical assumption makes NPE susceptible to posterior errors in the \textit{open world} due to model misspecification or ``out-of-distribtuion'' (OOD) data \citep{yang2021generalized}, as we discuss next.

\subsection{Model Misspecification in Simulation-Based Inference}\label{sec:defining-model-misspecification}

When modeling a complex system or process, we typically assume an unknown (true) generator $\M^*$, which yields an unknown (true) distribution $\observed{\x} \sim p^*(\x)$ and is available to the data analyst only via a finite realization (i.e., actually observed data $\observed{\x}$).
According to a common definition \citep{frazier_model_2020,white1982,masegosa2020learning,lotfi2022bayesian}, the generative model $\M$ is \textit{well-specified} if a ``true'' parameter $\thetab^*\in\Theta$ exists, such that the likelihood function matches the data-generating distribution,
\begin{equation}\label{eq:mms-conventional}
    \exists\thetab^*: p(\x \given \thetab^*, \mathcal{M}) = p^*(\x),
\end{equation}
and \textit{misspecified} otherwise.
This likelihood-centered definition is well-established and sensible in many domains of traditional Bayesian inference.

In \emph{simulation-based} inference, however, there is an additional difficulty regarding model specification: 
Simulation-based training (see Eq.~\ref{eq:kl_bf}) takes the expectation with respect to the model-implied distribution~$p(\x\given\M)$, not necessarily the ``true'' distribution $p^*(\x)$ that generates the observed data.
Thus, optimal convergence does not imply correct amortized inference or faithful prediction in the real world when there is a simulation gap, that is, when the assumed training model $\M$ deviates critically from the unknown true generative model $\M^*$.
Crucially, even if the generative model $\M$ is well-specified according to the likelihood-centered definition in Eq.~\ref{eq:mms-conventional}, finite training with respect to a ``wrong'' prior (predictive) distribution will likely result in insufficient learning of relevant parameter (and data) regions.
This scenario could also be framed as ``out-of-simulation'' (OOSim) by analogy with the out-of-distribution (OOD) problem in machine learning applications \citep{yang2021generalized}.
In fact, we observe in \textbf{Experiment \numberGaussianMeans} that a misspecified prior distribution worsens posterior inference just like a misspecified likelihood function.
What is more, the same issues also apply when the observed data lies within the support of the assumed model \textit{but only populates a narrow subspace}.
While this is not strictly an \textit{out}-of-simulation setting, the simulation-based neural network training will possibly be inefficient because the generative forward model rarely yields relevant training examples.
Hence, posterior performance may be impeded given small simulation budgets---a common scenario in real-life applications where data is scarce and training times limited.

Thus, our adjusted definition of model misspecification \emph{in the context of simulation-based inference} considers the entire prior predictive distribution $p(\x\given\M)$ relative to the data-generating process $p^*(\x)$:
A generative model $\mathcal{M}$ is well-specified if the information loss through modeling $p^*(\x)$ with $p(\x\given\M)$ is below an acceptance threshold $\vartheta$,
\begin{equation}
    \mathbb{D}\big[p(\x \given \mathcal{M})\,||\,p^*(\x)\big] < \vartheta,
\end{equation}
and misspecified otherwise.
The symbol $\mathbb{D}$ denotes a divergence metric between the data distributions implied by reality and by the model (i.e., the marginal likelihood). 
A natural choice for $\mathbb{D}$ would stem from the family of $\mathcal{F}$-divergences, such as the KL divergence.
We choose the Maximum Mean Discrepancy (MMD),
\begin{equation}\label{eq:MMD:MMD-kernel-trick}
    \mathbb{MMD}^2\big[p^*(\x)\,||\,p(\x \given \mathcal{M})\big] =
    \mathbb{E}_{
    x, \x'\sim 
    p^*(\x)}\big[\kappa(\x, \x')\big]
    + \mathbb{E}_{
    \x, \x'\sim 
    p(\x \given \mathcal{M})}\big[\kappa(\x, \x')\big]
    - 2 \mathbb{E}_{
    \begin{mysubarray}
      \x{\ }&\sim&p^*(\x) 
      \\ 
      \x'&\sim&p(\x \given \mathcal{M})
    \end{mysubarray}
    }\big[\kappa(\x, \x')\big],
\end{equation}
where $\kappa(\cdot,\cdot)$ is a positive definite kernel.
We can tractably estimate the MMD based on finite samples from $p(\x \given \mathcal{M})$ and $p^*(\x)$ and its analytic value equals zero if and only if the two densities are equal \citep{Gretton2012}.
In our empirical evaluation, we use sums of Gaussian kernels with different widths $\sigma_i$ as an established and flexible universal kernel \citep{Muandet2017}.
However, \citet{Ardizzone2018} argue that kernels with heavier tails may improve performance by yielding superior gradients for outliers.
Thus, we repeated all experiments with a sum of inverse multiquadratic kernels \citep{Tolstikhin2017}, and find that the results are essentially equal.

Our adjusted definition of model misspecification no longer assumes the existence of a \textit{true} parameter vector~$\thetab^*$ (cf. Eq.~\ref{eq:mms-conventional}).
Instead, we focus on the \textit{marginal likelihood} $p(\x \given \mathcal{M})$ which represents the entire prior predictive distribution of a model and does not commit to a single most representative parameter vector.
In this way, multiple models whose marginal distributions are representative of $p^*(\x)$ can be considered well-specified without any reference to some hypothetical ground-truth $\thetab^*$, which may not even exist for opaque systems with unknown properties.

\subsection{Structured Summary Statistics}\label{sec:structured-summary-statistics}
In simulation-based inference, summary statistics have a dual purpose because (i) they are fixed-length vectors, even if the input data $\x$ have variable length; and (ii) they usually contain crucial features of the data, which simplifies neural posterior inference.
However, in complex real-world scenarios such as COVID-19 modeling (see \textbf{Experiment \numberCovid}), it is not feasible to rely on hand-crafted summary statistics.
Thus, combining neural posterior estimation with \emph{learned summary statistics} leverages the benefits of summary statistics (i.e., compression to fixed-length vectors) while avoiding the virtually impossible task of manually designing summariesfor complex models.

In simulation-based inference, the summary network $h_{\psib}$ acts as an interface between the data $\x$ and the inference network $f_{\phib}$.
Its role is to learn maximally informative summary vectors of fixed size $S$ from complex and structured observations (e.g., sets of $i.\-i.\-d.\-$ measurements or multivariate time series).
Since the learned summary statistics are optimized to be
maximally informative for posterior inference, they are forced to capture underlying data regularities (see Section~\ref{sec:related-work}).
Therefore, we deem the summary network's representation $\z = h_{\psib}(\x)$ as an adequate target to detect simulation gaps.%

Specifically, we prescribe an $S$-dimensional multivariate unit (aka.\ standard) Gaussian distribution to the summary space, $p\big(\z=h_{\psib}(\x) \given \mathcal{M}\big) \approx \mathcal{N}(\z \given\0, \mathbb{I})$, by minimizing the MMD between summary network outputs and random draws from a unit Gaussian distribution. 
To ensure that the summary vectors comply with the support of the Gaussian density, we use a linear (bottleneck) output layer with $S$ units in the summary network.
A random vector in summary space takes the form $h_{\psib}(\x) \equiv \z \equiv  (z_1, \ldots, z_S)\in\mathbb{R}^S$. 
The extended optimization objective follows as
\begin{equation}\label{eq:bf_kl_mmd}
    \psib^*,\phib^* = \argmin_{\psib, \phib}
    \mathbb{E}_{\jointm}\Big[-\log q_{\phib}\big(\thetab\,\big|\, h_{\psib}(\x), \mathcal{M}\big)
    \Big]
    + \gamma\,\mathbb{MMD}^2\Big[p\big(h_{\psib}(\x)\,\big|\, \mathcal{M}\big)\,\Big|\Big|\,\mathcal{N}\big(\0, \mathbb{I}\big)\Big]
\end{equation}
with a hyperparameter $\gamma$ to control the relative weight of the MMD term.
Intuitively, this objective encourages the approximate posterior $q_{\phib}\big(\thetab\given h_{\psib}(\x), \mathcal{M}\big)$ to match the correct posterior and the summary distribution $\smash{p\big(h_{\psib}(\x) \given \mathcal{M}\big)}$ to match a unit Gaussian. %
The extended objective does not constitute a theoretical trade-off between the two terms, since the MMD merely reshapes the summary distribution in an information-preserving manner. 
In practice, the extended objective may render optimization of the summary network more difficult, but our experiments suggest that it does not restrict the quality of the amortized posteriors.

This method is also directly applicable to hand-crafted summary statistics, which already have fixed length and usually contain rich information for posterior inference.
Thus, the task of the summary network $h_{\psi}$ simplifies to transforming the hand-crafted summary statistics to a unit Gaussian (Eq.~\ref{eq:bf_kl_mmd}) to enable model misspecification via distribution matching during test time (see below).
We apply our method to a problem with hand-crafted summary statistics in \textbf{Experiment \numberCS}.

\subsection{Detecting Model Misspecification with Finite Data}
Once the simulation-based training phase is completed, we can generate $M$ validation samples $\smash{\{\thetab^{(m)}, \x^{(m)}\}}$ from our generative model $\M$ and pass them through the summary network to obtain a sample of latent summary vectors $\smash{\{\z^{(m)}\}}$, where $\smash{\z=h_{\psib}(\x)}$ denotes the summary network output.
This sample contains important convergence information: If $\smash{\z}$ is approximately unit Gaussian, we can assume a structured summary space given the training model $\M$.
This enables misspecification diagnostics via distribution checking during inference on real data (algorithm in Appendix).

Let $\smash{\{\observed{\x}^{(n)}\}}$ be an \emph{observed} sample, either simulated from a different generative model, or arising from real-world observations with an unknown generator. 
Before invoking the inference network, we pass this sample through the summary network to obtain the summary statistics for the sample: $\smash{\{\observed{\z}^{(n)}\}}$.
We then compare the validation summary distribution $\{\z^{(m)}\}$ with the summary statistics of the observed data $\smash{\{\observed{\z}^{(n)}\}}$ according to the sample-based MMD estimate $\smash{\widehat{\text{MMD}}^2\left(p(\z)\,||\, p(\observed{\z})\right)}$ \citep{Gretton2012}. %
Importantly, we are not limited to pre-determined sizes of simulated or real-world data sets, as the MMD estimator is defined for arbitrary $M$ and $N$. %
To allow MMD estimation for data sets with single instances ($N=1$ or $M=1$), we do not use the unbiased MMD version from \citet{Gretton2012}. 
Singleton data sets are an important use case for our method in practice, and potential advantages of unbiased estimators do not justify exclusion of such data.
To enhance visibility, the figures in the experimental section will plot the square root of the squared MMD estimate.

Whenever we estimate the MMD from finite data, its estimates vary according to a sampling distribution and we can resort to a frequentist hypothesis test to determine the probability of observed MMD values under well-specified models.
Although this sampling distribution under the null hypothesis is unknown, we can estimate it from multiple sets of simulations from the generative model, $\smash{\{\z^{(m)}\}}$ and $\smash{\{\z^{(n)}\}}$, with $M$ large and $N$ equal to the number of real data sets.
Based on the estimated sampling distribution, we can obtain a critical MMD value for a fixed Type I error probability $(\alpha)$ and compare it to the one estimated with the observed data. 
In general, a larger $\alpha$ level corresponds to a more conservative modeling approach: A larger type I error implies that more tests reject the null hypothesis, which corresponds to more frequent model misspecification alarms and a higher chance that incorrect models will be recognised.
The Type II error probability ($\beta$) of this test will generally be high (i.e., the \emph{power} of the test will be low) whenever the number of real data sets $N$ is very small.
However, we show in \textbf{Experiment \numberCovid} that even as few as $5$ real data sets suffice to achieve $\beta \approx 0$ for a complex model on COVID-19 time series.

\newpage
\section{Experiments}

\subsection{Experiment \numberGaussianMeans: 2D Gaussian Means}\label{sec:experiment-toy-conjugate}

\begin{figure}[t]
    \centering
    \begin{subfigure}[t]{0.40\linewidth}
        \includegraphics[width=\linewidth]{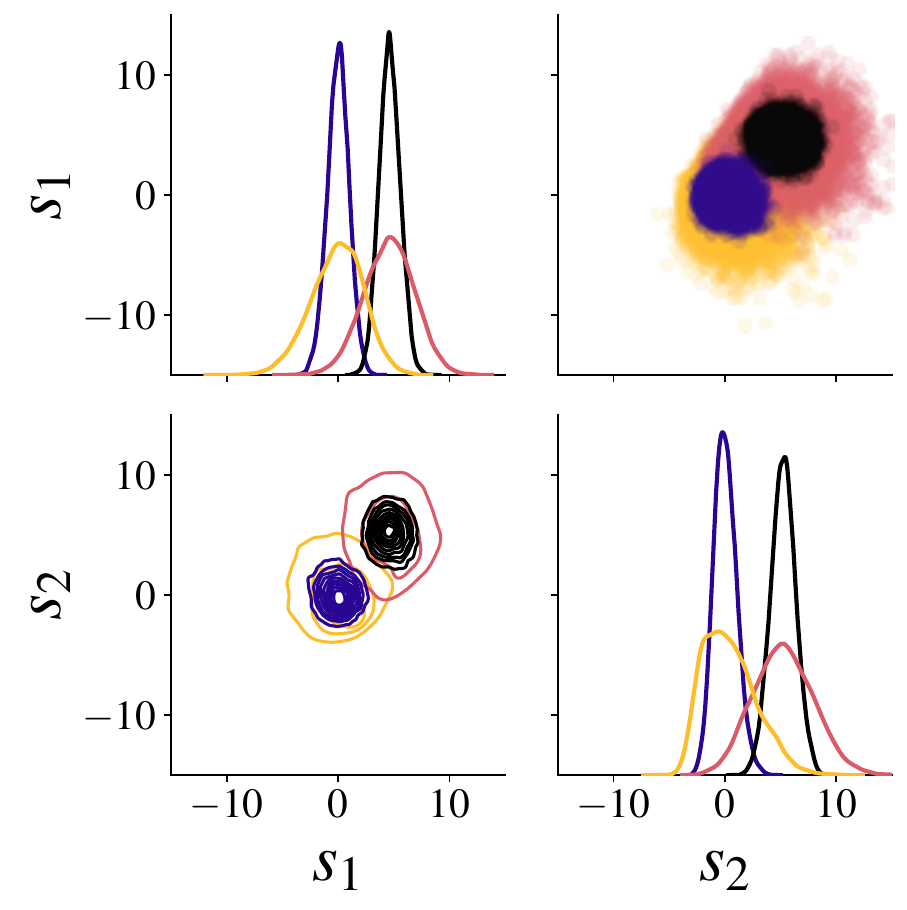}
    \end{subfigure}\hspace*{1cm}
    \begin{subfigure}[t]{0.40\linewidth}
        \includegraphics[width=\linewidth]{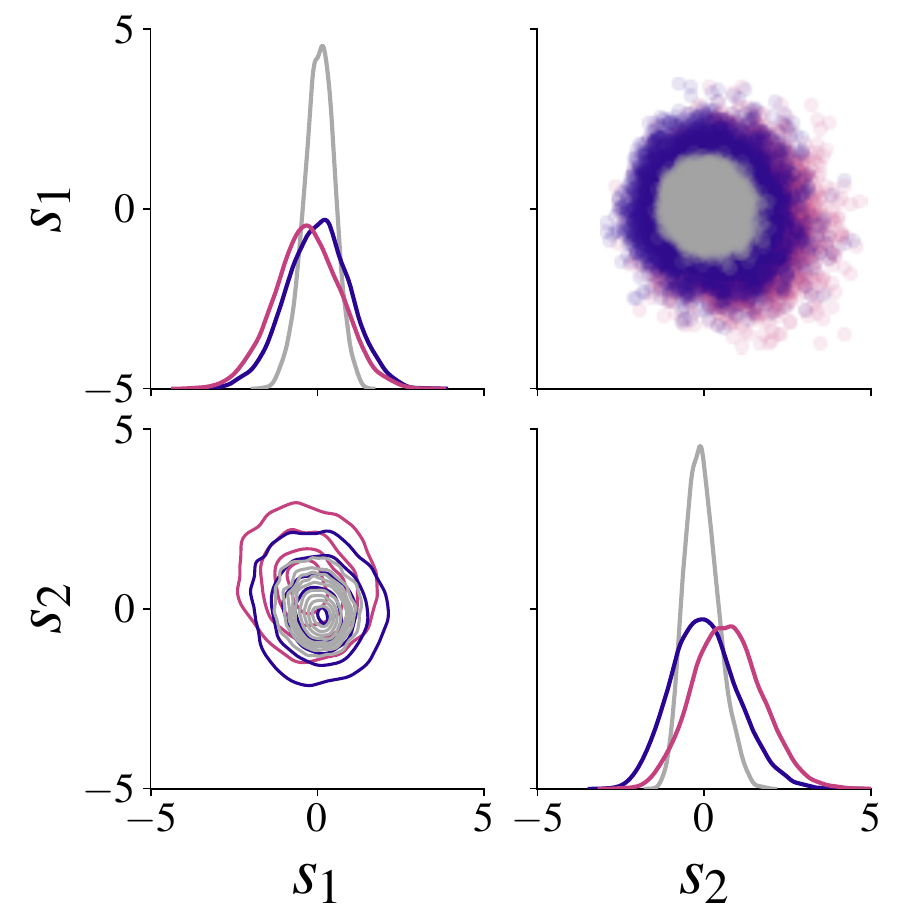}
    \end{subfigure}\\
    \includegraphics[width=0.80\linewidth]{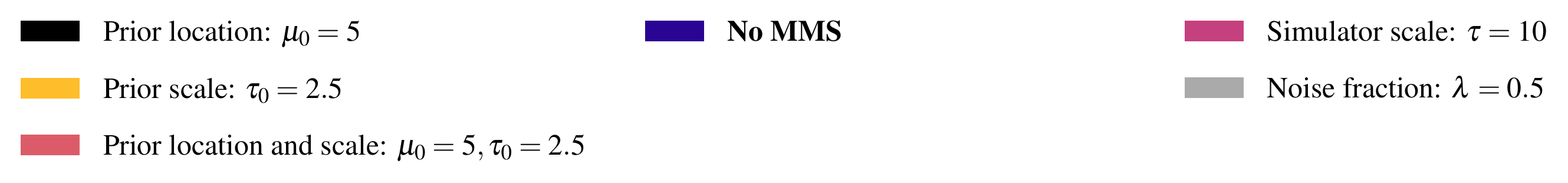}
    \caption{\textbf{Experiment \numberGaussianMeans.} Summary space samples for the minimal sufficient summary network ($S=2$) from a well-specified model $\M$ (blue) and several misspecified configurations. \textbf{Left:} Prior misspecification can be detected. \textbf{Right:} Noise misspecification can be detected, while simulator scale misspecification is indistinguishable from the validation summary statistics.}
    \label{fig:mvn:pairplot}
\end{figure}

\begin{table}[b]
    \centering
    \begin{tabular}{l|ll}
        \bfseries Model (MMS) &\bfseries Prior &\bfseries Likelihood\\
        \hline
        $\mathcal{M}$ (No MMS) &
        $\mub\sim\mathcal{N}(\0, \mathbb{I})$&
        $\x_k\sim\mathcal{N}(\mub, \mathbb{I})$
        \\
        $\mathcal{M}_P$ (Prior) &
        $\mub\sim\mathcal{N}(\mub_0, \tau_0\mathbb{I}), \tau_0\in\mathbb{R}^+$&
        $\x_k\sim\mathcal{N}(\mub, \mathbb{I})$\\
        $\mathcal{M}_S$ (Simulator) &
        $\mub\sim\mathcal{N}(\0, \mathbb{I})$&
        $\x_k\sim\mathcal{N}(\mub, \tau\mathbb{I}), \tau\in\mathbb{R}^+$
        \\
        $\mathcal{M}_N$ (Noise)&
        $\mub\sim\mathcal{N}(\0, \mathbb{I})$ & 
        $\x_k\sim\lambda\cdot\mathrm{Beta}(2, 5)+(1-\lambda)\cdot\mathcal{N}(\mub, \mathbb{I})%
        $
    \end{tabular}%
    \caption{\textbf{Experiment \numberGaussianMeans.} Investigated model misspecifications (MMS). $\mathcal{N}(\mub, \Sigmab)$ denotes a (multivariate) Gaussian distribution with location $\mub$ and covariance $\Sigmab$.}
    \label{tab:mvn-mean-misspecifications}
\end{table}
We set the stage by estimating the means of a $2$-dimensional conjugate Gaussian model with $K=100$ observations per data set and a known analytic posterior in order to illustrate our method.
This experiment contains the Gaussian examples from \citet{frazier_model_2020} and \citet{ward_robust_2022}, and extends them by (i) studying misspecifications beyond the likelihood variance (see below); and (ii) implementing continuously widening simulation gaps, as opposed to a single discrete misspecification.
The data generating process is defined as
\begin{equation}
    \x_k \sim \mathcal{N}(\x\given\mub, \Sigmab) \quad\text{for } k = 1,...,K\qquad \text{with }\;\;
    \mub \sim \mathcal{N}(\mub\given\mub_0, \Sigmab_0).
\end{equation}

As a summary network, we use a permutation invariant network \citep{invariant} with $S=2$ output dimensions, which equal the number of minimal sufficient statistics implied by the analytic posterior.
The terms ``minimal'', ``sufficient'', and ``overcomplete'' refer to the inference task and \emph{not} to the data. Thus, $S=2$ summary statistics are \emph{sufficient} to solve the inference task, namely recover two means.
For training the posterior approximator, we set the prior of the generative model $\M$ to a unit Gaussian and the likelihood covariance $\Sigmab$ to an identity matrix.

We induce prior misspecification by altering the prior location $\mub_0$ and covariance $\Sigmab_0=\tau_0\mathbb{I}$ (diagonal covariance, controlled through the factor $\tau_0$).
Further, we achieve misspecified likelihoods by manipulating the likelihood covariance $\Sigmab=\tau\mathbb{I}$ (diagonal covariance, controlled through $\tau$).
We induce noise misspecification by replacing a fraction $\lambda\in[0,1]$ of data $\x$ with samples from a scaled $\text{Beta}(2,5)$ distribution.

\begin{figure}[t]
    \centering
    \begin{subfigure}[b]{0.49\linewidth}
    \begin{subfigure}[c]{0.9\linewidth}%
    \setlength\tabcolsep{2pt}%
    \begin{tabular}{ccccc}
    && \multicolumn{2}{c}{\textbf{Model Misspecification}} & \\
        & &
        \textbf{Prior} &
        \textbf{Simulator} \textbf{\& noise} 
        \\
        \multirow{2}{*}{\hspace*{-0.1cm}\rotatebox[origin=c]{90}{\textbf{Summary Network}}} &
        \rotatebox[origin=c]{90}{\textbf{minimal}} &
        \raisebox{-0.48\height}{\includegraphics[width=0.45\linewidth]{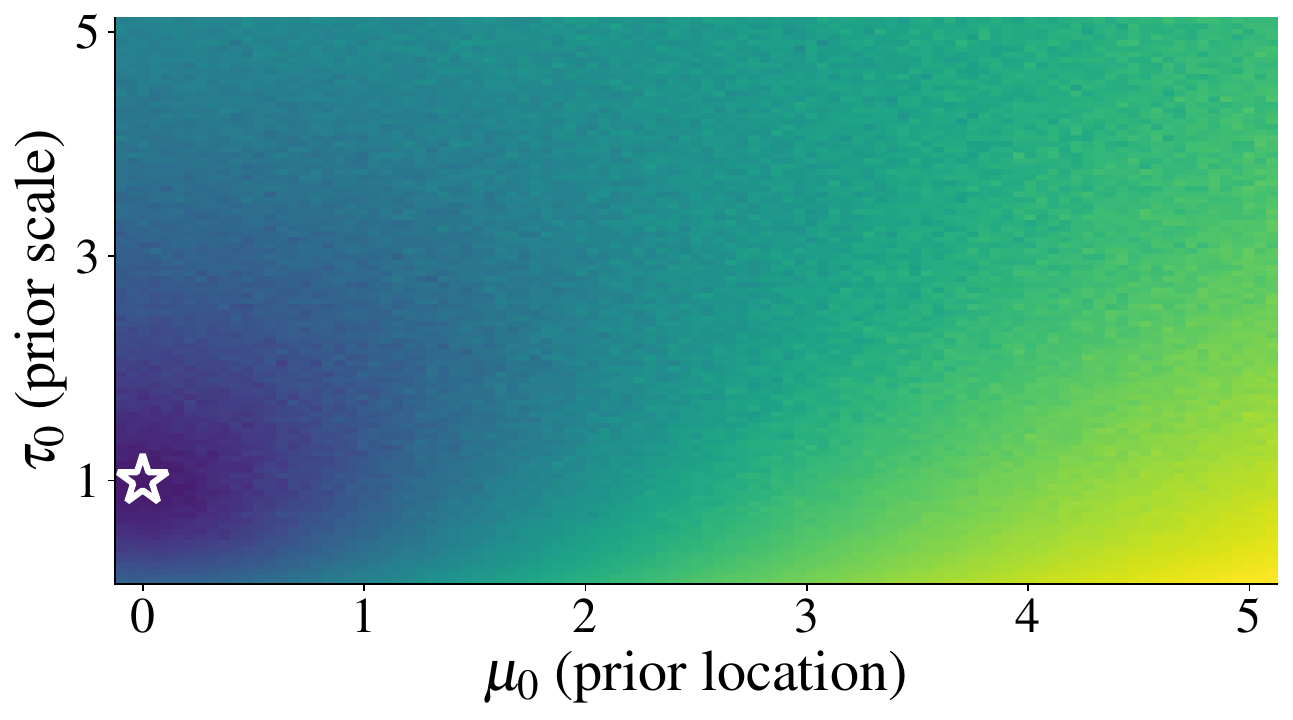}} &
        \raisebox{-0.48\height}{\includegraphics[width=0.46\linewidth]{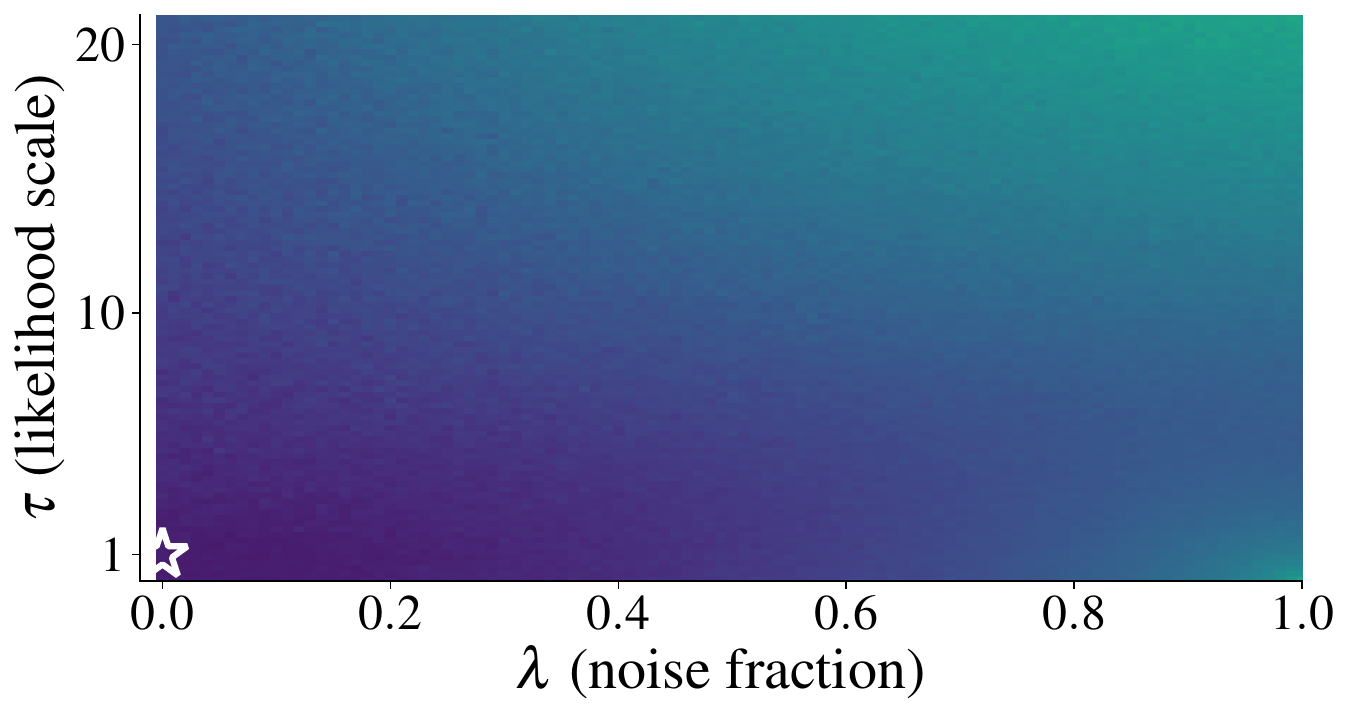}}
        \\
        &\rotatebox[origin=c]{90}{\textbf{overcomplete}} &
        \raisebox{-0.5\height}{\includegraphics[width=0.45\linewidth]{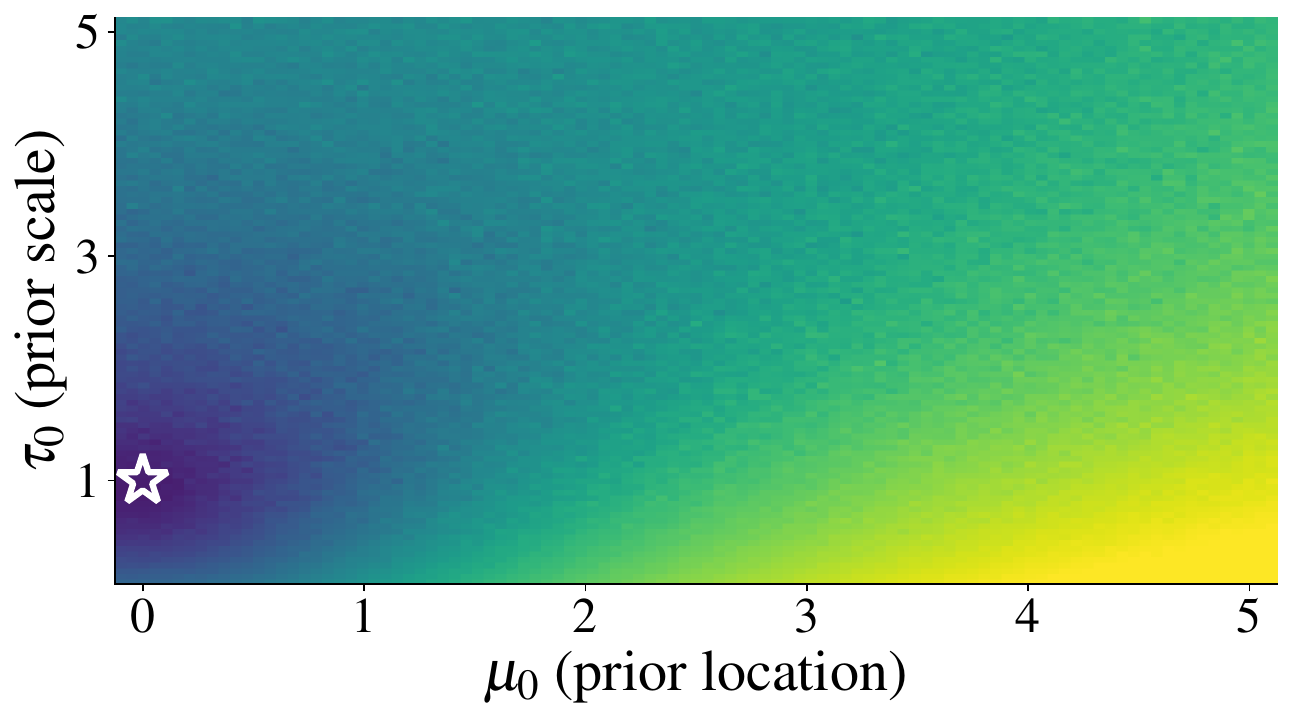}} &
        \raisebox{-0.5\height}{\includegraphics[width=0.46\linewidth]{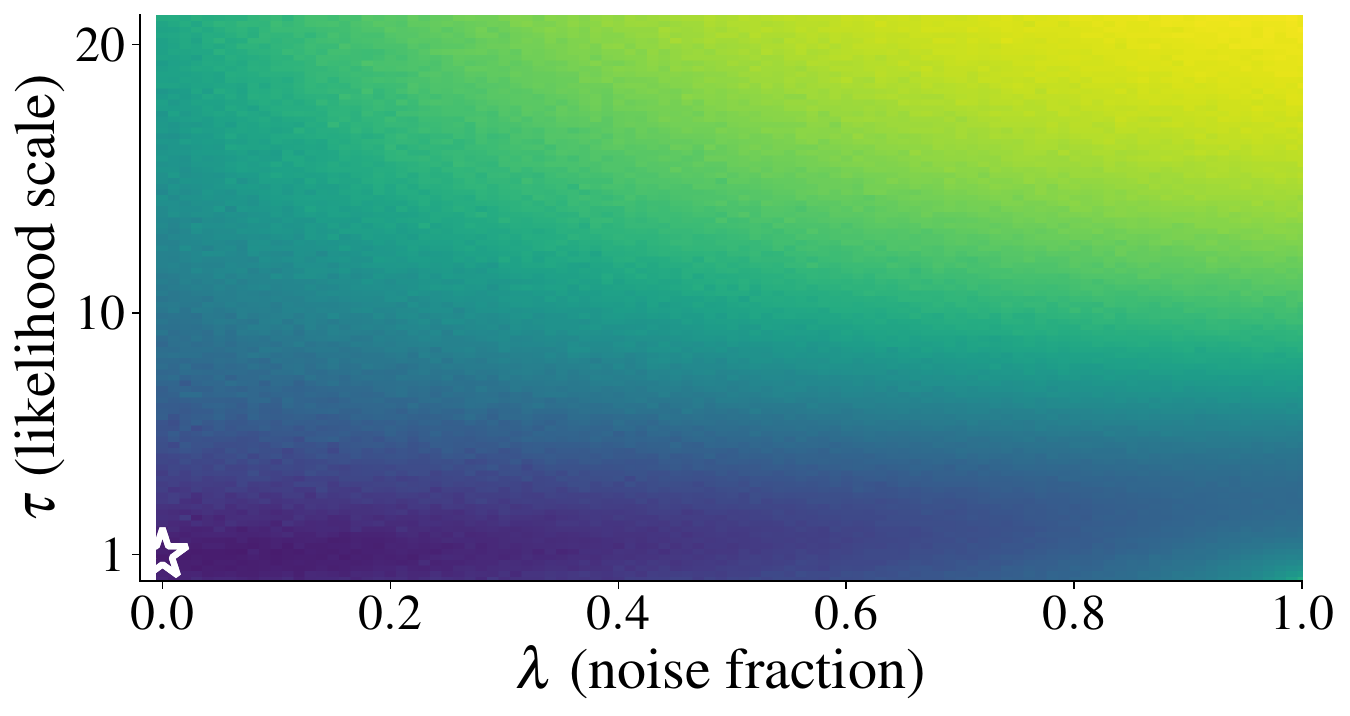}}
    \end{tabular}%
    \end{subfigure}%
    \begin{subfigure}[c]{0.07\linewidth}
    \hspace*{3mm}\includegraphics[width=\linewidth, clip, trim=9.8cm 0cm 0.2cm 0cm]{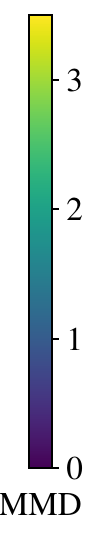}
    \end{subfigure}
    \caption{Summary MMD by misspecification severity.}
    \label{fig:mvn:mmd}
    \end{subfigure}
\begin{subfigure}[b]{0.49\linewidth}
    \begin{subfigure}[c]{0.9\linewidth}%
    \setlength\tabcolsep{2pt}%
    \begin{tabular}{ccccc}
    && \multicolumn{2}{c}{\textbf{Model Misspecification}} & \\
        & &
        \textbf{Prior} &
        \textbf{Simulator \& noise} 
        \\
        \multirow{2}{*}{\hspace*{-0.1cm}\rotatebox[origin=c]{90}{\textbf{Summary Network}}} &
        \rotatebox[origin=c]{90}{\textbf{minimal}} &
        \raisebox{-0.48\height}{\includegraphics[width=0.45\linewidth]{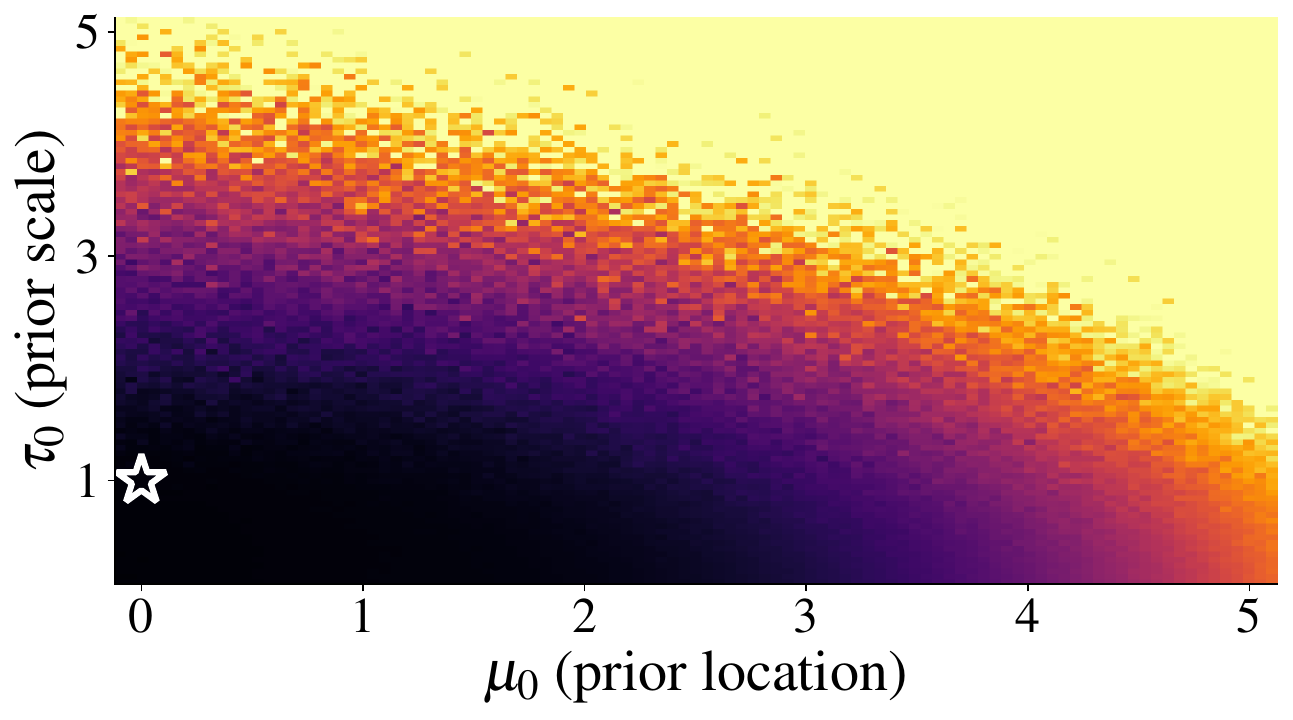}} &
        \raisebox{-0.48\height}{\includegraphics[width=0.46\linewidth]{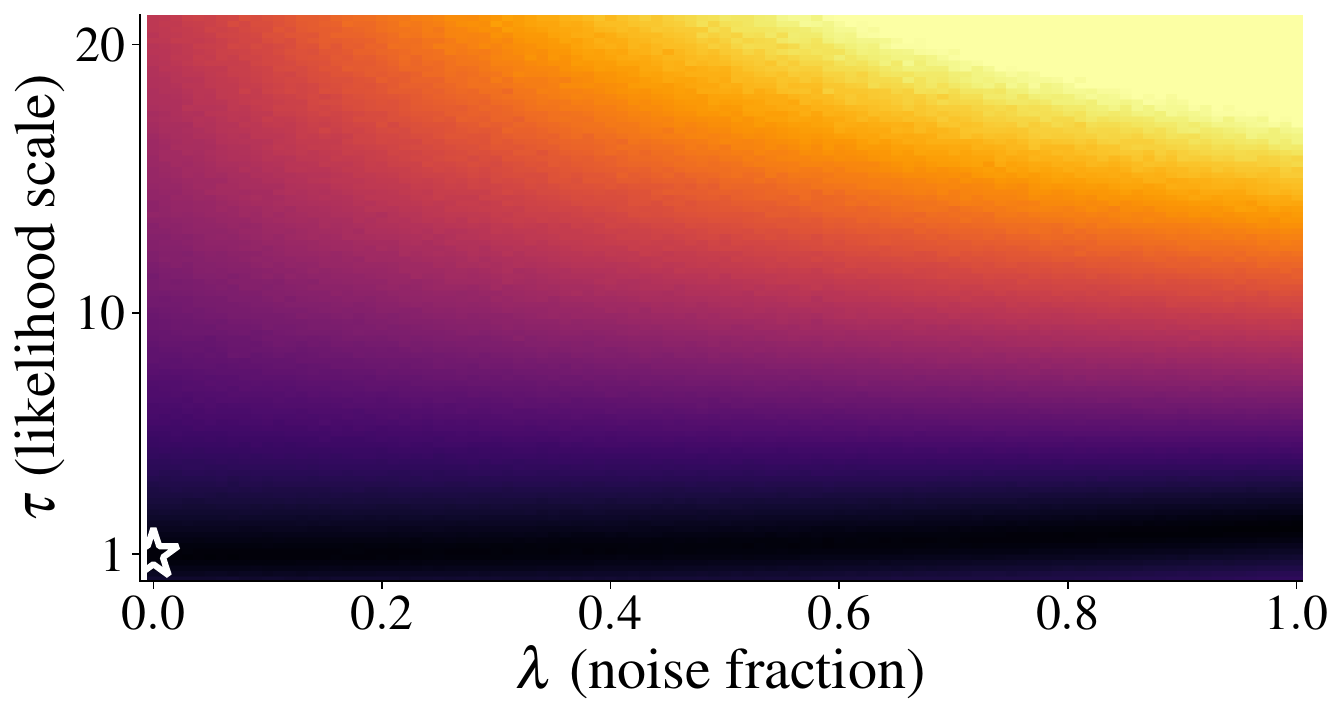}}
        \\
        &\rotatebox[origin=c]{90}{\textbf{overcomplete}} &
        \raisebox{-0.5\height}{\includegraphics[width=0.45\linewidth]{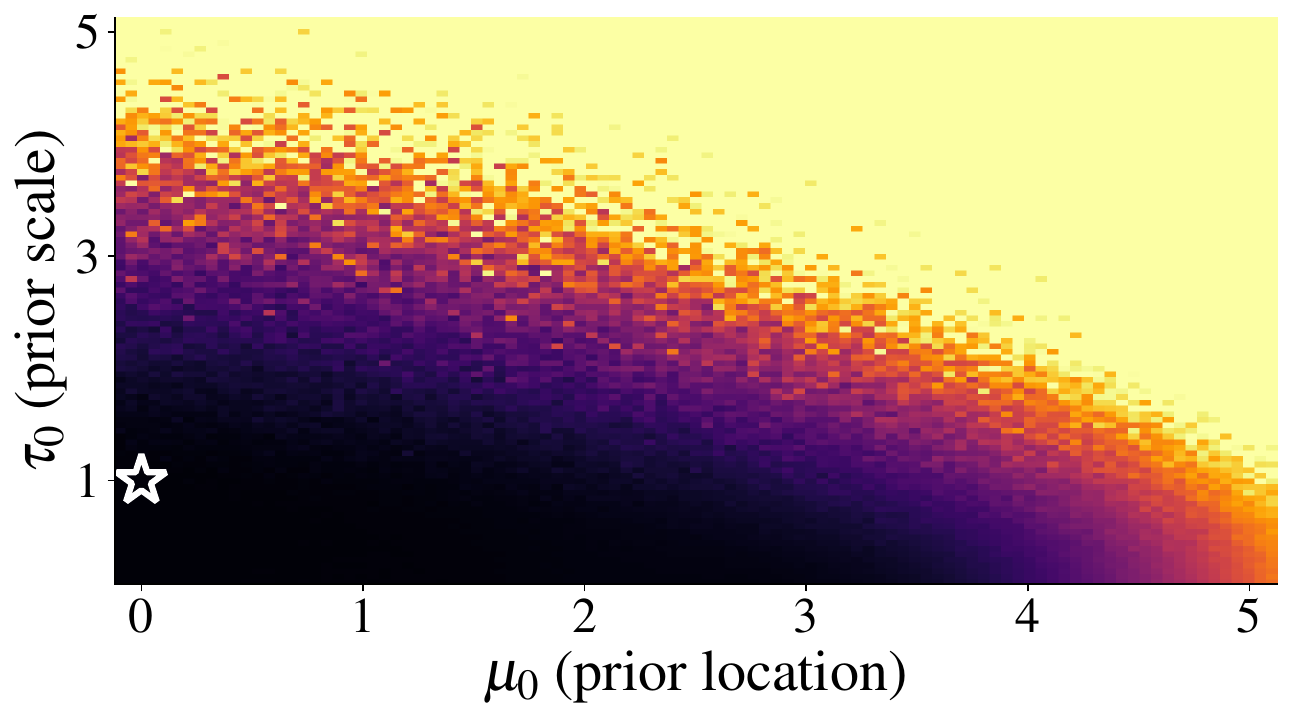}} &
        \raisebox{-0.5\height}{\includegraphics[width=0.46\linewidth]{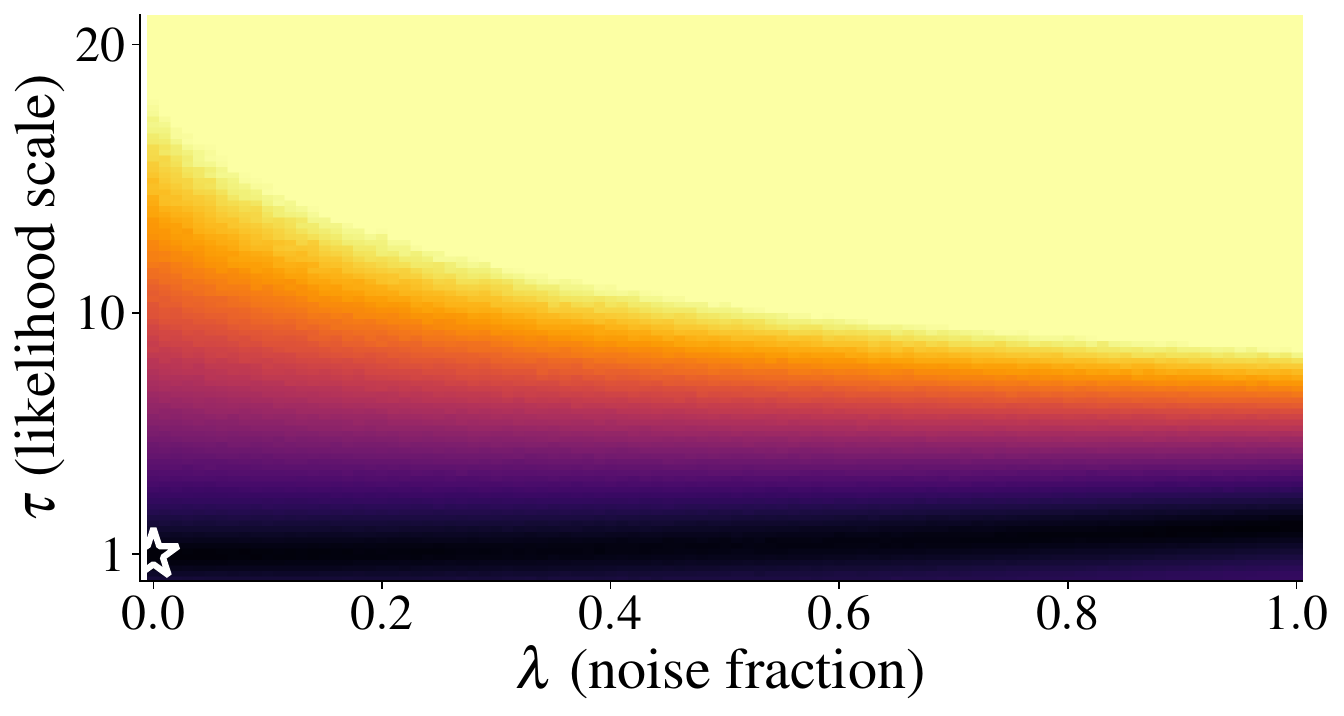}}
    \end{tabular}%
    \end{subfigure}%
    \begin{subfigure}[c]{0.09\linewidth}
    \hspace*{3mm}\includegraphics[width=\linewidth, clip, trim=9.5cm 0cm 0.2cm 0cm]{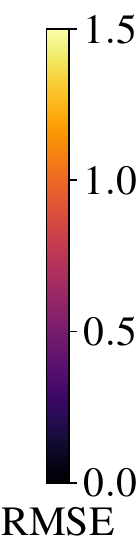}%
    \end{subfigure}
    \caption{Posterior error by misspecification severity.}
    \label{fig:mvn:rmse}
\end{subfigure}
    \caption{\textbf{Experiment \numberGaussianMeans.} Summary space discrepancy (MMD to training distribution) and posterior error (RMSE of correct vs.\ analytic posterior means) as a function of misspecification severity. White stars indicate the well-specified model configuration (i.e., equal to the training model $\M$), where both MMD and posterior error are low.}
\end{figure}

\textit{Results.}
The neural posterior estimator trained to minimize the augmented objective (Eq.~\ref{eq:bf_kl_mmd}) exhibits excellent recovery and calibration \citep{talts_validating_2020,sailynoja_graphical_2021} in the well-specified case, as shown in the Appendix.
All prior misspecifications manifest themselves in anomalies in the summary space which are directly detectable through visual inspection of the $2$-dimensional summary space in \autoref{fig:mvn:pairplot} (left).
Note that the combined prior misspecification (location and scale) exhibits a summary space pattern that combines the location and scale of the respective location and scale misspecifications.
However, based on the $2$-dimensional summary space, misspecifications in the fixed parameters of the likelihood ($\tau$) and mixture noise are not detectable via an increased MMD (see \autoref{fig:mvn:mmd}, top right).

We further investigate the effect of an \emph{overcomplete} summary space, namely $S=4$ learned summary statistics with an otherwise equal architecture.
In addition to prior misspecifications, the overcomplete summary network also captures misspecifications in the noise and simulator via the MMD criterion (see \autoref{fig:mvn:mmd}, bottom row).
Furthermore, the misspecifications in the noise and simulator are visually detectable in the summary space (see Appendix).
Recall that the $2$-dimensional summary space fails to capture these misspecifications (see \autoref{fig:mvn:mmd}, top right).

\textit{Posterior Errors.}
While detecting model misspecification might be valuable by itself for applied modelers, one fundamental question remains: \textit{Are the inference results of NPE even impaired by model misspecification?}
To answer this question, we compute the error in posterior recovery as a function of the misspecification severity.
To ease visualization, we use the RMSE of the approximated posterior mean from $L$ samples $\hat{\overline{\thetab}}^{(n)} = \frac{1}{L}\sum_{l=1}^L\thetab^{(l, n)}$ against the analytic posterior mean $\overline{\thetab}^{(n)}$ across $N$ data sets.\footnote{Since the approximate posterior in the Gaussian model is likely to be symmetric---and the analytic posterior is symmetric by definition---we deem the posterior mean as an appropriate evaluation target for the RMSE across data sets.
In fact, error metrics over several posterior quantiles (i.e., $Q_{25}$, $Q_{50}$ and $Q_{75}$) in the place of posterior means yield identical results.
Other common metrics (i.e., MSE and MAE) yield identical results.}

\autoref{fig:mvn:rmse} illustrates that more severe model misspecifications generally coincide with a larger error in posterior estimation across all model misspecifications for both $S=2$ and $S=4$ learned summary statistics, albeit with fundamental differences, as explained in the following.
When processing data emerging from models with misspecified noise and simulator (see \autoref{fig:mvn:rmse}, right column), minimal and overcomplete summary networks exhibit a drastically different behavior:
While the minimal summary network cannot detect noise or simulator simulation gaps, its posterior estimation performance is not heavily impaired either (see \autoref{fig:mvn:rmse}, top right).
On the other hand, the overcomplete summary network is able to capture noise and simulator misspecifications, but also incurs larger posterior inference error (see \autoref{fig:mvn:rmse}, bottom right).
This might suggest a trade-off between model misspecification detection and posterior inference error, depending on the number of learnable summary statistics.

From a practical modeling perspective, researchers might wonder how to choose the number of learnable summary statistics.
While an intuitive heuristic might suggest ``the more, the merrier'', the observed results in this experiment beg to differ depending on the modeling goals.
If the focus in a critical application lies in detecting potential simulation gaps, it might be advantageous to utilize a large (overcomplete) summary vector.
However, modelers might also desire a network which is as robust as possible during test time, opting for a smaller number of summary statistics.
\textbf{Experiments \numberCovid} addresses this question for a complex non-linear time series model where the number of sufficient summary statistics is unknown.

\textit{Replication with the SNPE-C method.}
Our method successfully detects model misspecification when we use non-amortized sequential neural posterior estimation \citep[SNPE-C;][]{apt} with a structured summary space (see Appendix for details).
The results are largely equivalent to those obtained with amortized NPE, as implemented in the BayesFlow framework \citep{radev2023joss}.

\subsection{Experiment \numberCS: Cancer and Stromal Cell Model}

This experiment illustrates model misspecification detection in a marked point process model of cancer and stromal cells \citep{jones-todd_identifying_2019}.
We use the implementation of \citet{ward_robust_2022} with hand-crafted summary statistics and showcase the applicability of our method in scenarios where good summary statistics are already known.
The inference parameters are three Poisson rates $\lambda_c, \lambda_p, \lambda_d$, and the setup in \citet{ward_robust_2022} extracts four hand-crafted summary statistics from the 2D plane data: (1--2) number of cancer and stromal cells; (3--4) mean and maximum distance from stromal cells to the nearest cancer cell.

We achieve misspecification during inference by mimicking necrosis, which often occurs in core regions of tumors.
A Bernoulli distribution with parameter $\pi$ controls whether a cell is affected by necrosis or not.
Consequently, $\pi=0$ implies no necrosis (and thus no simulation gap), and $\pi=1$ entails that all cells are affected.
The experiments by \citet{ward_robust_2022} study a single misspecification scenario, namely the case $\pi=0.75$ in our implementation.
In order to employ our proposed method for model misspecification detection, we add a small summary network $h_{\psib}:\mathbb{R}^4\to\mathbb{R}^4$ consisting of three hidden fully connected layers with $64$ units each.
This network $h_{\psib}$ merely transforms the hand-crafted summary statistics into a $4$-D unit Gaussian, followed by NPE for posterior inference.

\begin{figure}[t]
    \centering%
        \begin{subfigure}[t]{0.45\linewidth}%
            \includegraphics[width=\linewidth]{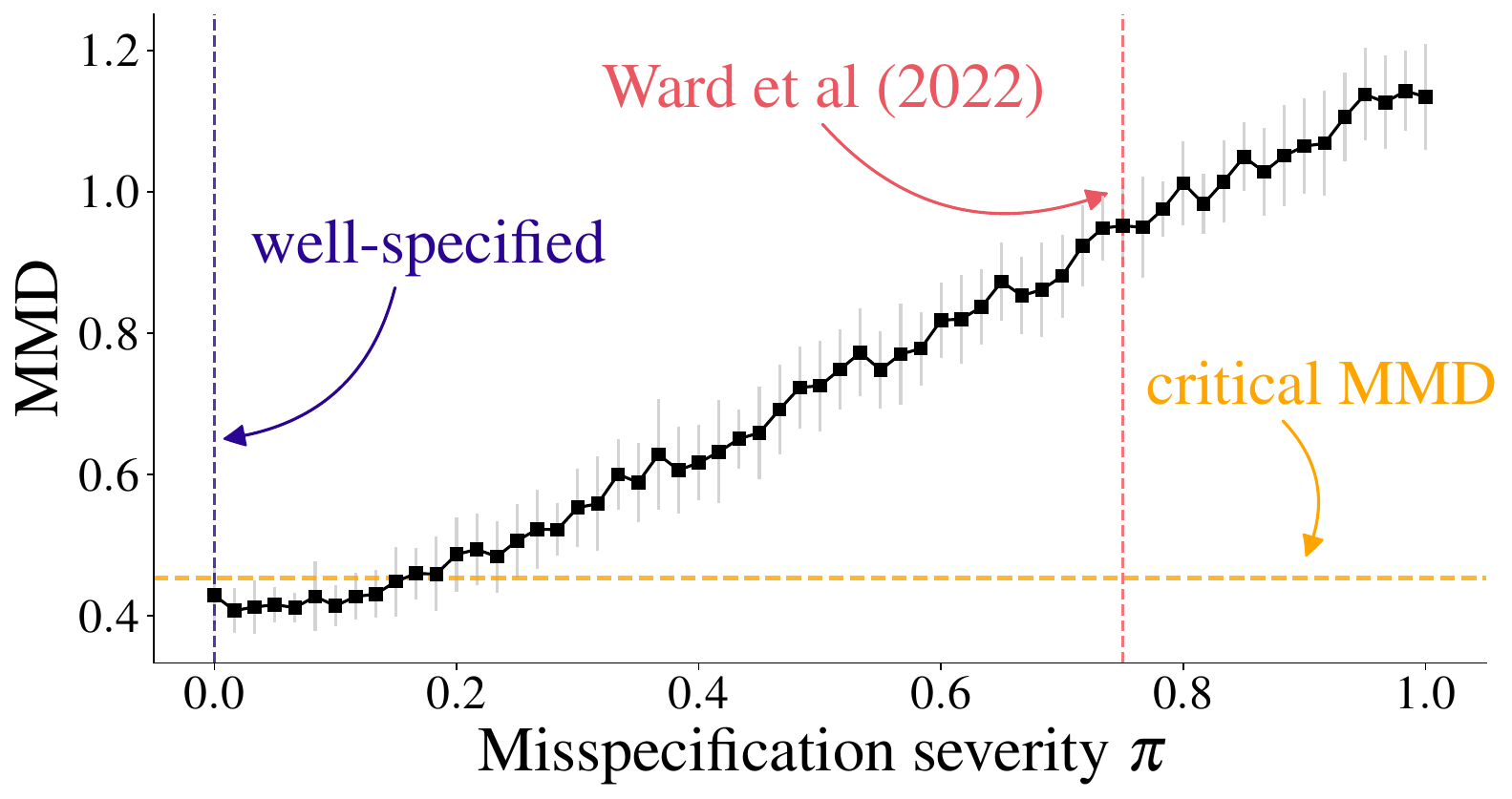}%
            \caption{MMD by misspecification severity.}%
            \label{fig:cs:mms-mmd}%
        \end{subfigure}
        \hfill
        \begin{subfigure}[t]{0.50\linewidth}%
                \includegraphics[width=\linewidth]{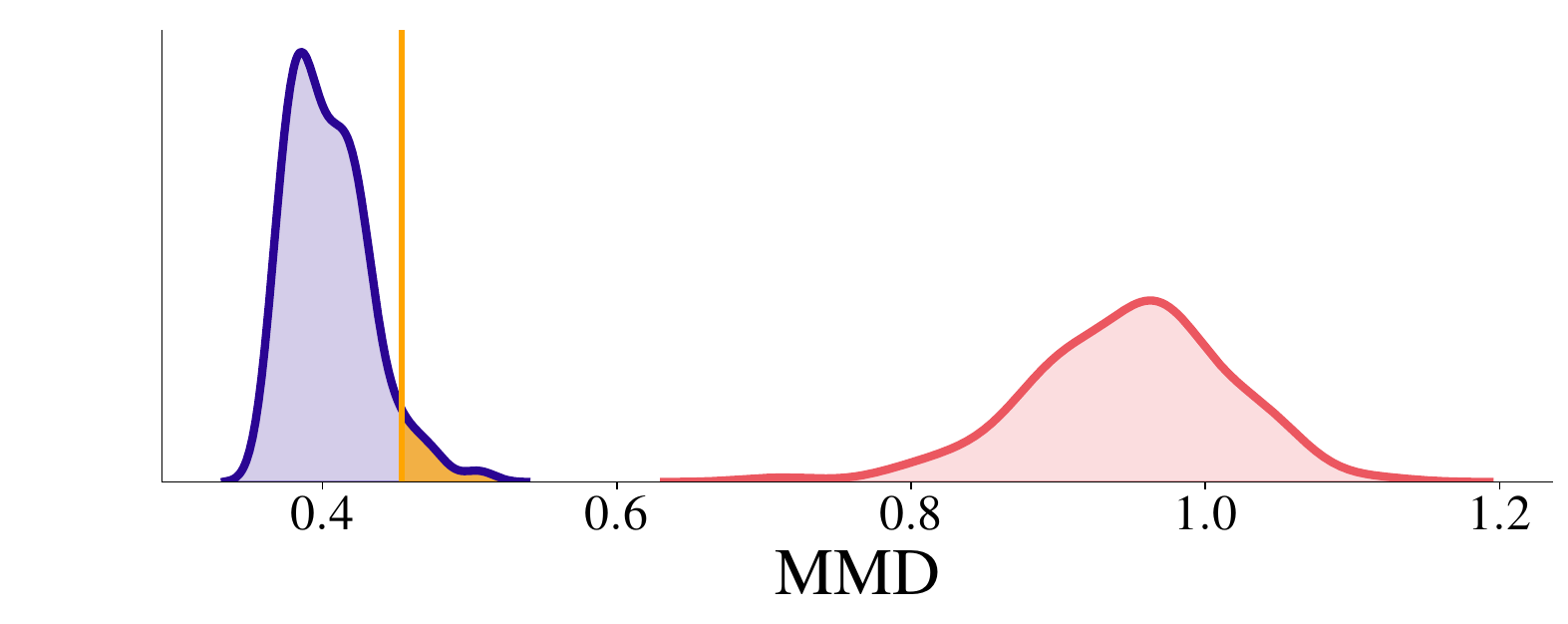}%
                \caption{Power for $\pi=0.75$ is essentially 1.}%
            \label{fig:cs:power}%
        \end{subfigure}\\
    \includegraphics[width=0.8\linewidth]{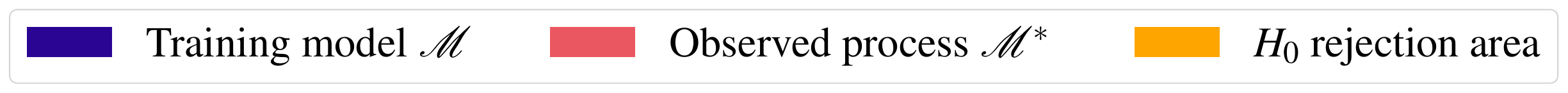}
    \caption{\textbf{Experiment \numberCS.} MMD increases with misspecification severity (\subref{fig:cs:mms-mmd}; mean, SD of 20 repetitions). Our test easily detects the setting from \citet{ward_robust_2022} (\subref{fig:cs:power}).
    }
\end{figure}

\textit{Results.}
Our MMD misspecification score increases with increasingly severe model misspecification (i.e., increasing necrosis rate $\pi$), see \autoref{fig:cs:mms-mmd}.
What is more, for the single misspecification $\pi=0.75$ studied by \citet{ward_robust_2022}, we illustrate (i) the power of our proposed hypothesis test; and (ii) the summary space distribution for misspecified data.
The power ($1-\beta$) essentially equals $1$, as shown in \autoref{fig:cs:power}: The MMD sampling distributions under the training model ($H_0$) and under the observed data generating process ($\M^*$) are completely separated.

\subsection{Experiment \numberDDM: Drift-Diffusion Model of Decision Making}
\label{sec:ddm-experiment} 
\begin{wrapfigure}[18]{r}{0.6\textwidth}
\vspace{-1.6em}
    \centering
    \includegraphics[width=\linewidth]{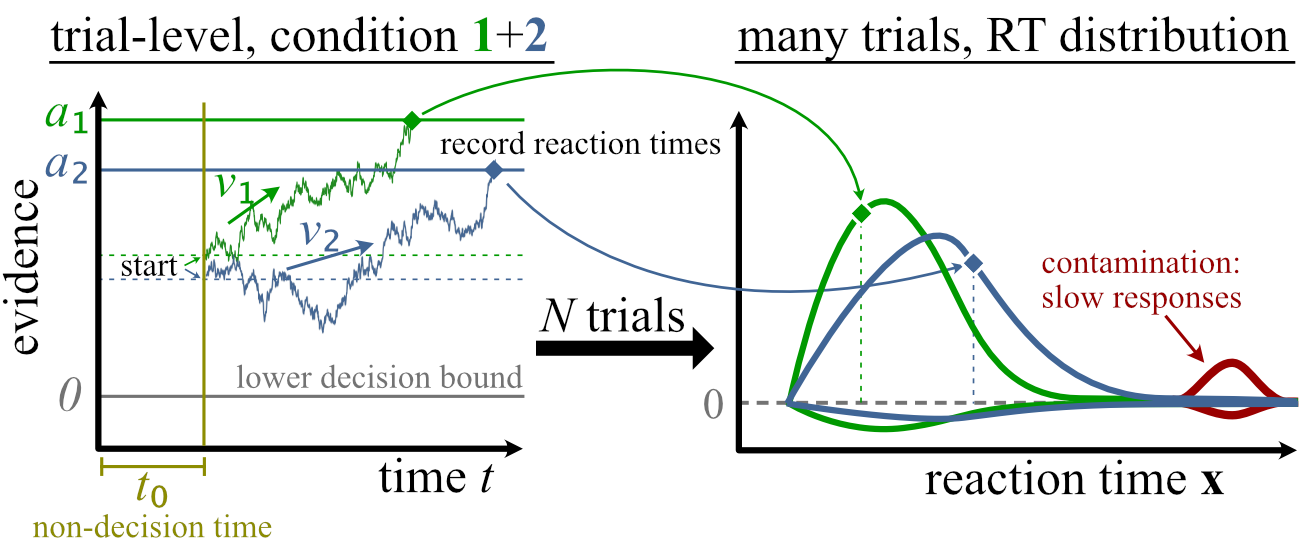}
    \caption{\textbf{Experiment 3.} In each trial, the drift-diffusion model (DDM) models the information uptake in experimental condition $j\in\{1,2\}$ as a random walk with drift $v_j$ and initial non-decision time $t_0$.
    Once the evidence reaches a decision threshold $a_j$, the associated reaction time is recorded. After $N$ trials, the $i.\,i.\,d.\,$ set of reaction times $\x$ is used to infer the mechanistic parameters $\thetab=(v_1, v_2, a_1, a_2, t_0)$.}
    \label{fig:ddm-overview}
\end{wrapfigure}
This experiment aims to (i) apply the new optimization objective to a complex model of human decision making; (ii) tackle strategies to determine the required number of learned summary statistics in more complex applications; and (iii) compare the posterior estimation of amortized NPE under a misspecified model with a reference posterior from the gold-standard non-amortized Markov Chain Monte Carlo algorithm (HMC-MCMC).%

We focus on the drift diffusion model (DDM)---a cognitive model describing reaction times (RTs) in binary decision tasks \citep{Ratcliff2008} which is well amenable to amortized inference \citep{Radev2020bayesflow-cognition}.
The DDM assumes that perceptual information for a choice alternative accumulates continuously according to a Wiener diffusion process. 
Thus, the change in information in experimental condition $j$ follows a random walk with drift and Gaussian noise: $v\mathrm{d}t + \xi \sqrt{\mathrm{d}t}$ with $\xi\sim\mathcal{N}(0, 1)$.
Our model implementation features five free parameters $\thetab = (v_1, v_2, a_1, a_2, t_0)$ which produce data stemming from two simulated conditions.
The summary network is a permutation-invariant network which reduces $i.\,i.\,d.\ $RT data sets to $S=10$ summary statistics each.

We realize a simulation gap by simulating typically observed contaminants: fast guesses (e.g., due to inattention), very slow responses (e.g., due to mind wandering), or a combination of the two.
The parameter $\lambda$ controls the fraction of the observed data which is contaminated  (see Section~\ref{sec:app:ddm} in the Appendix for implementation details).
For the comparison with the MCMC reference posterior from Stan, we simulate $100$ uncontaminated DDM data sets from the true model $\M$, as well as three scenarios (fast guesses, slow responses, fast and slow combined) with a fraction of $\lambda = 10\%$ of contaminated observations.

\begin{figure}[t]
    \begin{subfigure}[t]{0.33\linewidth}
        \includegraphics[width=\linewidth]{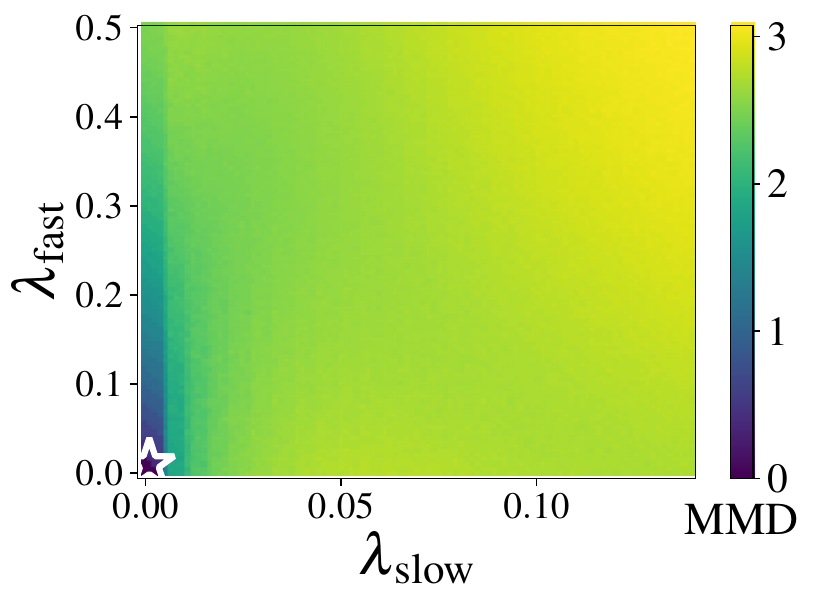}
        \caption{MMD as a function of misspecification severity.}
        \label{fig:exp:ddm-mmd-contamination}
    \end{subfigure}
    \hfill
    \begin{subfigure}[t]{0.62\linewidth}
        \includegraphics[width=\linewidth]{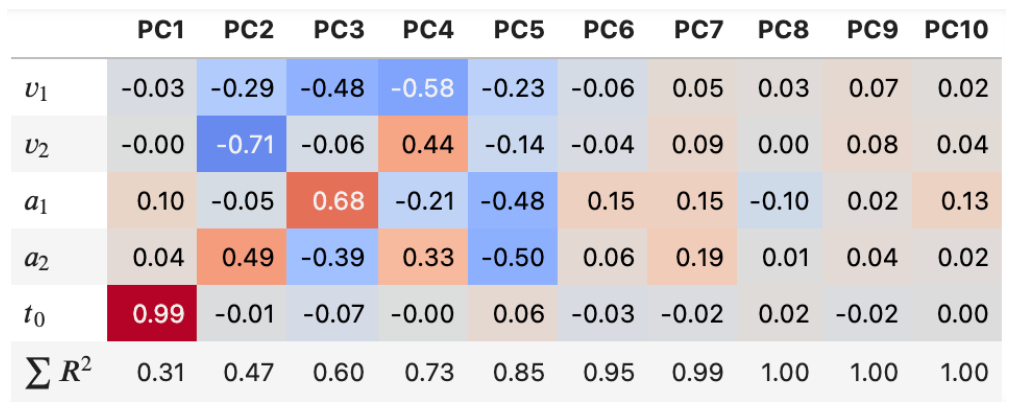}
        \caption{Correlation between parameters $\thetab$ and principal components (PCs) of learned summary statistics.}
        \label{fig:exp:ddm-pca}
    \end{subfigure}
    \caption{\textbf{Experiment \numberDDM.} Contamination of reaction times is detectable with our method (\subref{fig:exp:ddm-mmd-contamination}), and the principal components from the learned summary statistics coincide with the true parameters $\thetab$ (\subref{fig:exp:ddm-pca}).
    White star indicates the configuration for the well-specified model (i.e., training model $\M$ without contamination).}
    \label{fig:exp:ddm-figure}
\end{figure}

\textit{Results.} During inference, our criterion reliably detects the induced misspecifications: Increasing fractions $\lambda$ of contaminants (fast, slow, and combined) manifest themselves in increasing MMD values (see \autoref{fig:exp:ddm-mmd-contamination}).
The results of applying PCA to the summary network outputs $\{\observed{\z}^{(n)}\}$ for the well-specified model (no contamination) are illustrated in \autoref{fig:exp:ddm-pca}.
We observe that the first five principal components exhibit a large overlap with the true model parameters $\thetab$ and jointly account for 85\% of the variance in the summary output.
Furthermore, the drift rates and decision thresholds within conditions are entangled (i.e., $v_1, a_1$ and $v_2, a_2$).
This entanglement mimics the strong posterior correlations observed between these two parameters.
In practical applications, dimensionality reduction might act as a guideline for determining the number of minimally sufficient summary statistics or parameter redundancies for a given problem.

\begin{table}[b]
    \centering
    \renewcommand{\arraystretch}{1.1}
    \begin{tabular}{l|c|c}
        \textbf{Model (Contamination)} & \textbf{Posterior error} & \textbf{Summary space MMD}\\
        \hline
        $\M_{\ }$: True model                  & $0.25\,[0.13, 0.56]$    & $0.45\,[0.42, 0.52]$ \\
        $\M_{1}$: Fast contaminants               & $2.66\,[1.44, 3.40]$    & $2.68\,[2.61, 2.74]$ \\
        $\M_{2}$: Slow contaminants               & $0.55\,[0.23, 1.01]$    & $1.18\,[1.13, 1.26]$ \\
        $\M_{3}$: Fast \& slow contaminants      & $1.90\,[0.83, 3.18]$    & $2.33\,[2.19, 2.43]$
    \end{tabular}
    \caption{
    \textbf{Experiment \numberDDM.} Posterior error of the approximate neural posterior (MMD to reference MCMC posterior; median and 95\% CI).
    The bootstrapped MMD values for the summary statistics of the $100$ investigated data sets and $1\,000$ samples from the true uncontaminated model $\M$
    illustrate that posterior errors are mirrored by anomalies in the summary space and thus detectable.
    }
    \label{tab:ddm:stan-bf}
\end{table}

For the comparison with the MCM reference posterior from Stan, we juxtapose $4\,000$ samples from the approximate neural posterior with $4\,000$ samples obtained from the Stan sampler after ensuring MCMC convergence and sufficient sampling efficiency for each data set (see \autoref{fig:exp:ddm:stan-bf} for an illustration). 
Because Stan is currently considered state-of-the-art for likelihood-based Bayesian inference, we assume the Stan samples are representative of the \emph{correct posterior under the potentially misspecified model} (see Section~\ref{sec:posterior-inference-errors}).
In order to quantify the difference between the posterior samples from NPE (BayesFlow) and the MCMC reference (Stan), we use the MMD criterion as well.
When no model misspecification is present, the posterior samples from NPE (BayesFlow) and MCMC (Stan) match almost perfectly (see \autoref{fig:exp:ddm:stan-bf:clean}).
This means that our augmented optimization objective still enables correct posterior approximation under well-specified models.
In contrast, the results in \autoref{fig:exp:ddm:stan-bf:slow} and \autoref{tab:ddm:stan-bf} clearly indicate that the amortized posteriors deteriorate as a result of the induced misspecification.
Moreover, these results closely mirror the overall detectability of misspecification obtained by matching the summary representations of $1000$ data sets from the true uncontaminated process with the representations of the $100$ data sets for each of the above scenarios via MMD (see \autoref{tab:ddm:stan-bf}).
This confirms the hypothesis that our proposed MMD criterion for quantifying model misspecification acts as a proxy for the posterior estimation error, which is of hallmark importance but typically unknown in practice due to a lack of ground-truths.

\subsection{Experiment \numberCovid: Epidemiological Model for COVID-19}\label{sec:experiment-covid}

\begin{figure}[t]
    \centering
    \begin{subfigure}[t]{0.50\linewidth}
        \includegraphics[width=\linewidth]{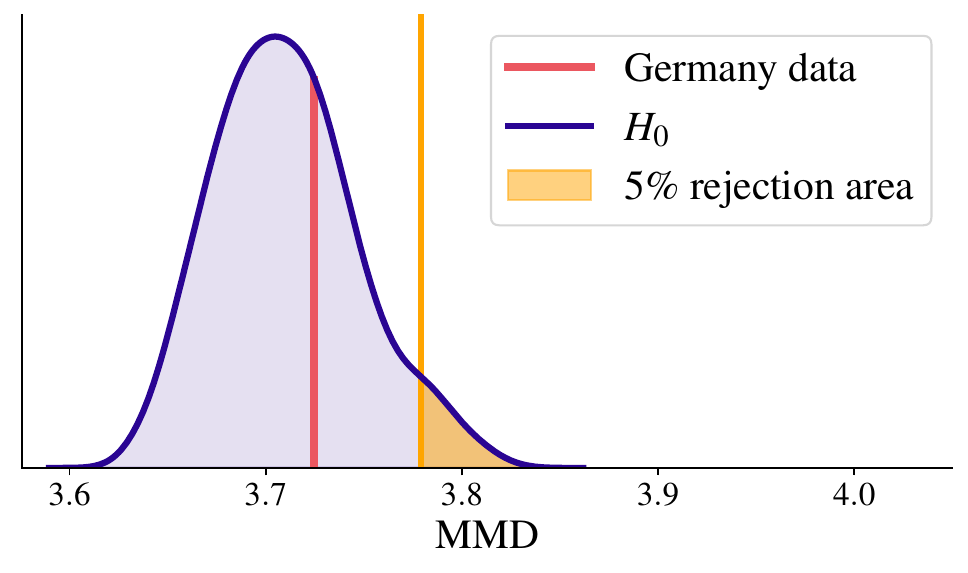}
        \caption{Representation of Germany's COVID-19 time series \mbox{w.\-r.\-t.} the MMD distribution under the null hypothesis $H_0: p^*(\x)=p(\x\given\mathcal{M})$.}
        \label{fig:exp:covid:mmd-real-data}        
    \end{subfigure}
    \hfill
    \begin{subfigure}[t]{0.44\linewidth}
        \includegraphics[width=\linewidth]{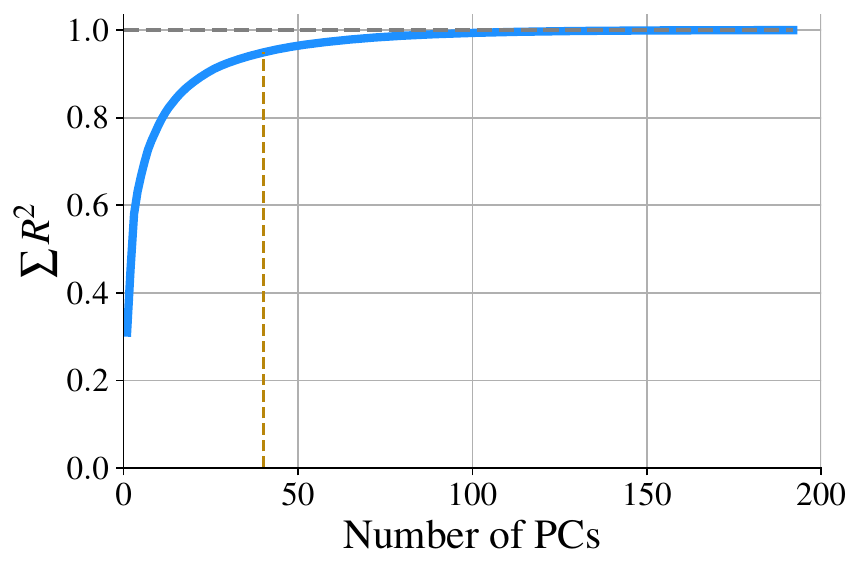}
        \caption{Cumulative explained variance ratio for the COVID-19 summary statistics as a function of the number of principal components (PCs).}
        \label{fig:exp:covid:PCA-exp-var}        
    \end{subfigure}
    \caption{\textbf{Experiment 4.} Our model misspecification hypothesis test indicates that the assumed model is well-specified for the real observed COVID data in Germany (a). Further, linear dimensionality reduction reveals that 40 principal components can explain 95\% of the variance in learned summary statistics.}
\end{figure}

As a challenging real-world example, we treat a high-dimensional compartmental model representing the early months of the COVID-19 pandemic in Germany \citep{outbreak}.
We investigate the utility of our method to detect simulation gaps in a much more realistic and non-trivial extension of the SIR settings in \citet{lueckmann_benchmarking_2021} and \citet{ward_robust_2022} with substantially increased complexity.
Moreover, we perform inference on real COVID-19 data from Germany and use our new method to test whether the model used in \citet{outbreak} is misspecified, possibly undermining the trustworthiness of political conclusions that are based on the inferred posteriors.
To achieve this, we train an NPE setup identical to \citet{outbreak} but using our new optimization objective (Eq.~\ref{eq:bf_kl_mmd}) to encourage a structured summary space.
We then simulate $1000$ time series from the training model $\mathcal{M}$ and $1000$ time series from three misspecified models: (i) a model $\mathcal{M}_1$ without an intervention sub-model; (ii) a model $\mathcal{M}_2$ without an observation sub-model; (iii) a model $\mathcal{M}_3$ without a ``carrier'' compartment \citep{covid_germany}.

\begin{table*}[b]
    \centering
    \renewcommand{\arraystretch}{1.2}
    \begin{tabular}{c|ccc|ccc}
        & 
        \multicolumn{3}{c|}{\bfseries Bootstrap MMD} & 
        \multicolumn{3}{c}{ \bfseries Power ($1-\beta$)} \\
        \textbf{Model}& $N=1$  & $N=2$ & $N=5$ & $N=1$ & $N=2$ & $N=5$\\
        \hline
        $\mathcal{M}_{\ }$ & $3.70\,[3.65, 3.79]$ & $2.61\,[2.54, 2.91]$ & $1.66\,[1.59, 1.84]$ & --- & --- & ---\\
        $\mathcal{M}_1$ & $3.76\,[3.72, 3.80]$ & $2.86\,[2.62, 3.16]$ & $2.11\,[1.82, 2.50]$ & $.998$ & $.958$ & $\approx1.0$ \\
        $\mathcal{M}_2$ & $3.80\,[3.73, 3.83]$ & $2.81\,[2.65, 3.00]$ & $2.01\,[1.82, 2.19]$ & $.789$ & $.804$ & $\approx1.0$ \\
        $\mathcal{M}_3$ & $3.78\,[3.74, 3.83]$ & $2.81\,[2.68, 3.11]$ & $2.07\,[1.92, 2.41]$ & $.631$ & $.690$ & $\approx1.0$ \\
    \end{tabular}
    \caption{
    \textbf{Experiment \numberCovid.} Results for different variations of the COVID-19 compartmental model (median and 95\% CI of 100 bootstrap samples).
    Our proposed detection criterion reliably flags data from misspecified models, with essentially no false-negatives for as few as $N=5$ observations.
    }
    \label{tab:covid19-models-mmd}
\end{table*}

\textit{Results.}
\autoref{tab:covid19-models-mmd} shows the MMD between the summary representation of $N=1,2,5$ bootstrapped time series from each model and the summary representation of the $1000$ time series from model $\mathcal{M}$ (see the Appendix for bootstrapping details).
We also calculate the power ($1-\beta$) of our hypothesis test for each misspecified model under the sampling distribution estimated from $1\,000$ samples of the $1\,000$ time series from $\mathcal{M}$ at a type I error probability of $\alpha=.05$.
We observe that the power of the test rapidly increases with more data sets and the Type II error probability ($\beta$) is essentially zero for as few as $N=5$ time series.

As a next step, we pass the reported COVID-19 data between 1 March and 21 April 2020 \citep{dong_interactive_2020} through the summary network and compute the critical MMD value for a sampling-based hypothesis test with an $\alpha$ level of $.05$ (see \autoref{fig:exp:covid:mmd-real-data}). 
The MMD of the Germany data is well below the critical MMD value---it essentially lies in the bulk of the distribution.
Hence, we conclude that the assumed training model $\M$ is well-specified for this time period.
Finally, we perform linear dimensionality reduction (PCA) on the summary space and find that the first 40 principal components jointly explain $95\%$ of the variance in the $192$-dimensional summary space outputs (see \autoref{fig:exp:covid:PCA-exp-var}). 
Thus, a $40$-dimensional learned summary vector might provide a good approximation of the true (unknown) minimally sufficient summary statistics and render inference less fragile under potential misspecifications (cf.\ \textbf{Experiment \numberGaussianMeans}).

\newpage
\subsection{Experiment \numberGIN: Style Inference of EMNIST Image Simulator}

\begin{figure*}
    \includegraphics[width=\linewidth, trim = 0cm 3cm 0cm 0cm, clip]{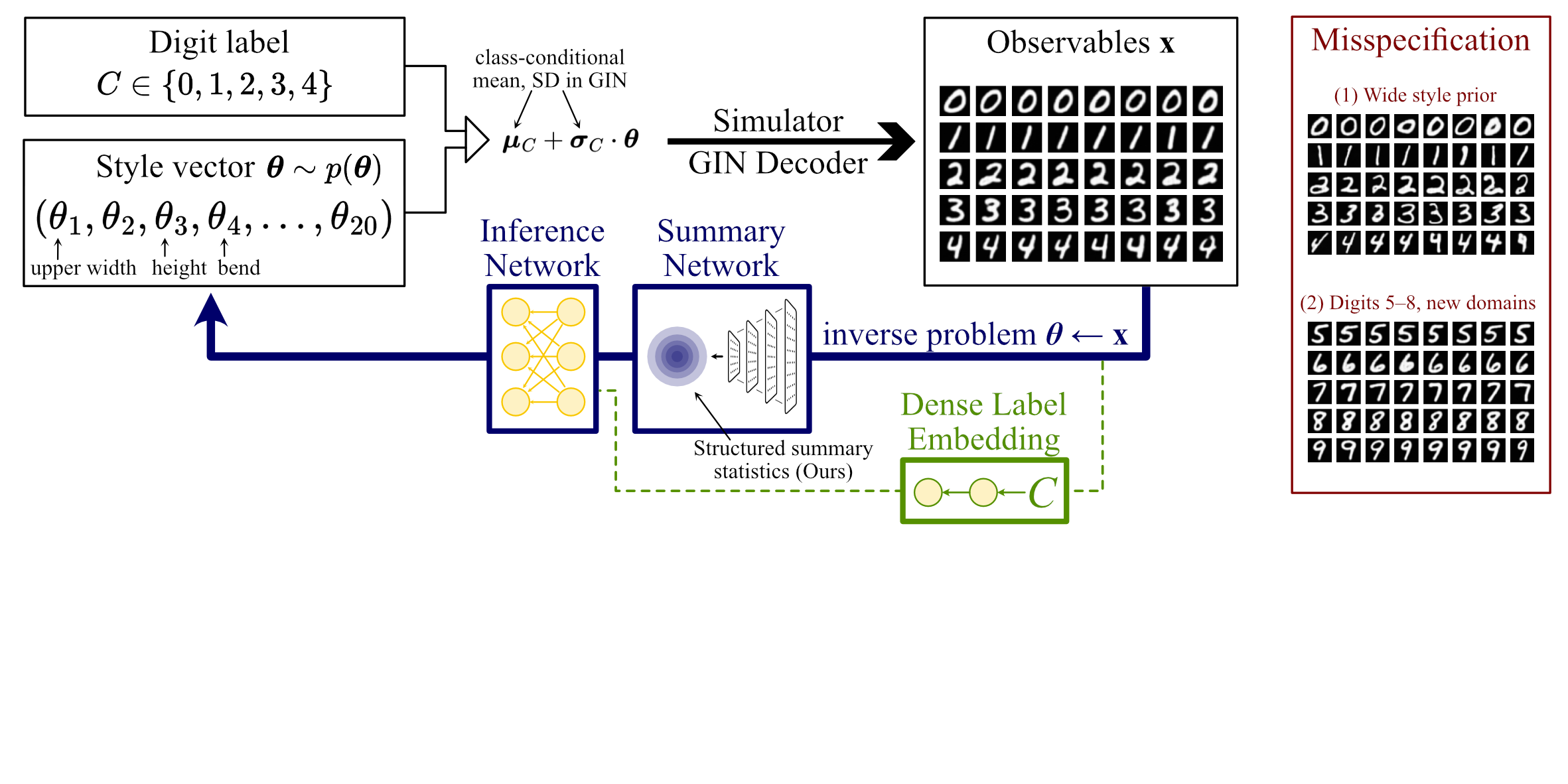}
    \caption{\textbf{Experiment 5.} Overview of the training objective.
    The latent style vector $\thetab\in\mathbb{R}^{20}$ and the digit label $C$ parameterize the GIN decoder \citep{Sorrenson2020GIN} which simulates a realistic hand-drawn digit $\x$.
    The \textbf{\color{darkblue}inverse problem} aims to recover the posterior distribution $p(\thetab\given\x)$ of plausible styles given an observed image $\x$.
    Our method can detect anomalies in the structured learned summary statistics that originate from (i) \textbf{\textcolor[rgb]{0.5, 0, 0}{atypical styles}} due to a wider prior $p(\thetab)$; and (ii) \textbf{\textcolor[rgb]{0.5, 0, 0}{domain shifts to new digits}}.
    In 
    \textbf{\textcolor[rgb]{0.388, 0.557, 0.157}{Experiment 2}}, we additionally pass the embedded digit label to the inference network to mimic knowledge about a domain shift in practical applications.
    }
    \label{fig:emnist-overview}
\end{figure*}

As a final experiment, we demonstrate the sensitivity of our detection method on a pre-trained generative network which we treat as an image simulator (i.e., an implicit likelihood model), see \autoref{fig:emnist-overview} for an overview.
We use a general incompressible-flow network \citep[GIN;][]{Sorrenson2020GIN} which has been trained on the digits of the EMNIST data set \citep{cohen2017emnist}.
The GIN learns a class-conditional bijective mapping from the image domain to a disentangled latent space, which can be interpreted as a nonlinear independent component analysis (ICA).
We treat the decoder (latent $\mapsto$ image) of the pre-trained GIN as a simulation program and only use the 20 dimensions with the highest variance of the latent space, which encode global (across all digits) and local (digit-specific) style features \citep[see][for a detailed analysis]{Sorrenson2020GIN}.
This introduces uncertainty, because we omit information from the GIN's learned bijective mapping.
We use NPE to compensate for the induced uncertainty by approximating the inverse problem, namely, performing probabilistic inference on the 20 latent style dimensions (inference targets $\thetab$) based on simulated images (observables $\x$).
The summary network is a simple CNN that learns a 64-dimensional representation of the observable image data $\x$.
The inference network is tasked with learning  the Bayesian posterior distribution $p(\thetab\given\x)$ of the latent styles $\thetab\in\mathbb{R}^{20}$ conditional on the observable images $\x\in\mathbb{R}^{784}$.
We realize the inference network as an affine coupling flow with 10 affine coupling layers, 512 units per layer, $10^{-4}$ kernel regularization, and dropout layers with a $0.15$ dropout rate.

In this experiment, we induce model misspecification via two strategies: First, we train the neural approximator on a narrow subset of possible styles by halving the dispersion of the style prior $p(\thetab)$.
During inference, out-of-distribution styles can easily be generated by sampling from a wider style prior.
Second, we only train the neural approximator on a subset of all ``domains'', namely the digits 0--4.
During inference, we achieve out-of-domain samples for the digits 5--8.

We further investigate two variations of this experiment.
In \textbf{Variation 1 (class-agnostic)}, the neural posterior approximator is trained and evaluated without any explicit label information (digits 0--9).
In \textbf{Variation 2 (class-informed)}, a simple 1D embedding of the digit index is passed as an additional condition to the invertible inference network.

\textit{Results.}
NPE exhibits excellent recovery of the latent style vectors of in-domain samples with our modified misspecification-aware optimization objective in Eq.~\ref{eq:bf_kl_mmd} (see \autoref{app:emnist-gin} for detailed recovery plots of all inferred style dimensions).
Moreover, NPE shows a heavily impaired performance (RMSE and bias to ground-truth) under extreme styles and domain shifts (see \autoref{tab:mms-emnist}).
Crucially, our proposed detection criterion successfully detects extreme style shifts (wider prior $p(\thetab)$) as well as domain shifts to unseen digits (5--9) via a significantly increased MMD score compared to the in-domain MMD score.
These effects persist regardless of whether the neural approximators are class-agnostic (variation 1) or class-informed (variation 2).
These results suggest that our proposed detection method is a reliable proxy to gauge the trustworthiness of NPE under a potentially misspecified probabilistic model and alert the user about unfaithful inference in case of unknown (variation 1) or known (variation 2) domain shifts.

\begin{table}[t]
\centering
\caption{\textbf{Experiment 5.} 
Our method consistently detects atypically extreme styles, as well as domain shifts to the unseen digits 5--9, as indexed by a large MMD score compared to the in-distribution baseline (first row).
The neural posterior approximator shows impaired performance in these unknown domains, as quantified by higher RMSE and bias relative to the ground-truth styles.
}\label{tab:mms-emnist}
\begin{tabular}{>{\bfseries}l | c c c | c c c}
\toprule
& \multicolumn{3}{c|}{\textbf{Variation 1: Class-agnostic}} & \multicolumn{3}{c}{\textbf{Variation 2: Class-informed}} \\
 & \textbf{MMD} & \textbf{RMSE} & \textbf{Bias} & \textbf{MMD} & \textbf{RMSE} & \textbf{Bias}  \\
\midrule
In distribution, digits 0--4 & 0.020 & 0.109 & 0.004 & 0.026 & 0.124 & 0.006\\
Extreme styles, digits 0--4 & 1.592 & 0.418 & 0.073 & 0.165 & 0.896 & 0.086 \\
New domain: Digit 5 & 4.031 & 0.860 & 0.565 & 2.576 & 1.050 & 0.774 \\
New domain: Digit 6 & 3.485 & 0.973 & 0.859 & 2.248 & 1.223 & 1.372 \\
New domain: Digit 7 & 3.786 & 0.993 & 0.915 & 2.033 & 1.021 & 0.968 \\
New domain: Digit 8 & 3.873 & 0.844 & 0.549 & 2.393 & 0.917 & 0.573 \\
New domain: Digit 9 & 5.702 & 0.745 & 0.506 & 2.470 & 0.693 & 0.440 \\
\bottomrule
\end{tabular}
\end{table}

\section{Conclusions}\label{sec:discussion}

This paper approached a fundamental problem in amortized simulation-based Bayesian inference, namely, flagging potential posterior errors due to model misspecification.
We argued that misspecified models might cause so-called \emph{simulation gaps}, resulting in deviations between simulations during training time and actual observed data at test time.
We further showed that simulation gaps can be detrimental for the performance and faithfulness of simulation-based inference relying on neural networks.
We proposed to increase the networks' awareness of posterior errors by compressing simulations into a structured latent summary space induced by a modified optimization objective in an unsupervised fashion. 
We then applied the maximum mean discrepancy (MMD) estimator, equipped with a sampling-based hypothesis test, as a criterion to spotlight discrepancies between model-implied and actually observed distributions in summary space.
While we focused on the application to NPE \citep[\textit{BayesFlow} implementation;][]{bayesflow,Radev2023bayesflow} and SNPE \citep[sbi implementation;][]{tejero2020sbi}, our proposed method can be easily integrated into other inference algorithms and frameworks as well.
Most notably, while we focus on \textit{detection}, our reliable misspecification detection method can be easily combined with techniques that aim to \textit{mitigate} model misspecification.
Our method can pinpoint individual data sets where inference is potentially unfaithful, and local non-amortized techniques may be used as a targeted post-hoc correction for these problematic data sets.
Such a two-step approach might consist of (i) detection with our method; and (ii) targeted mitigation with local methods, and constitute a promising alternative to current global regularization techniques.
Our methods are implemented in the open-source BayesFlow library (\href{http://www.bayesflow.org}{www.bayesflow.org}) and can be seamlessly integrated into an end-to-end workflow for amortized simulation-based inference.

\subsection*{Acknowledgments}
We acknowledge support by Cyber Valley CyVy-RF-2021-16 (MS, PCB), Deutsche Forschungsgemeinschaft EXC-2075 - 390740016 (MS, PCB), the Informatics for Life initiative (UK), and Deutsche Forschungsgemeinschaft EXC-2181 - 390900948 (STR).

\subsection*{Data Availability}
Our methods are implemented in the free open-source library BayesFlow (\href{http://www.bayesflow.org}{www.bayesflow.org}) and can be seamlessly integrated into amortized Bayesian workflows.

\bibliography{references}
\bibliographystyle{tmlr}

\clearpage
\begin{appendices}

\begin{center}
    \Large\bfseries Appendix
\end{center}

\counterwithin{figure}{section}

\section{Theoretical Implications}
\label{sec:theoretical-implications}

Attaining the global minimum of the neural posterior estimation loss with an arbitrarily expressive neural architecture $\{h_{\psib^*}, f_{\phib^*}, \mathcal{M}\}$ implies that (i) the inference and summary network jointly amortize the analytic posterior $p(\thetab \given \x, \mathcal{M})$; and (ii) the summary network transforms the data $p(\x \given \mathcal{M})$ into a unit Gaussian in summary space: $p\big(\z = h_{\psib^*}(\x)\big) = \mathcal{N}(\z \given \mathbf{0}, \mathbb{I})$.
According to (i), the set of inference network parameters $\phib^*$ is a minimizer of the (expected) negative log posterior learned by the inference network,
\begin{equation}
    \phib^* = \argmin_{\phib}  
    \mathbb{E}_{p(\z)} \mathbb{E}_{p(\thetab \given \z)}\Big[-\log q_{\phib}\big(\thetab\given \z\big)\Big].
\end{equation}
At the same time, (ii) ensures that deviances in the summary space (according to MMD) imply differences in the data generating processes,
\begin{equation}\label{eq:implications:mmd-greater-zero}
    \underbrace{\mathbb{MMD}^2\big[p^*(\z)\,||\,p(\z\given\M)\big] > 0}_{\text{summary space difference}} \implies \underbrace{\mathbb{MMD}^2\big[p^*(\x)\,||\,p(\x \given \mathcal{M})\big] > 0}_{\text{data space difference}},
\end{equation}
since a deviation of $p^*(\z=h_{\psib^*}(\x))$ from a unit Gaussian means that the summary network is no longer transforming samples from $p(\x \given \mathcal{M})$. 

Accordingly, the LHS of Eq.~\ref{eq:implications:mmd-greater-zero} no longer guarantees that the inference network parameters $\phib^*$ are maximally informative for posterior inference. 
The preceding argumentation also motivates our augmented objective, since a divergence of summary statistics for observed data $\observed{\z}=h_{\psib}(\observed{\x})$ from a unit Gaussian signalizes a deficiency in the assumed generative model $\M$ and a need to revise the latter. 
We also hypothesize and show empirically that we can successfully detect simulation gaps in practice even when the summary network outputs have not exactly converged to a unit Gaussian (e.g., in the presence of correlations in summary space, cf.\ \textbf{Experiment \numberDDM} and \textbf{Experiment \numberCovid}).

However, the converse of Eq.~\ref{eq:implications:mmd-greater-zero} is not true in general.
In other words, a discrepancy in data space (non-zero MMD on the RHS of Eq.~\ref{eq:implications:mmd-greater-zero}) does \emph{not} generally imply a difference in summary space (non-zero MMD on the LHS of Eq.~\ref{eq:implications:mmd-greater-zero}).
To illustrate this via a counter-example, consider the assumed Gaussian generative model $\M$ defined by
\begin{equation}\label{eq:implications:process-1}
    \begin{aligned}
        \mu &\sim p(\mu), \\
        x_1,x_2  &\sim \mathcal{N}(\mu, \sigma^2 = 2),
    \end{aligned}
\end{equation}
for $N = 2$ observations and a summary network with a single-output ($S = 1$).
Since the variance is fixed, the only inference target is the mean $\mu$.

Then, an optimal summary network $\psib^*$ outputs the minimal sufficient summary statistic for recovering the mean, namely the empirical average: $h_{\psib^*}(x_1, x_2) = \bar{x} \equiv (x_1 + x_2) / 2$. 
Consequently, the distribution in summary space is given as $p(\bar{x}) = \mathcal{N}(0, 1)$.\footnote{This follows from the property $\text{Var}(\bar{x}) = \text{Var}((x_1 + x_2)/ 2) = (\text{Var}(x_1) + \text{Var}(x_2)) / 2^2 = 1$ for independent $x_1, x_2$.}
In terms of the MMD criterion, we see that 
\begin{equation}
    \mathbb{MMD}^2\big[\underbrace{p(h_{\psib^*}(x_1, x_2))}_{=\mathcal{N}(0, 1)}\,||\,\mathcal{N}(0, 1)\big] = 0.
\end{equation}

Now, suppose that the real data are actually generated by a different model $\M^*$ with $\observed{x}\sim p^*(\x)$ given by 
\begin{equation}\label{eq:implications:process-2}
    \begin{aligned}
        \mu &\sim p(\mu),\\
        \observed{x}_1 &\sim \mathcal{N}(\mu, \sigma^2 = 1), \\
        \observed{x}_2 &\sim \mathcal{N}(\mu, \sigma^2 = 3).
    \end{aligned}
\end{equation}
Clearly, this process $p^*(\x)$ differs from $p(\x\given\M)$ (Eq.~\ref{eq:implications:process-1}) on the data domain according to the MMD metric: %
$\mathbb{MMD}^2\big[p^*(\x)\,||\,p(\x \given \mathcal{M})\big] > 0$. 
However, using the same calculations as above, we find that the summary space for the process $p^*(\x)$ also follows a unit Normal distribution: $p^*(\bar{x}) = \mathcal{N}(0, 1)$.
Thus, the processes $p^*(\x)$ and $p(\x\given\M)$ are indistinguishable in the summary space despite the fact that the first generative model $\M$ is clearly misspecified.

The above example shows that learning \emph{minimal sufficient summary statistics} for solving the inference task (i.e., the mean in this example) might not be optimal for detecting simulation gaps. 
In fact, increasing the output dimensions $S$ of the summary network $h_{\psib}$ would enable the network to learn structurally richer (overcomplete) sufficient summary statistics.
The latter would be invariant to fewer misspecifications and thus more useful for uncovering simulation gaps. 
In the above example, an \emph{overcomplete} summary network with $S = 2$ which simply copies and scales the two variables by their corresponding variances is able to detect the misspecification. 
\textbf{Experiment \numberGaussianMeans} studies the influence of the number of summary statistics in a controlled setting and provides empirical illustrations.
\textbf{Experiments \numberDDM} and \textbf{\numberCovid} further address the choice of the number of summary statistics in more complex models of decision making and disease outbreak.
Next, we describe how to practically detect simulation gaps during inference using only \emph{finite realizations} from $\M$ and $\mathcal{M}^*$.

\section{Posterior Inference Errors}\label{sec:posterior-inference-errors}

Given a generative model $\mathcal{M}$, the \emph{analytic posterior under the potentially misspecified model} $p(\thetab \given \x,\mathcal{M})$ always exists, even if $\mathcal{M}$ is misspecified for the data $\x$.
Obtaining a trustworthy approximation of the analytic posterior is the fundamental basis for any follow-up inference (e.g., parameter estimation or model comparison) and must be at least an intermediate goal in real world applications.
Assuming optimal convergence under a misspecified model $\mathcal{M}$, the amortized posterior $q_{\phib}\big(\thetab \given \z = h_{\psib}(\x), \mathcal{M}\big)$ still corresponds to the analytic posterior $p(\thetab \given \x,\mathcal{M})$, as any transformed $\observed{\x}$ arising from $p^*(\x)$ has non-zero density in the latent Gaussian summary space.\footnote{We assume that we have no hard limits in the prior or simulator in $\M$.}
Thus, the posterior approximator should still be able to obtain the correct pushforward density under $\mathcal{M}$ for any query $\observed{\x}$.
However, optimal convergence can never be achieved after finite training time, so we need to address its implications for the validity of amortized simulation-based posterior inference in practice.

Given finite training data, the summary and inference networks will mostly see simulations from the \emph{typical set} $\mathcal{T}(\M) \subset \mathcal{X}$ of the generative model $\M$, that is, training instances whose self-information $-\log p(\x \given \mathcal{M})$ is close to the entropy $\mathbb{E}\big[-\log p(\x \given \mathcal{M})\big]$.
In high dimensional problems, the typical set will comprise a rather small subset of the possible outcome space, 
determined by a complex interaction between the components of $\M$ \citep{betancourt2017}.
Accordingly, good convergence in practice may mean that i) only observations from $\mathcal{T}(\M)$ actually follow the approximate Gaussian in latent summary space and ii) the inference network has only seen enough training examples in $\mathcal{T}(\M)$ to learn accurate posteriors for observables $\x \in \mathcal{T}(\M)$, but remains inaccurate for well-specified, but rare $\x\not\in\mathcal{T}(\M)$.

Since atypical or improbable outcomes occur rarely during simulation-based training, they have negligible influence on the neural networks' loss.
Consequently, posterior approximation errors for observations outside of $\mathcal{T}(\M)$ can be large, simply because the networks have not yet converged in these unusual regions, and the highly non-linear mapping of the inference network still deviates considerably from the true solution.
Better training methods might resolve this problem in the future, but for now our proposed MMD criterion reliably signals low fidelity posterior estimates by quantifying the ``distance from the typical generative set'' $\mathcal{T}(\M)$ in the structured summary space.

Moreover, we hypothesize and demonstrate empirically in the following experiments that the difference between the correct posterior $p(\thetab \given \x, \mathcal{M})$ and the approximate posterior $q_{\phib}(\thetab \given h_{\psib}(\x), \mathcal{M})$ for misspecified models increases as a function of MMD, and thus the latter also measures the amount of misspecification.
Therefore, our MMD criterion serves a dual purpose in practice: It can uncover potential simulation gaps and, simultaneously, signal errors in posterior estimation of rare (but valid) events.

\section{Description of the Method as Algorithm}

\begin{algorithm}[H]
\caption{Misspecification-aware amortized Bayesian inference. The algorithm illustrates online learning for simplicity. The training phase can utilize any other learning paradigm as well (e.g., round-based). Thus, details like batch size or round indices are omitted in the training phase of the algorithm.}
\label{alg:mms}
\begin{algorithmic}[1]
\State \textbf{Training phase}:
\Repeat
\State Sample parameters and data $(\thetab, \x)$ from the specified generative model $\M$.
\State Pass the data $\x$ through the summary network: $\z = h_{\psib}(\x)$.
\State Compute $-\log q_{\phib}\big(\thetab\given \z, \mathcal{M}\big)$.
\State Compute loss function.
\State Update neural network parameters $\psib, \phib$  via backpropagation.
\Until{convergence to $\widehat{\psib}, \widehat{\phib}$}
\\
\State \textbf{Inference phase} \textit{(given $N$ observed or test data sets $\{\observed{\x}^{(n)}\}$, query $L$ draws from the amortized posterior, use $M$ draws from the validation summary distribution)}:
\For{$m=1,\ldots, M$}
\State Re-use data set $\x^{(m)}$ from the training phase or simulate a new one from $\mathcal{M}$.
\State Pass the data set $\x^{(m)}$ through the converged summary network: $\z^{(m)} = h_{\widehat{\psib}}(\x^{(m)})$.
\EndFor
\For{$n=1,\ldots, N$}
\State Pass the observed data set $\observed{\x}^{(n)}$ through the summary network: $\observed{\z}^{(n)} = h_{\widehat{\psib}}(\observed{\x}^{(n)})$.
\State Pass $\observed{\z}^{(n)}$ through the posterior approximator for $L$ draws $\{\thetab^{(n, l)}\}$ from $q_{\widehat{\phib}}(\thetab \given \observed{\z}^{(n)})$
\EndFor
\State Estimate the MMD distance of the inference data $\{\observed{\z}^{(n)}\}$ from the validation summary space under the generative model $\M$ from training: $\widehat{\mathrm{MMD}}^2\left(\left\{\observed{\z}^{(n)} \right\} \,\middle|\middle|\,\left\{\z^{(m)} \right\}\right)$.
\State \Return Draws from the approximate posterior distribution for each queried data set $\observed{\x}^{(n)}$ $\{\thetab^{(n, l)}\}$ and $\widehat{\mathrm{MMD}}^2\left(\left\{\observed{\z}^{(n)} \right\} \,\middle|\middle|\,\left\{\z^{(m)} \right\}\right)$.
\end{algorithmic}
\end{algorithm}

\clearpage
\section{Additional Experiments}

\subsection{Experiment \numberGaussianMeansCov: 5D Gaussian Means and Covariance}
\begin{figure}[t]
    \begin{minipage}{.48\linewidth}
        \begin{subfigure}[t]{\linewidth}
            \includegraphics[width=\linewidth]{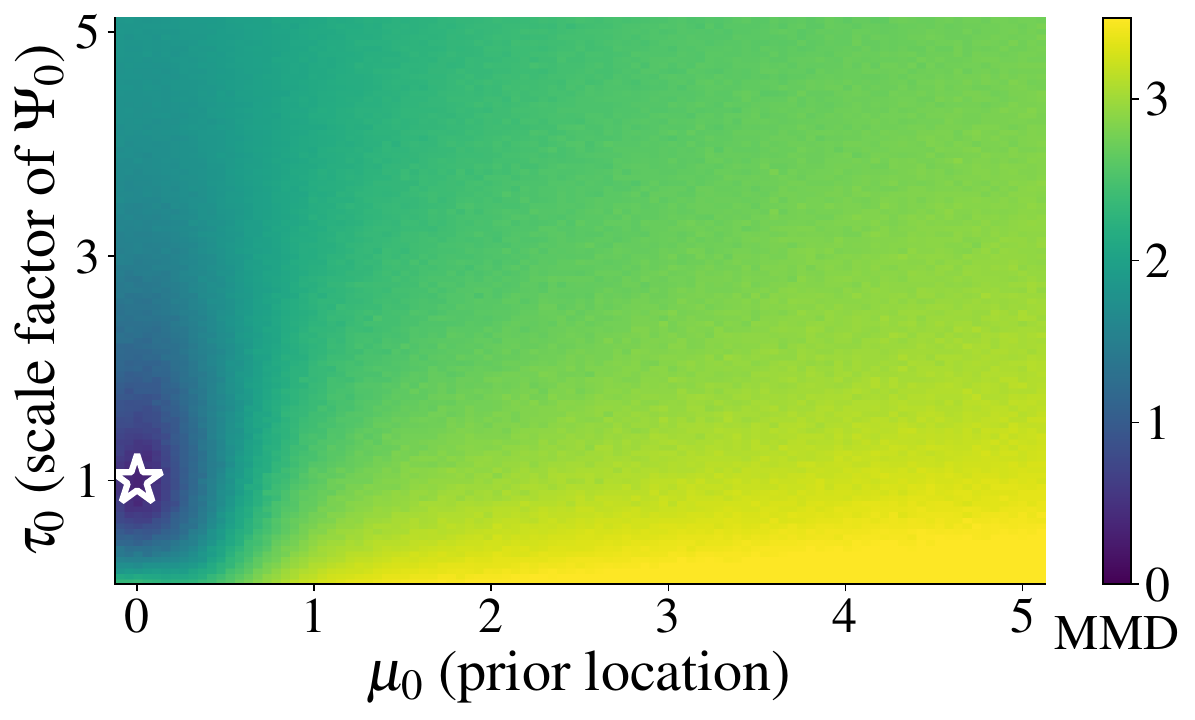}
            \caption{Prior misspecification.}
            \label{fig:app:mvn-full:MMD:prior}
        \end{subfigure}
    \end{minipage}
    \hfill
    \begin{minipage}{.48\linewidth}
        \begin{subfigure}[t]{\linewidth}
            \includegraphics[width=\linewidth]{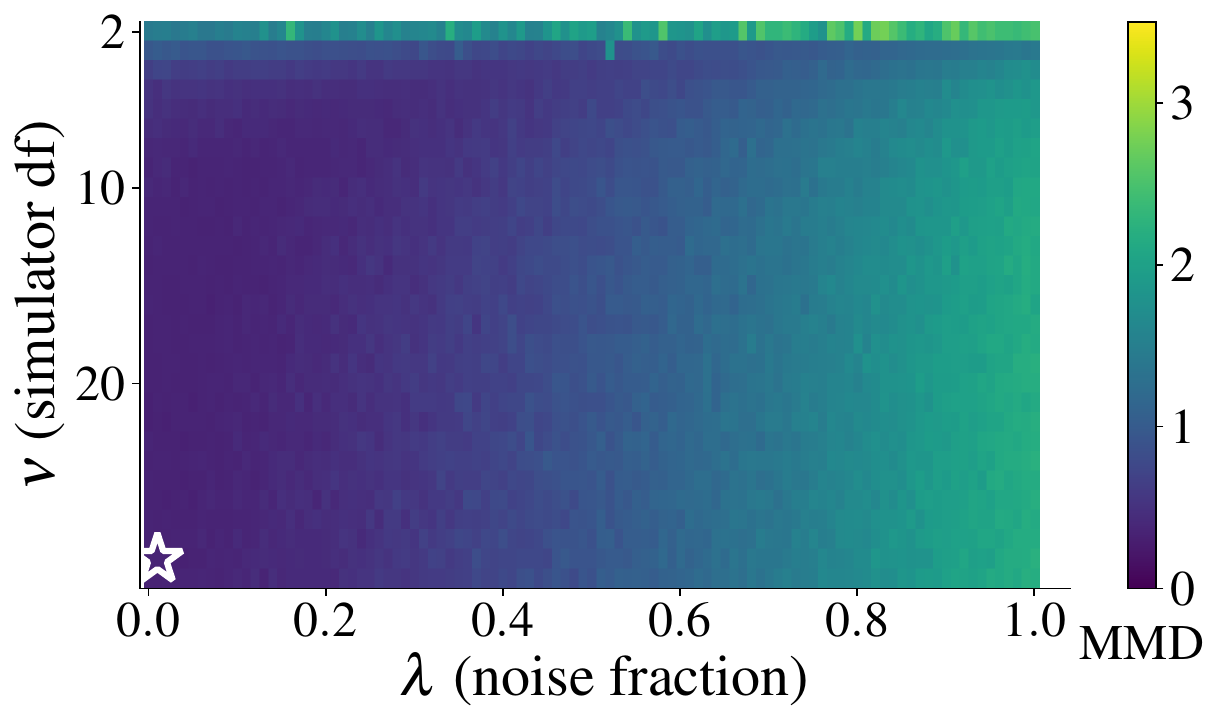}
            \caption{Simulator and noise misspecification.}
            \label{fig:app:mvn-full:MMD:likelihood-noise}
        \end{subfigure}
    \end{minipage}
    \hspace*{1cm}
\caption{\textbf{Experiment \numberGaussianMeansCov.} MMD as a function of model misspecification severity in the prior (left) as well as simulator and noise (right). 
All induced model misspecifications are detectable through an increased MMD.
White stars represent the configuration for the well-specified model (i.e., training model $\M$).
The deviating pattern in the top-most row of (\subref{fig:app:mvn-full:MMD:likelihood-noise}) is caused by the infinite variance of the Student $t$ likelihood with $\nu=2$ degrees of freedom: This is an extreme simulation gap with respect to the unit Gaussian model~$\M$, and consequently detected as such.}
\label{fig:app:mvn-full:MMD}
\end{figure}
This experiments extends the Gaussian conjugate model to higher dimensions and a more difficult task, i.e., recovering the means and the full covariance matrix of a $5$-dimensional Gaussian.
There is a total of $20$ inference parameters---5 means and 15 (co-)variances---meaning that $20$ summary statistics would suffice to solve the inference task.
The mean vector $\mub$ and the covariance matrix $\Sigmab$ are drawn from a joint prior, namely a normal-inverse-Wishart distribution \citep[$\NIW$;][]{Barnard2000}.
The normal-inverse-Wishart prior $\NIW(\mub, \Sigmab\given\mub_0, \lambda_0, \Psib_0, \nu_0)$ implies a hierarchical process, and the implementation details are described in Section~\ref{sec:app:mvn-full} in the Appendix.
We set the number of summary statistics to $S=40$ to balance the trade-off between posterior error and misspecification detection (see \textbf{Experiment \numberGaussianMeans}).
The model $\mathcal{M}$ used for training the networks as well as the induced model misspecifications (prior, simulator, and noise) are detailed in the Appendix.

\textit{Results.} The converged posterior approximator can successfully recover the analytic posterior for all inference parameters when no model misspecification is present.
Thus, our method does not impede posterior inference when the models are well-specified.
Since the summary space comprises $S=40$ dimensions, visual inspection is no longer feasible, and we resort to the proposed MMD criterion.
Both induced prior misspecifications---i.e., location and variance---are detectable through an increased MMD (see \autoref{fig:app:mvn-full:MMD:prior}).
Model misspecifications via a heavy-tailed simulator---i.e., Student-$t$ with $\nu=2$ degrees of freedom---, as well as Beta noise, are also detectable with our MMD criterion (see \autoref{fig:app:mvn-full:MMD:likelihood-noise}).

\section{Implementation Details}

\subsection{5D Gaussian with Covariance (Experiment \numberGaussianMeansCov)}

\label{sec:app:mvn-full}
The normal-inverse-Wishart prior $\NIW(\mub, \Sigmab\given\mub_0, \lambda_0, \Psib_0, \nu_0)$ implies a hierarchical prior. 
Suppose the covariance matrix $\Sigmab$ has an inverse Wishart distribution $\mathcal{W}^{-1}(\Sigmab\given\Psib_0, \nu_0)$  and the mean vector $\mub$ has a multivariate normal distribution $\mathcal{N}(\mub\given\mub_0, \frac{1}{\lambda_0}\Sigmab)$, then the tuple $(\mub, \Sigmab)$ has a normal-inverse-Wishart distribution $\NIW(\mub, \Sigmab\given\mub_0, \lambda_0, \Psib_0, \nu_0)$.
Finally, the likelihood is Gaussian:$\x_k\sim\mathcal{N}(\mub, \Sigmab)\;\text{for}\; k=1,\ldots, K$.

For a multivariate Gaussian with unknown mean and unknown covariance matrix, the analytic joint posterior $p(\mub_p, \Sigmab_p\given\x)$ follows a normal-inverse Wishart distribution again:
\begin{equation}
\begin{aligned}
    (\mub_p, \Sigmab_p \given \x) & \sim\text{N-}\mathcal{W}^{-1}(\mub_p, \Sigmab_p\given\mub_K, \lambda_K, \Psib_K, \nu_K)\quad\text{with}\\
    \mub_K & = \dfrac{\lambda_0\mub_0 + K\bar{\x}}{\lambda_0+K},\quad 
    \lambda_K =\lambda_0+K,\quad
    \nu_K  = \nu_0+K,\\
    \Psib_K & = \Psib_0+\sum\limits_{k=1}^K(\x_k-\bar{\x})(\x_k-\bar{\x})^T+\dfrac{\lambda_0 K}{\lambda_0 + K}(\bar{\x}-\mub_0)(\bar{\x}-\mub_0)^T
\end{aligned}
\end{equation}
The marginal posteriors for $\mub_p$ and $\Sigmab_p$ then follow as \citep{Murphy2007}:
\begin{equation}\label{eq:mvn-full-cov:analytic-posterior:marginal}
\begin{aligned}
    \mub_p & \sim t_{\nu_K-D-1}\Big(\mub_p\,\Big|\,\mub_K, \frac{\Psib_K^{-1}}{\lambda_K(\nu_K-D+1)}\Big)\\
    \Sigmab_p & \sim \mathcal{W}^{-1}(\Sigmab_p\given\Psib_K, \nu_K)
\end{aligned}
\end{equation}

The model $\mathcal{M}$ used for training the networks as well as the types of induced model misspecifications are outlined in \autoref{tab:app:mvn-full-MMS}.
\begin{table*}[b]
    \centering
    \scriptsize
    \begin{tabular}{l|ll}
        \bfseries Model &\bfseries Prior &\bfseries Likelihood\\
        \hline
        $\mathcal{M}$ (No MMS) &
        $\big(\mub, \Sigmab\big)\sim\NIW(\0, 5, \mathbb{I}, 10)$ &
        $\x_k\sim\mathcal{N}(\mub, \Sigmab)$
        \\
        $\mathcal{M}_P$ (Prior) &
        $\big(\mub, \Sigmab\big)\sim\NIW(\mub_0, 5, \tau_0\mathbb{I},10)$&
        $\x_k\sim\mathcal{N}(\mub, \Sigmab)$\\
        $\mathcal{M}_S$ (Simulator) &
        $\big(\mub, \Sigmab\big)\sim\NIW(\0, 5, \mathbb{I},10)$&
        $\x_k\sim t_{\text{df}}(\mub, \Sigmab),\quad\text{df}\in\mathbb{N}_{>0}$
        \\
        $\mathcal{M}_N$ (Noise)&
        $\big(\mub, \Sigmab\big)\sim\NIW(\0, 5, \mathbb{I},10)$& 
        $\x_k\sim\lambda\cdot\mathrm{Beta}(2, 5)+(1-\lambda)\cdot\mathcal{N}(\mub, \Sigmab)$
    \end{tabular}
        \caption{Investigated model misspecifications (MMS) for the $5$-dimensional Gaussian with fully estimated covariance matrix. $\NIW(\mub_0, \lambda_0, \Psib,\nu)$ denotes a multivariate Normal-Inverse-Wishart distribution with location $\mub_0$, precision scaling factor $\lambda_0$, precision $\Psib$, and $\nu$ degrees of freedom.
    }
    \label{tab:app:mvn-full-MMS}
\end{table*}
In the evaluation, we compare the means of the approximate posterior samples with the first moment of the respective marginal analytic posterior from Eq.~\ref{eq:mvn-full-cov:analytic-posterior:marginal}. 
We evaluate correlation matrices with standard deviations on the diagonal. 
For the $t$ distributed posterior mean and inverse-Wishart distributed posterior covariance, we obtain \citep{Mardia1979}:
\begin{equation}
        \mathbb{E}(\mub_p) =\mub_K,\quad
        \mathbb{E}(\Sigmab_p) =\frac{\Psib_K}{\nu_K-D-1}
\end{equation}

\subsection{Cancer and Stromal Cells (Experiment \numberCS)}\label{app:cs}

As implemented by \citet{ward_robust_2022}, the CS model simulates the development of cancer and stromal cells.
The respective cell counts (total $N_c$, unobserved parents $N_p$, daughters for each parent $N_d^{(i)}$) are stochastically defined through Poisson distributions
\begin{equation}
        N_c \sim \mathrm{Poisson}(\lambda_c),\quad N_p \sim \mathrm{Poisson}(\lambda_c),\quad N_d^{(i)} \sim \mathrm{Poisson}(\lambda_c),\quad i=1,\ldots, N_p.
\end{equation}

The distance based metrics for the hand-crafted summary statistics 3--4 are empirically approximated from 50 stromal cells to avoid computing the full distance matrix \citep{ward_robust_2022}.
The prior distributions are specified as 
\begin{equation}
        \lambda_c \sim \mathrm{Gamma}(25, 0.03),\quad
        \lambda_d \sim \mathrm{Gamma}(5, 0.5),\quad
        \lambda_p \sim \mathrm{Gamma}(45, 3)
\end{equation}
where $\mathrm{Gamma}(a, b)$ denotes the Gamma distribution with location $a$ and rate $b$.
To induce model misspecification through necrosis, a Bernoulli variable $w_i\sim\mathrm{Bernoulli}(\pi)$ is sampled, removing cancer cells within a specified radius around the parent cell \citep{ward_robust_2022}.
Thus, the Bernoulli parameter $\pi$ controls the degree of misspecification, ranging from no misspecification ($\pi=0$; no necrosis) to maximal misspecification with respect to that parameter ($\pi=1$; necrosis of all susceptible cells).

\subsection{DDM (Experiment \numberDDM)}\label{sec:app:ddm}
The starting point of the evidence accumulation process is unbiased, $x_{t=0}=\frac{a}{2}$.
During training, all parameters are drawn from Gamma prior distributions:
\begin{equation}
    v_1, v_2, a_1, a_2, t_0 \sim\Gamma(5, 0.5).
\end{equation}
We first generate uncontaminated data $\x^*$ from the well-specified generative model $\M$. 
Second, we randomly choose a fraction $\lambda\in[0,1]$ of the data $\x^*$.
Third, we replace this fraction with data-dependent contaminants $\xi$,
\begin{equation}
    \begin{aligned}
    \text{Fast guesses:}\quad\xi & \sim\mathcal{U}\big(0.1, Q_{10}(\x^*)\big)\\
    \text{Slow responses:}\quad\xi & \sim\mathcal{U}\big(Q_{75}(\x^*), 10\big),
    \end{aligned}
\end{equation}
where $Q_k(\x^*)$ denotes the $k^{\text{th}}$ percentile of $\x^*$.
The asymmetry in percentiles between fast and slow responses arises from the inherent positive skewness of reaction time distributions. 
The fixed upper limit of slow response contamination is motivated by the maximum number of iterations of the utilized diffusion model simulator. 
The contamination procedure is executed separately for each condition and response type. 
If an experiment features both fast and slow contamination, the fraction $\lambda$ is equally split between fast and slow contamination. 
The uncontaminated data set is generated once and acts as a baseline for all analyses of an experiment, resulting in a baseline MMD of 0 since $\x^*$ is unaltered if $\lambda=0$.

\section{Bootstrapping Procedure}
\label{sec:app:bootstrap}
In \textbf{Experiment \numberCovid}, we estimate a sampling distribution of the MMD between samples from the specified training model $\M$ with $\x\sim p(\x\given\M^*)$ and samples from the (opaque) observed model $\M^*$ with $\observed{x}\sim p^*(\x)$.
Since simulating time series from the compartmental models is time-consuming, we opt for bootstrapping \citep{Stine1989} on $M=1\,000$ pre-simulated time series $\{\x^{(m)}\}$ from $\mathcal{M}$ and $N=1\,000$ pre-simulated time series $\{\observed{\x}^{(n)}\}$ from $\M^*$.
In each bootstrapping iteration, we draw $M=1\,000$ samples with replacement from $\{\x^{(m)}\}$ as well as $N_B\in\{1,2,5\}$ samples (with replacement) from $\{\observed{\x}^{(n)}\}$ and calculate the MMD between the sets of bootstrap samples.

\section{2D Gaussian Means: Further Results}

\subsection{Performance under well-specified model}

\begin{figure}[H]
    \centering
    \begin{subfigure}[t]{.49\textwidth}
     \includegraphics[width=\linewidth]{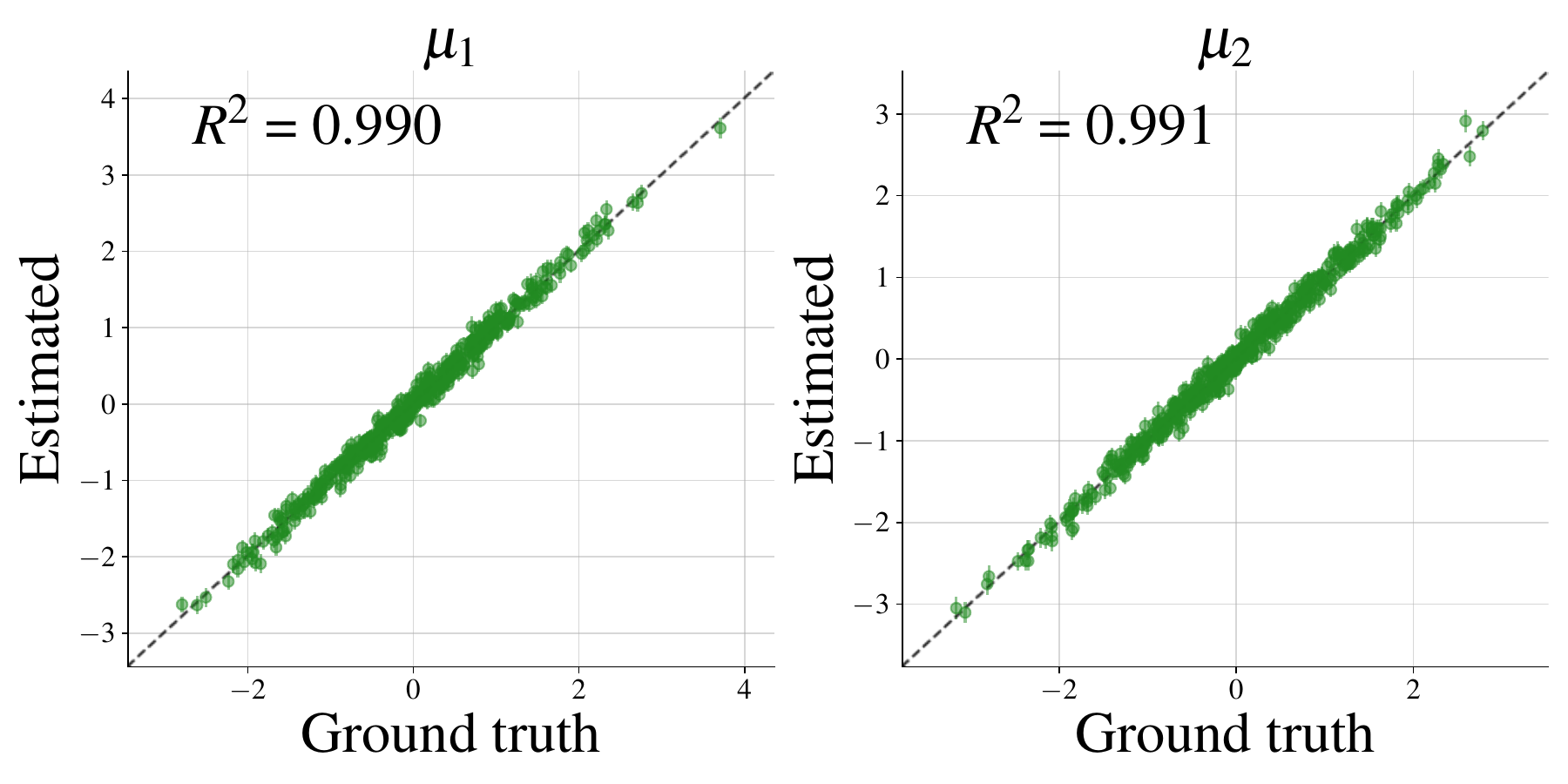}
     \caption{Excellent parameter recovery.}
     \label{fig:mvn:well-specified-performance:recovery}
    \end{subfigure}
    \hfill
    \begin{subfigure}[t]{.49\textwidth}
     \includegraphics[width=\linewidth]{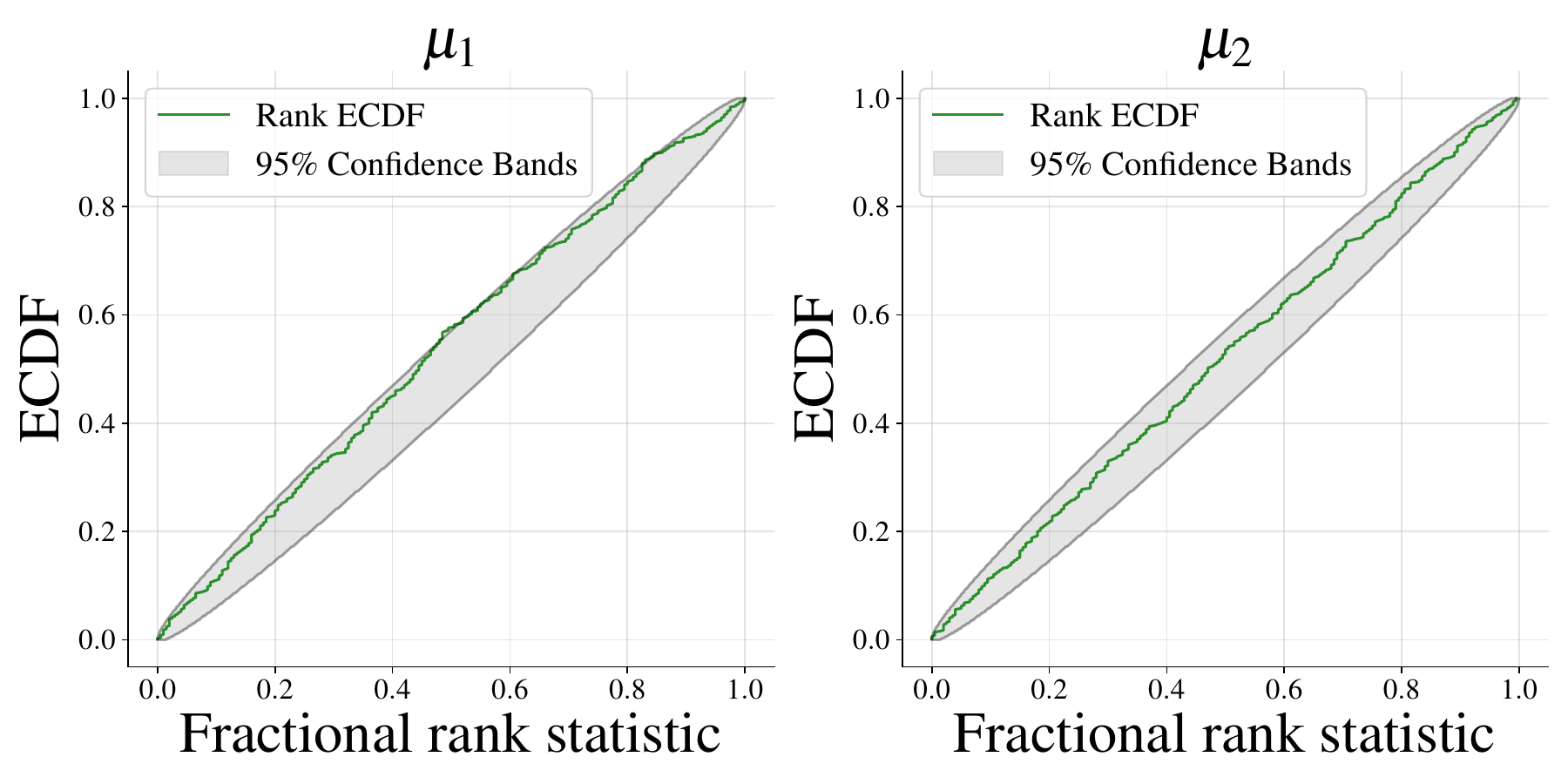}
     \caption{Excellent calibration.}
     \label{fig:mvn:well-specified-performance:calibration}
    \end{subfigure}
    \caption{\textbf{Experiment \numberGaussianMeans}. When the assumed model is well-specified for the observed data ($\M=\M^*$), both posterior recovery (\subref{fig:mvn:well-specified-performance:recovery}) and calibration (\subref{fig:mvn:well-specified-performance:calibration}) remain excellent with our adjusted optimization objective.}
    \label{fig:mvn:well-specified-performance}
\end{figure}

\subsection{Overcomplete Summary Statistics}
\begin{figure}[t]
    \centering
    \includegraphics[width=.46\linewidth]{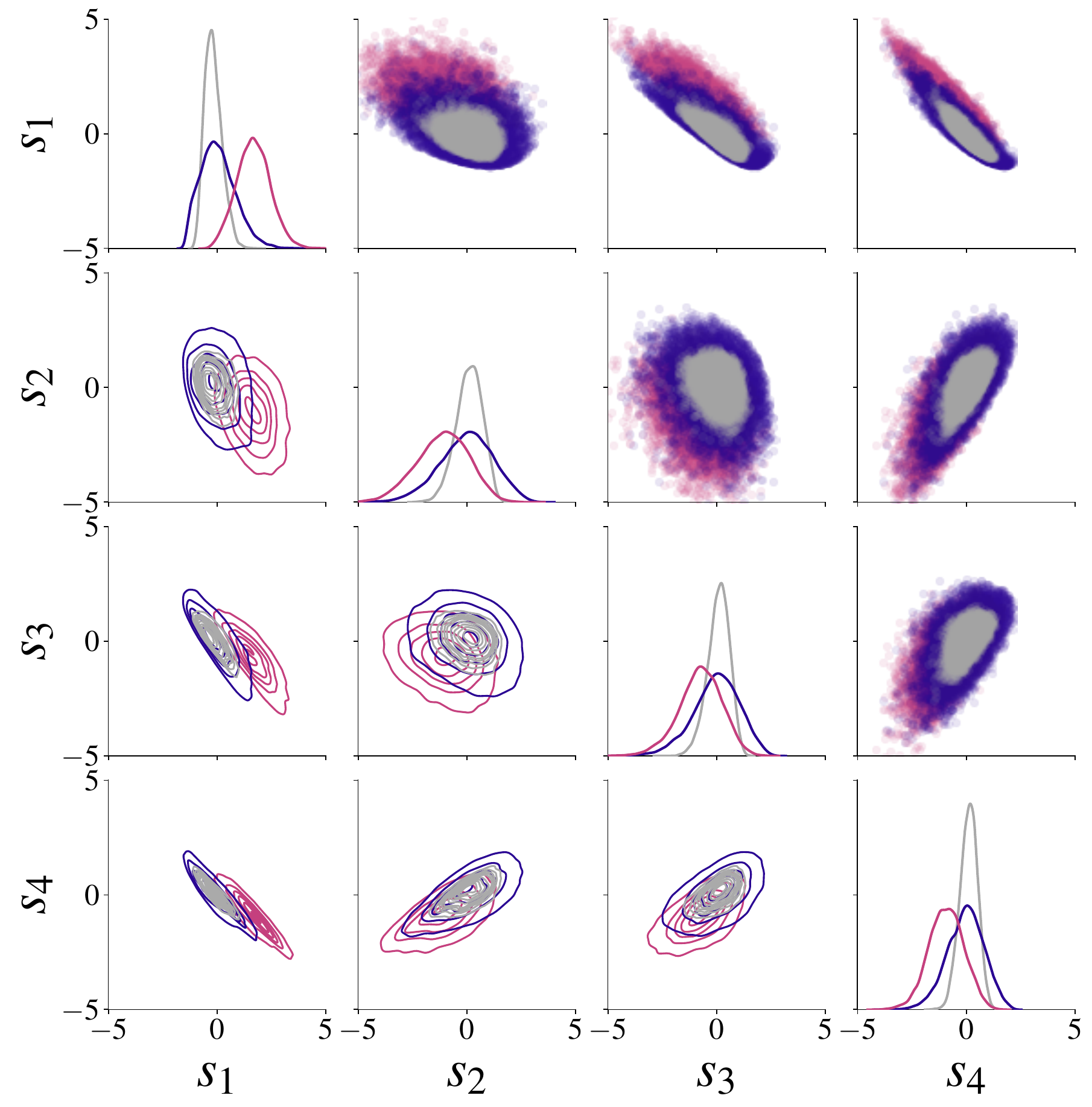}\\
    \includegraphics[width=0.7\linewidth]{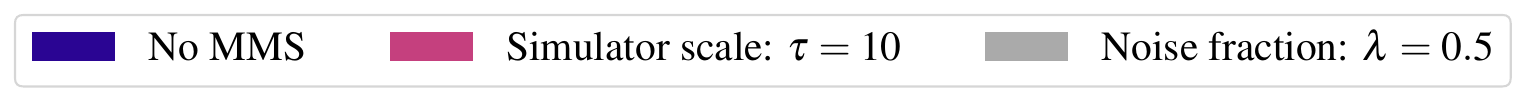}
    \caption{Pairplot of $10\,000$ latent summary space samples from the overcomplete summary network. Both noise (orange) and simulator (pink) misspecifications are distinguishable from the typical latent generative space (blue).}
    \label{fig:app:mvn-means:overcomplete:pairplot}
\end{figure}

\autoref{fig:app:mvn-means:overcomplete:pairplot} shows the latent summary space when overcomplete sufficient summary statistics ($S=4$) are used in \textbf{Experiment \numberGaussianMeans} to recover the means of a $2$-dimensional Gaussian.
Model misspecification with respect to both simulator and noise is detectable through anomalies in the latent summary space.
A network with $S=2$ summary statistics and otherwise equivalent architecture could not capture these types of model misspecification. 

\clearpage
\section{Replication of Experiment \numberGaussianMeans~with SNPE-C}
\label{app:snpe}

Here, we show the results of repeating \textbf{Experiment \numberGaussianMeans} with SNPE-C \citep{apt} in the sbi implementation \citep{tejero2020sbi} for posterior inference, instead of NPE in the BayesFlow implementation \citep{bayesflow}.
The results are largely equivalent to those obtained with NPE in BayesFlow.

\begin{figure}[H]
    \centering
    \begin{subfigure}[t]{0.40\linewidth}
        \includegraphics[width=\linewidth]{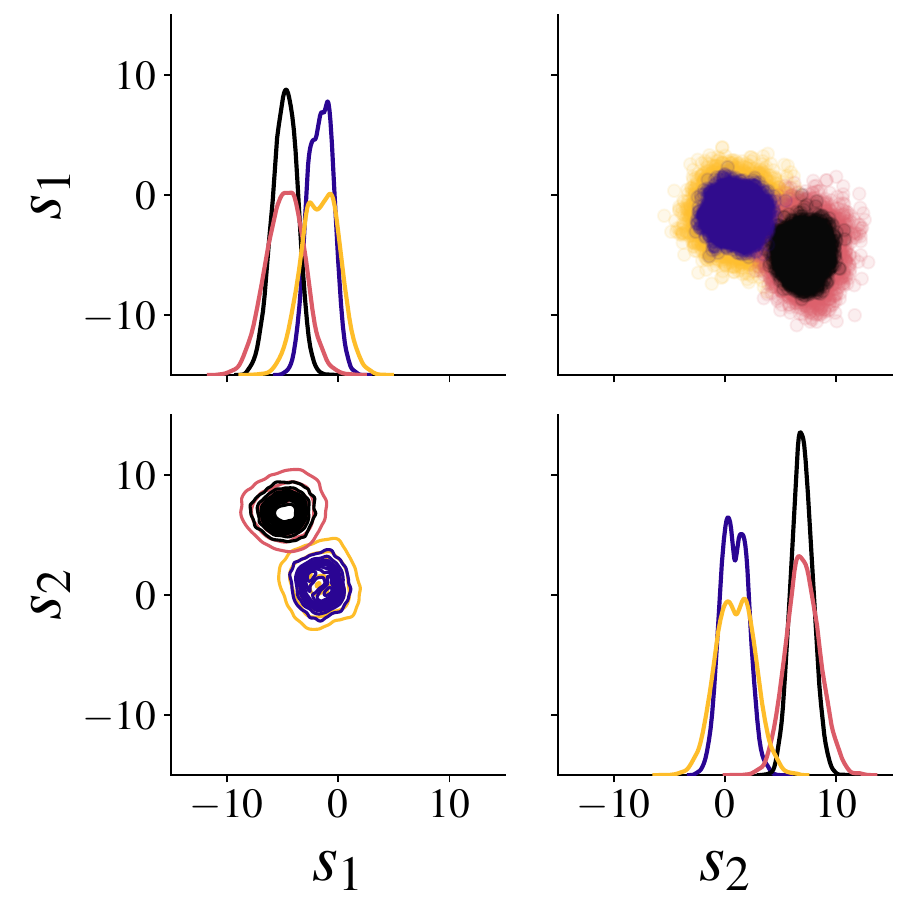}
    \end{subfigure}\hspace*{1cm}
    \begin{subfigure}[t]{0.40\linewidth}
        \includegraphics[width=\linewidth]{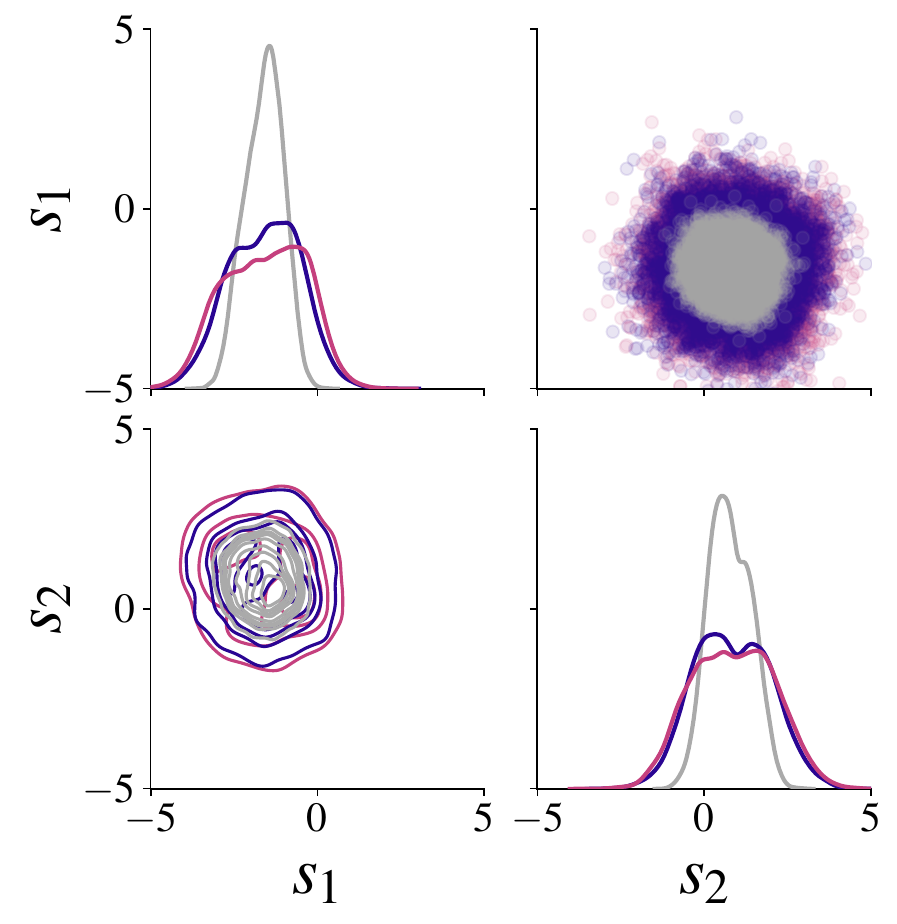}
    \end{subfigure}\\
    \includegraphics[width=0.80\linewidth]{abf_mvn_means_sufficient_pairplot_MMD_legend_new.pdf}
    \caption{\textbf{Experiment \numberGaussianMeans, SNPE-C.} Summary space samples for the minimal sufficient summary network ($S=2$) from a well-specified model $\M$ (blue) and misspecified configurations. \textbf{Left:} Prior misspecification can be detected. \textbf{Right:} Simulator scale misspecification is indistinguishable from the validation summary statistics.}
\end{figure}

\begin{figure}[H]
    \centering
    \begin{subfigure}[c]{0.7\linewidth}%
    \setlength\tabcolsep{2pt}%
    \begin{tabular}{ccccc}
    && \multicolumn{2}{c}{\textbf{Model Misspecification}} & \\
        & &
        \textbf{Prior} ($\M_P$) &
        \textbf{Simulator} ($\M_S$) \textbf{\& noise} ($\M_N$)
        \\
        \multirow{2}{*}{\hspace*{-0.1cm}\rotatebox[origin=c]{90}{\textbf{Summary Network}}} &
        \rotatebox[origin=c]{90}{\textbf{minimal}} &
        \raisebox{-0.48\height}{\includegraphics[width=0.40\linewidth]{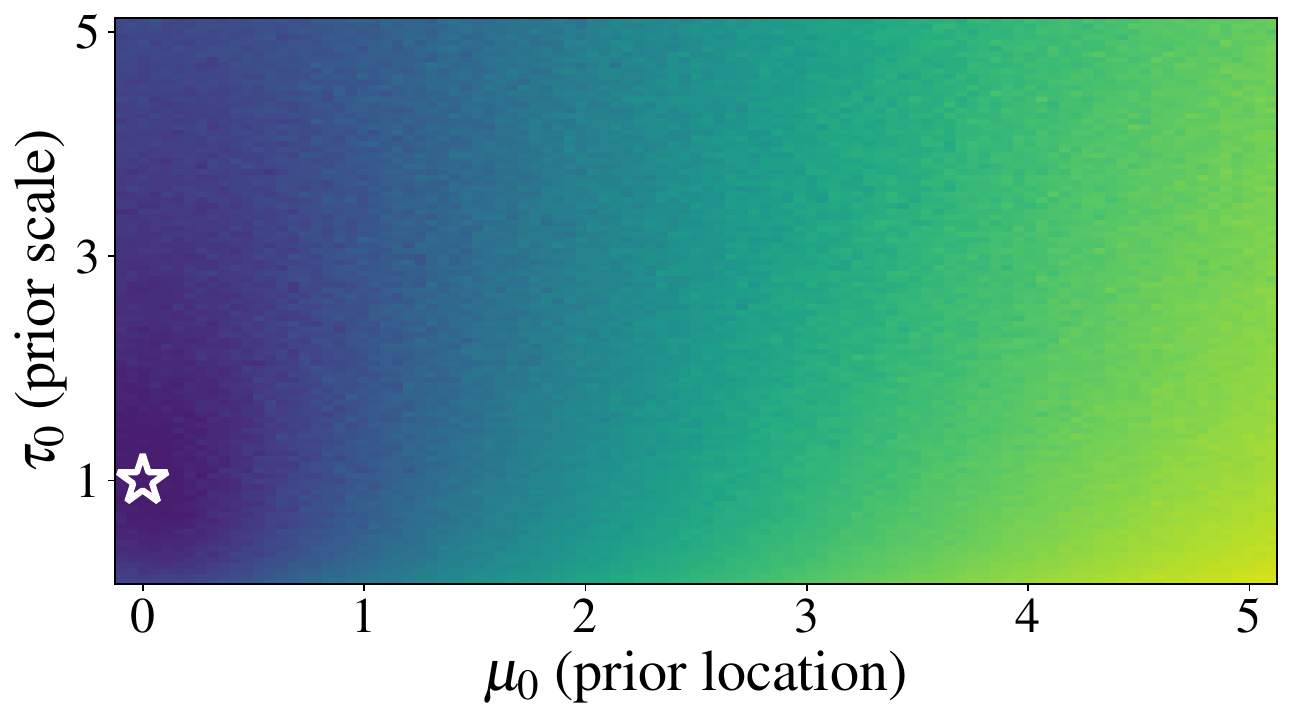}} &
        \raisebox{-0.48\height}{\includegraphics[width=0.40\linewidth]{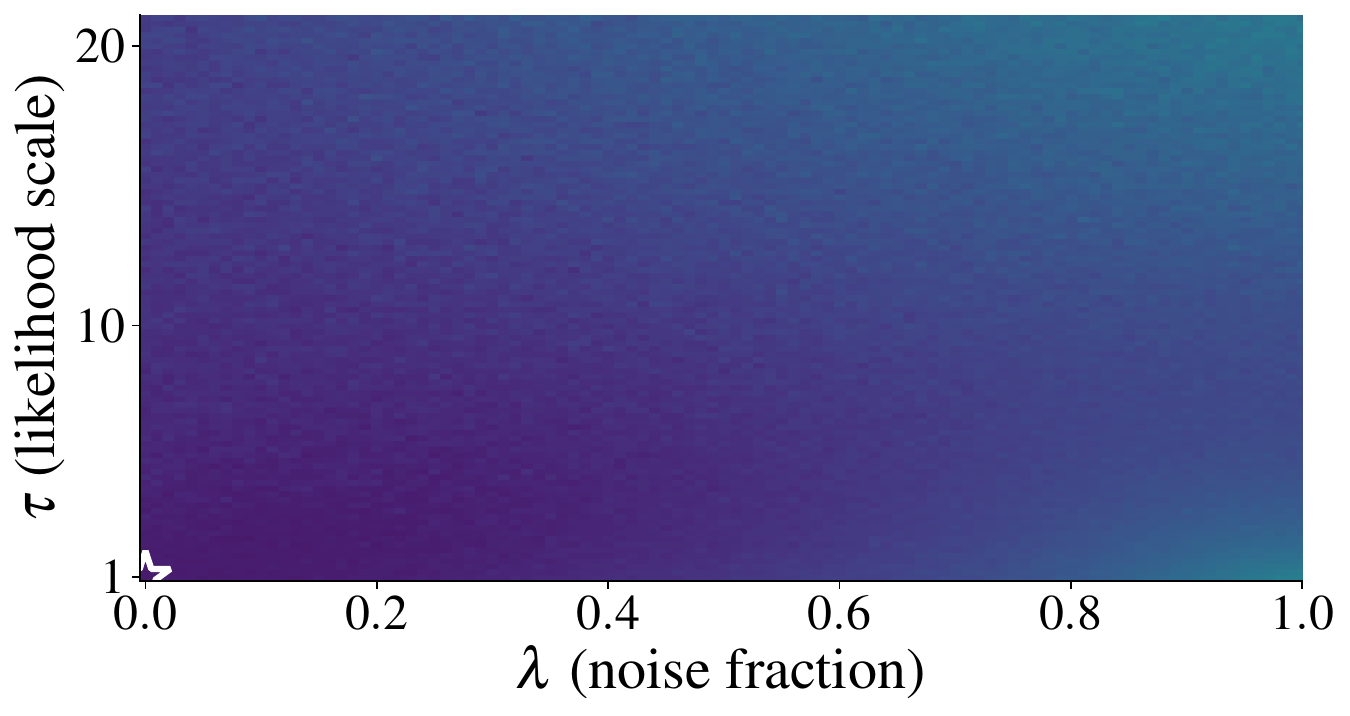}}
        \\
        &\rotatebox[origin=c]{90}{\textbf{overcomplete}} &
        \raisebox{-0.5\height}{\includegraphics[width=0.40\linewidth]{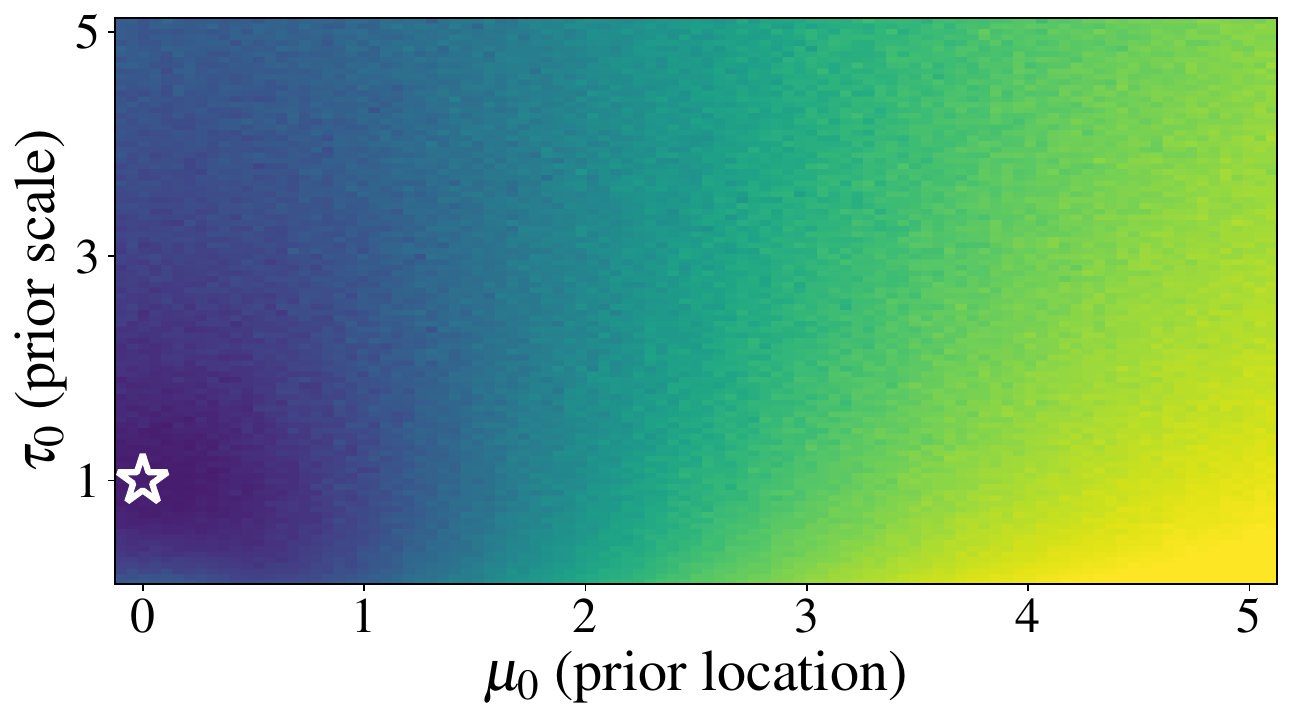}} &
        \raisebox{-0.5\height}{\includegraphics[width=0.40\linewidth]{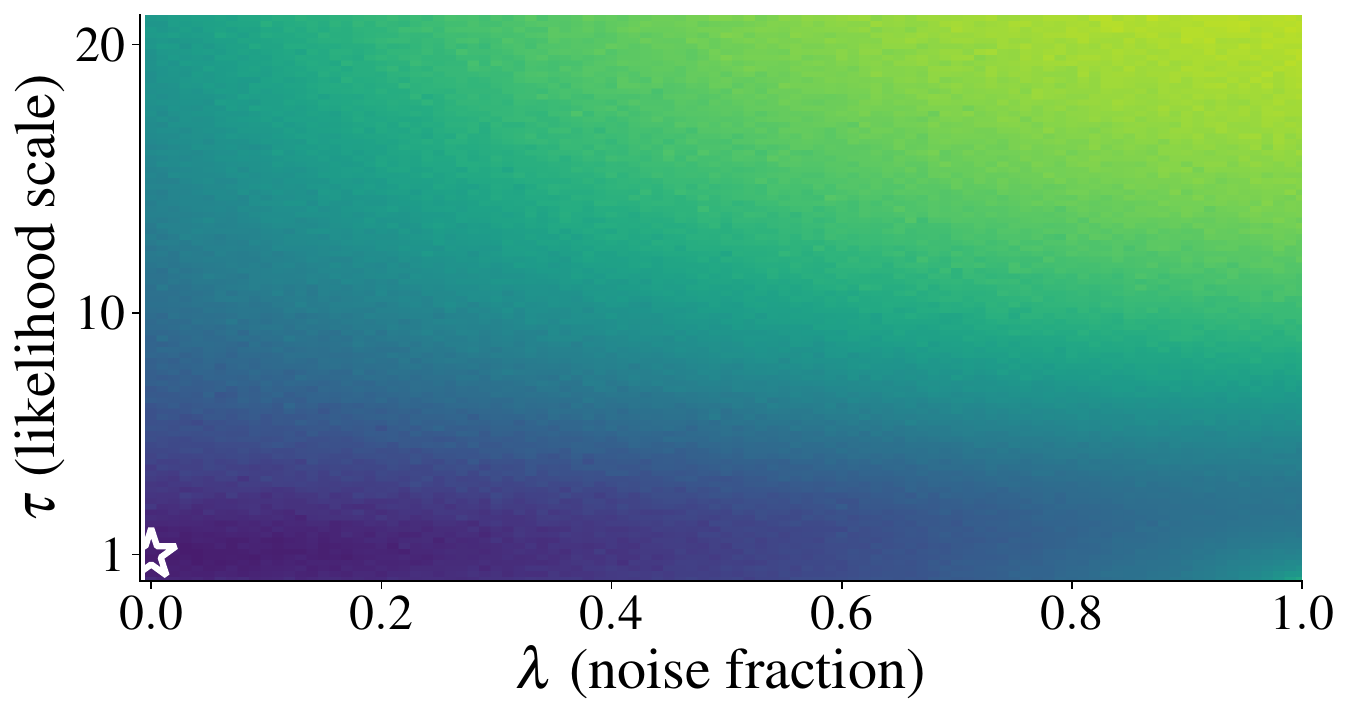}}
    \end{tabular}%
    \end{subfigure}%
    \begin{subfigure}[c]{0.06\linewidth}
    \includegraphics[width=\linewidth, clip, trim=9.8cm 0cm 0.2cm 0cm]{abf_mvn_means_mmd_heatmaps_colorbar.pdf}%
    \end{subfigure}
    \caption{\textbf{Experiment \numberGaussianMeans, SNPE-C.} Summary space MMD as a function of misspecification severity. White stars indicate the well-specified model configuration (i.e., equal to the training model $\M$).}
\end{figure}

\section{Epidemiological COVID-19 Model: Power Analysis}

\begin{figure}[H]
    \centering
    \begin{tabular}{cccc}
    & {\Large$N=1$} & {\Large$N=2$} & {\Large$N=5$}\\
    \rotatebox[origin=c]{90}{\raisebox{0.1cm}{\colsquare{observedcolor}}\Large$\M_1$} & 
    \raisebox{-0.48\height}{\includegraphics[width=.28\linewidth]{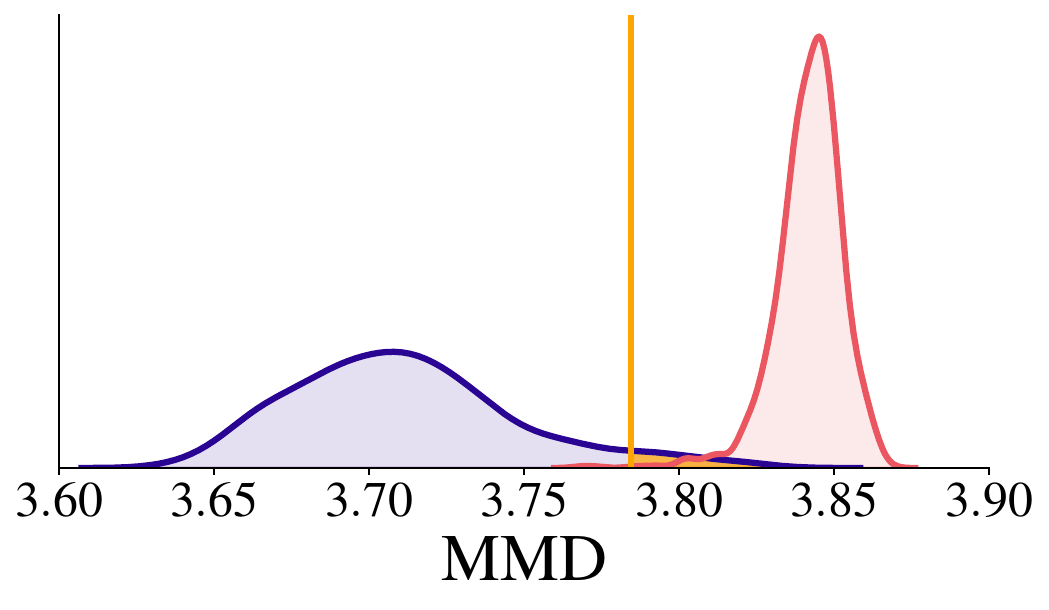}} &
    \raisebox{-0.48\height}{\includegraphics[width=.28\linewidth]{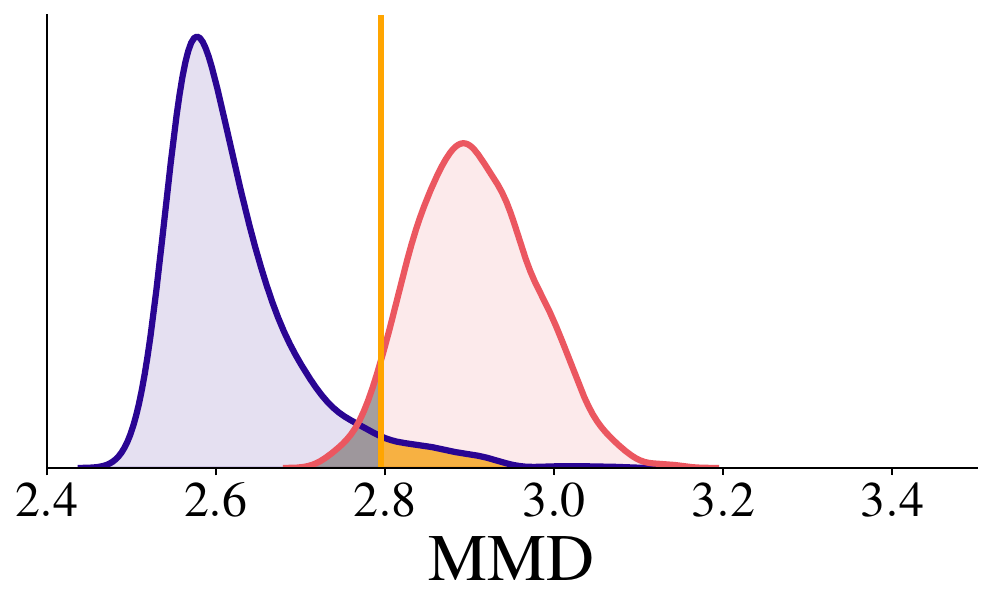}} &
    \raisebox{-0.48\height}{\includegraphics[width=.28\linewidth]{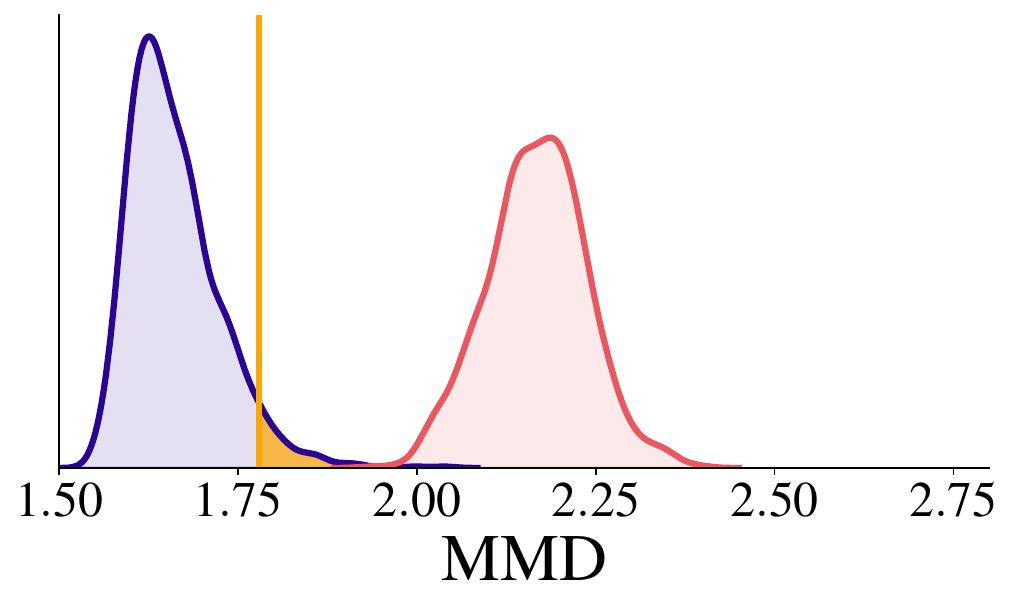}}\\
    \rotatebox[origin=c]{90}{\raisebox{0.1cm}{\colsquare{observedcolor}}\Large$\M_2$} & 
    \raisebox{-0.48\height}{\includegraphics[width=.28\linewidth]{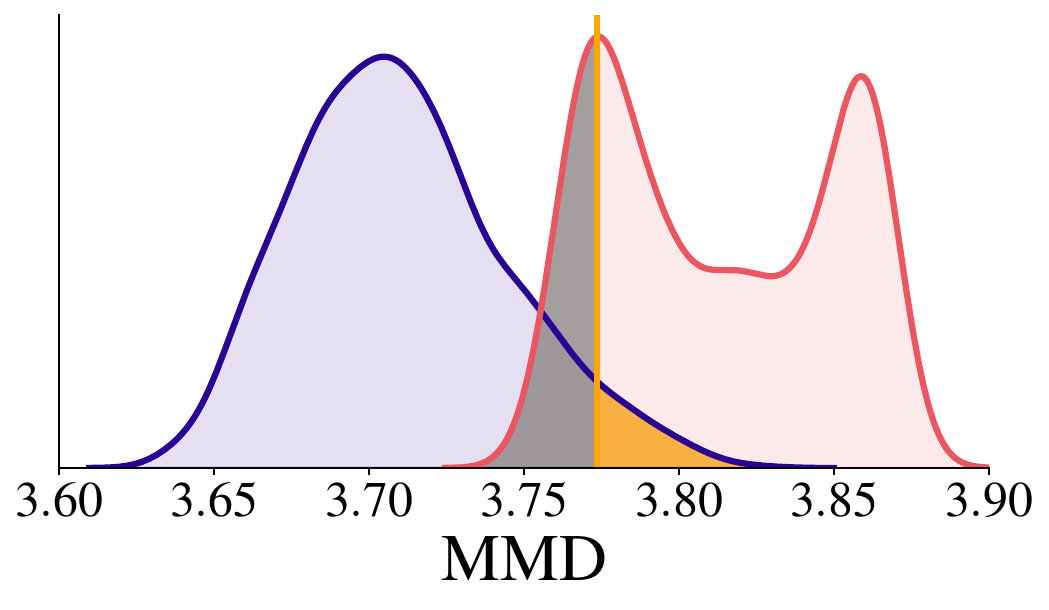}} &
    \raisebox{-0.48\height}{\includegraphics[width=.28\linewidth]{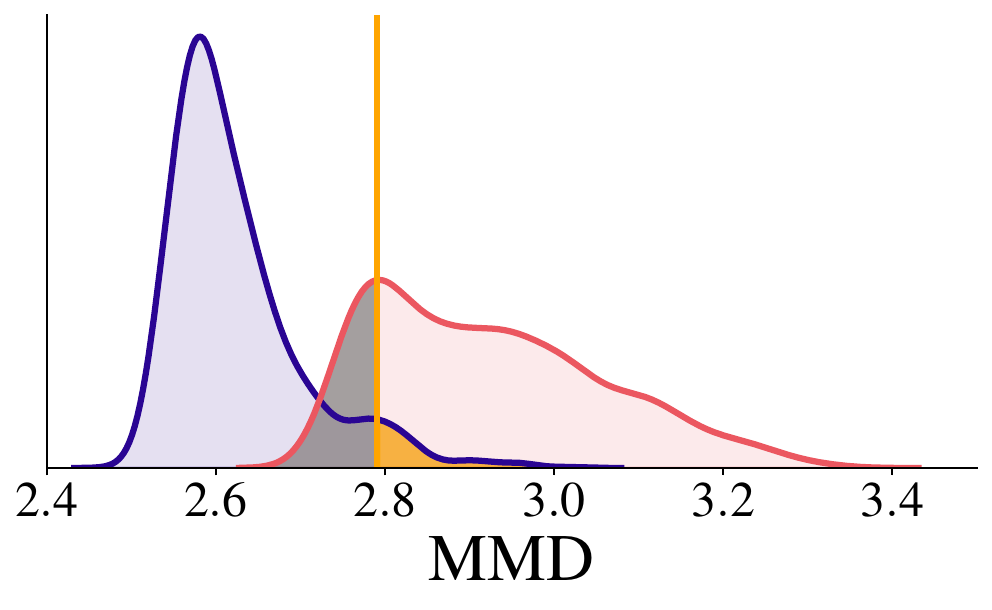}} &
    \raisebox{-0.48\height}{\includegraphics[width=.28\linewidth]{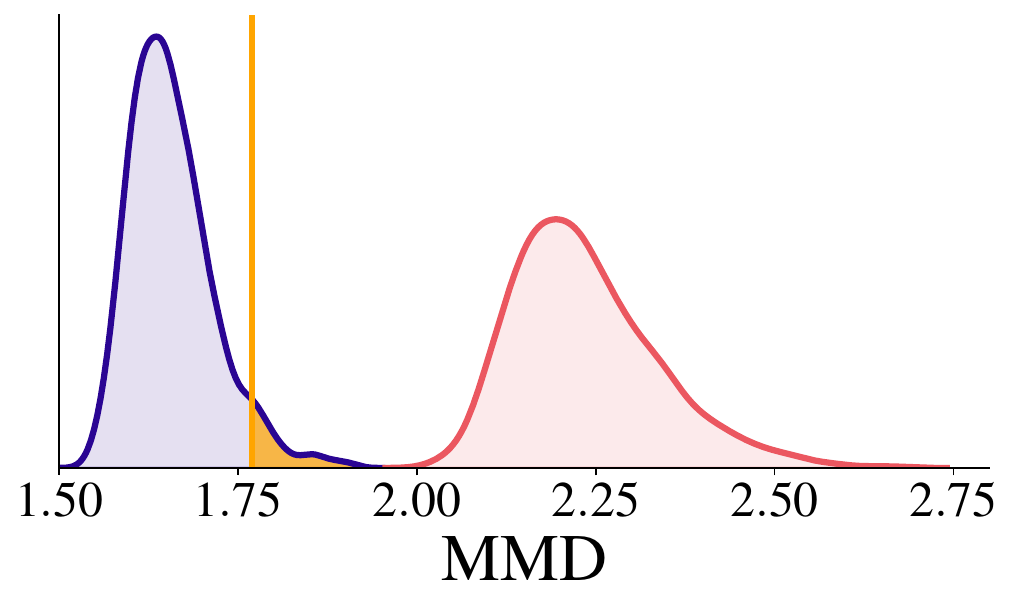}}\\
    \rotatebox[origin=c]{90}{\raisebox{0.1cm}{\colsquare{observedcolor}}\Large$\M_3$} & 
    \raisebox{-0.48\height}{\includegraphics[width=.28\linewidth]{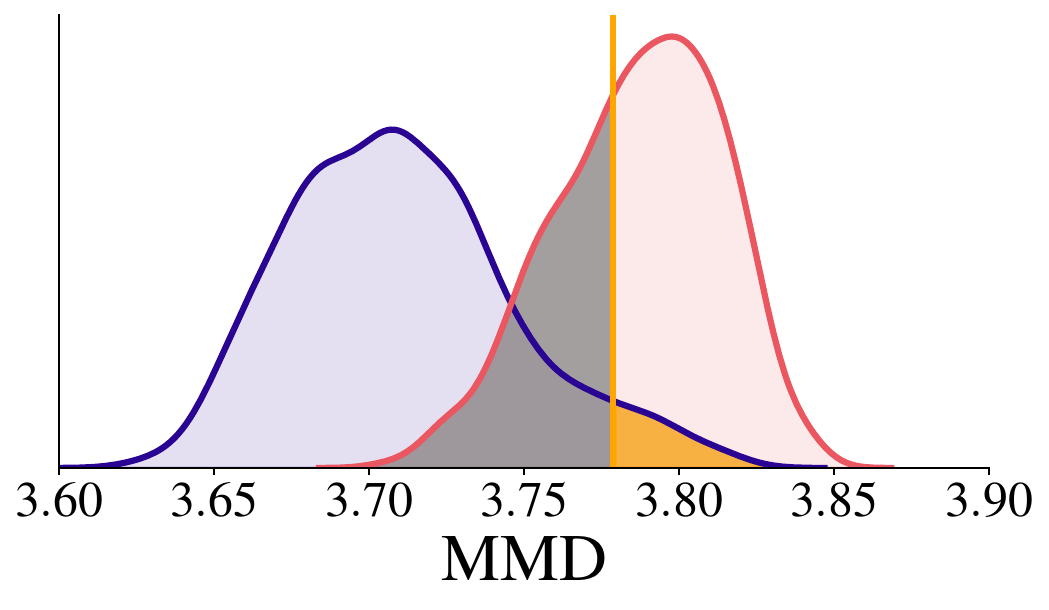}} &
    \raisebox{-0.48\height}{\includegraphics[width=.28\linewidth]{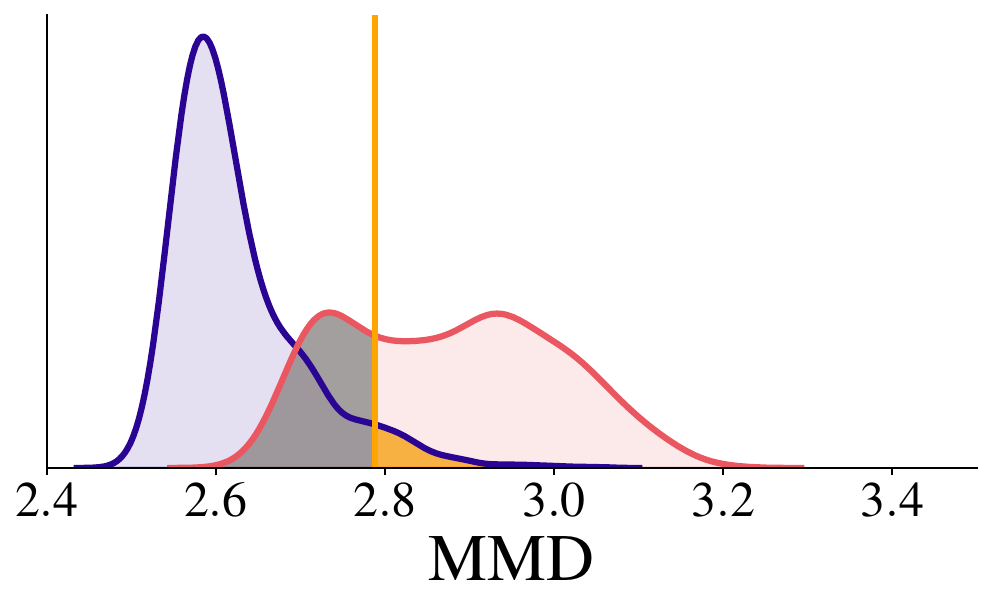}} &
    \raisebox{-0.48\height}{\includegraphics[width=.28\linewidth]{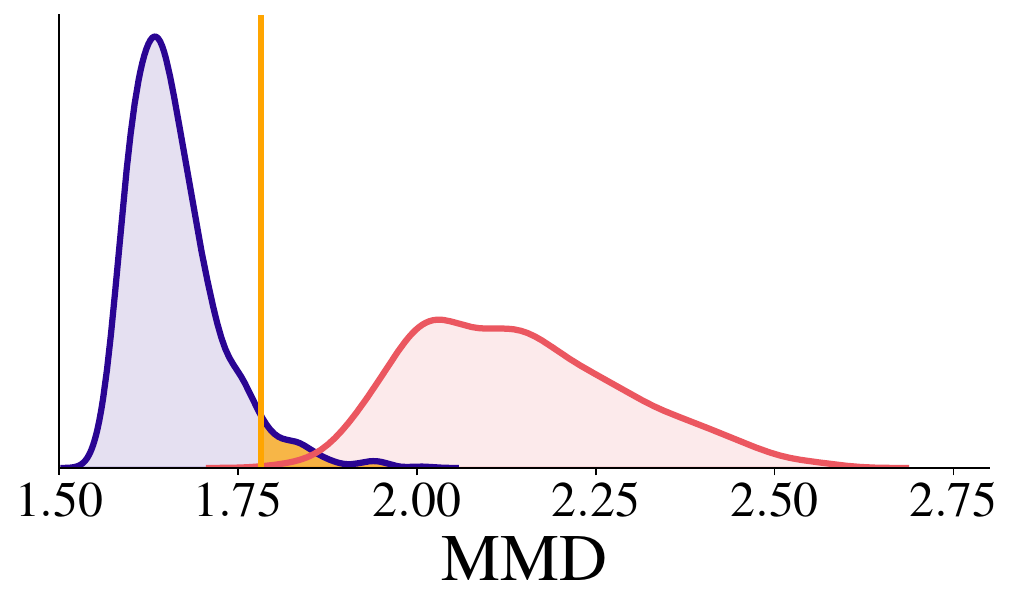}}\\
    \end{tabular}\\
    \includegraphics[width=1.0\linewidth]{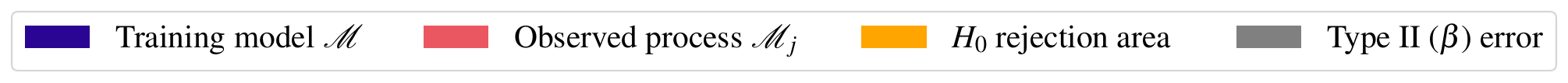}
    \caption{\textbf{Experiment \numberCovid.} Detailed illustration of the power analysis. As few as $N=5$ observed data sets suffice to achieve a negligible type II error $\beta\approx 0$.}
    \label{fig:app:covid:power}
\end{figure}
\vfill
\clearpage

\section{Style-Inference of EMNIST Image Simulator: Additional Results}\label{app:emnist-gin}

\begin{figure}[H]
    \centering
    \begin{subfigure}[b]{0.35\textwidth}
        \centering
        \includegraphics[width=\linewidth]{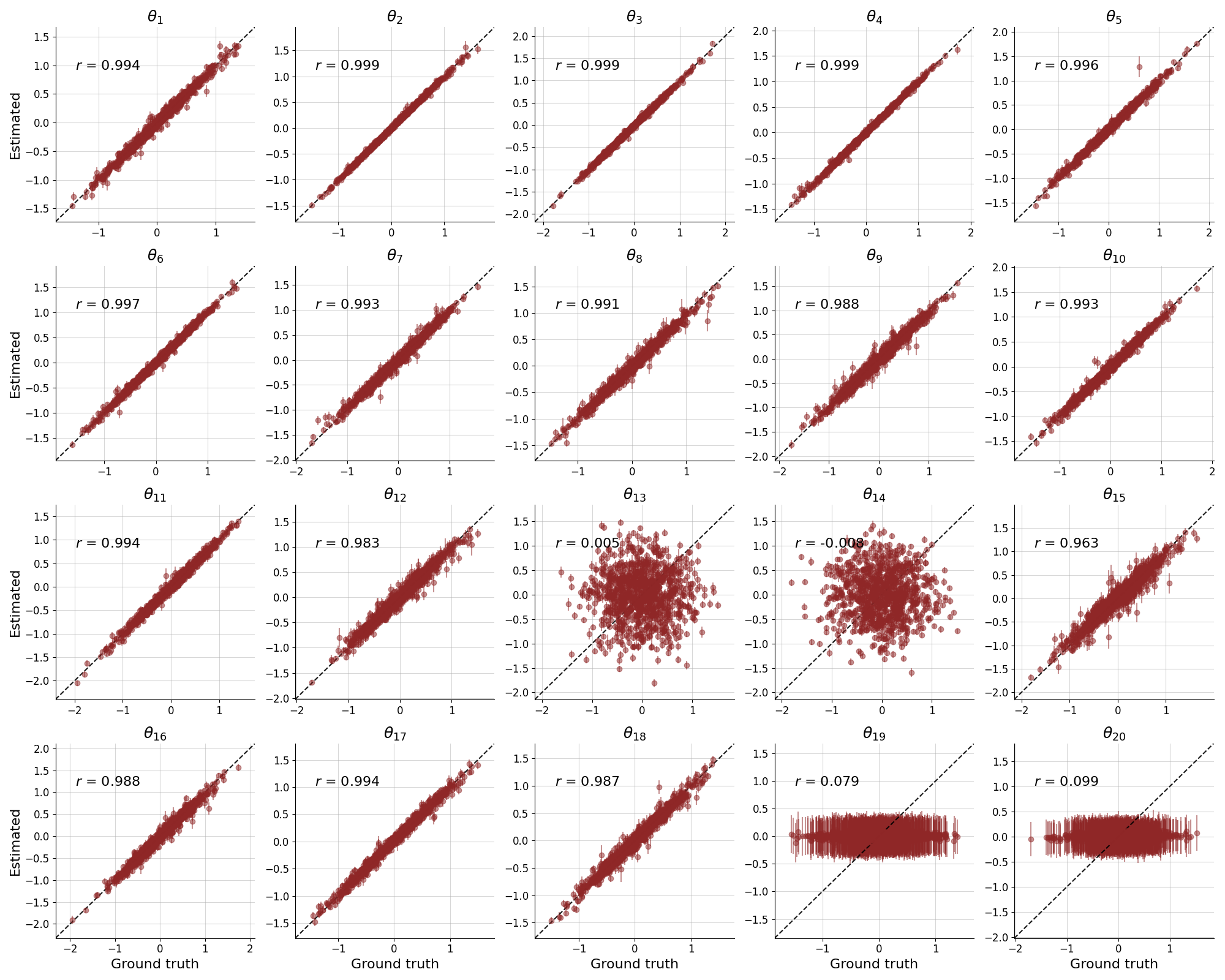}
        \caption{In distribution (digits 0--4)}
    \end{subfigure}
    \\
    \begin{subfigure}[b]{0.35\textwidth}
        \centering
        \includegraphics[width=\linewidth]{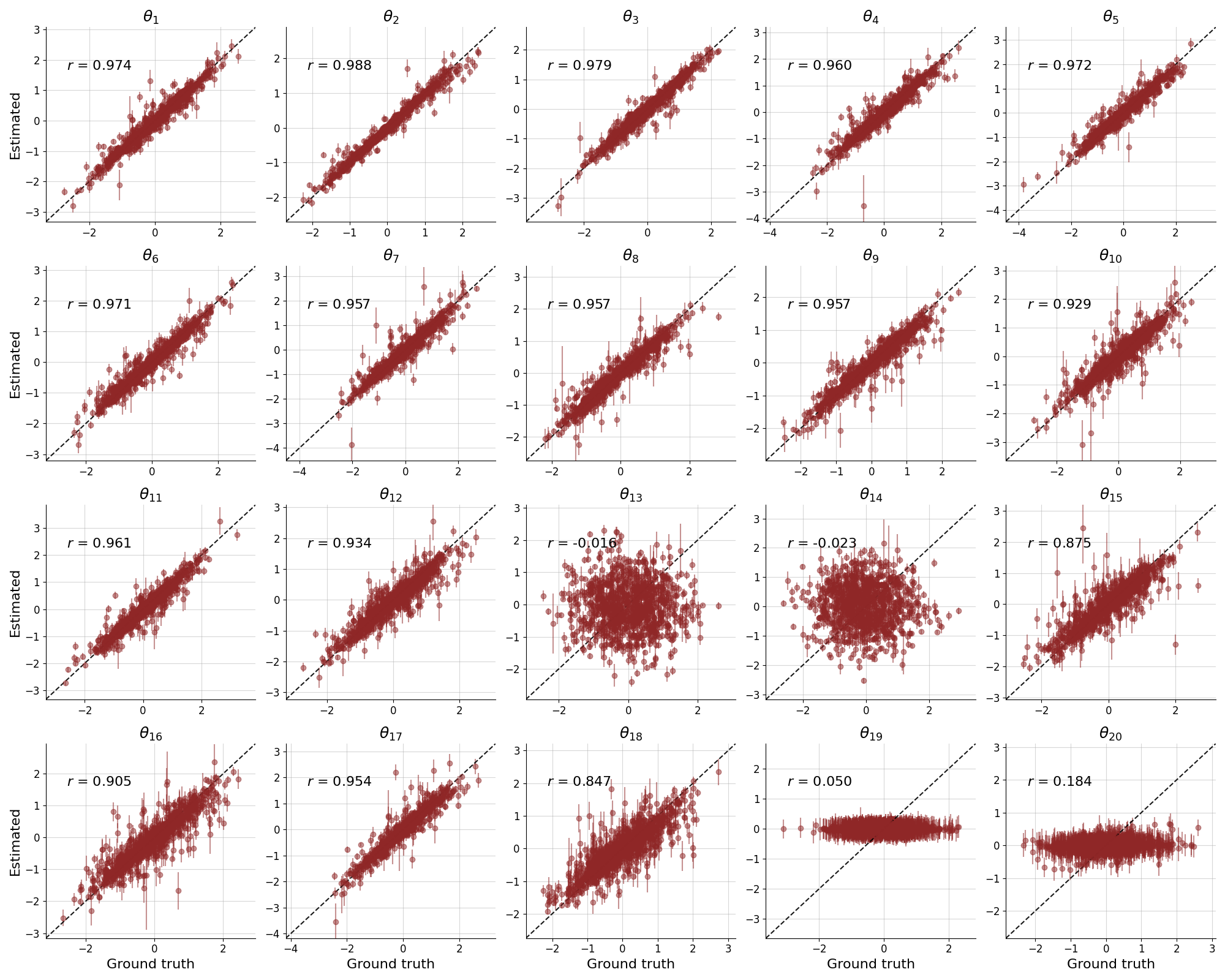}
        \caption{Extreme styles (digits 0--4)}
    \end{subfigure}
    \hspace*{1cm}
    \begin{subfigure}[b]{0.35\textwidth}
        \centering
        \includegraphics[width=\linewidth]{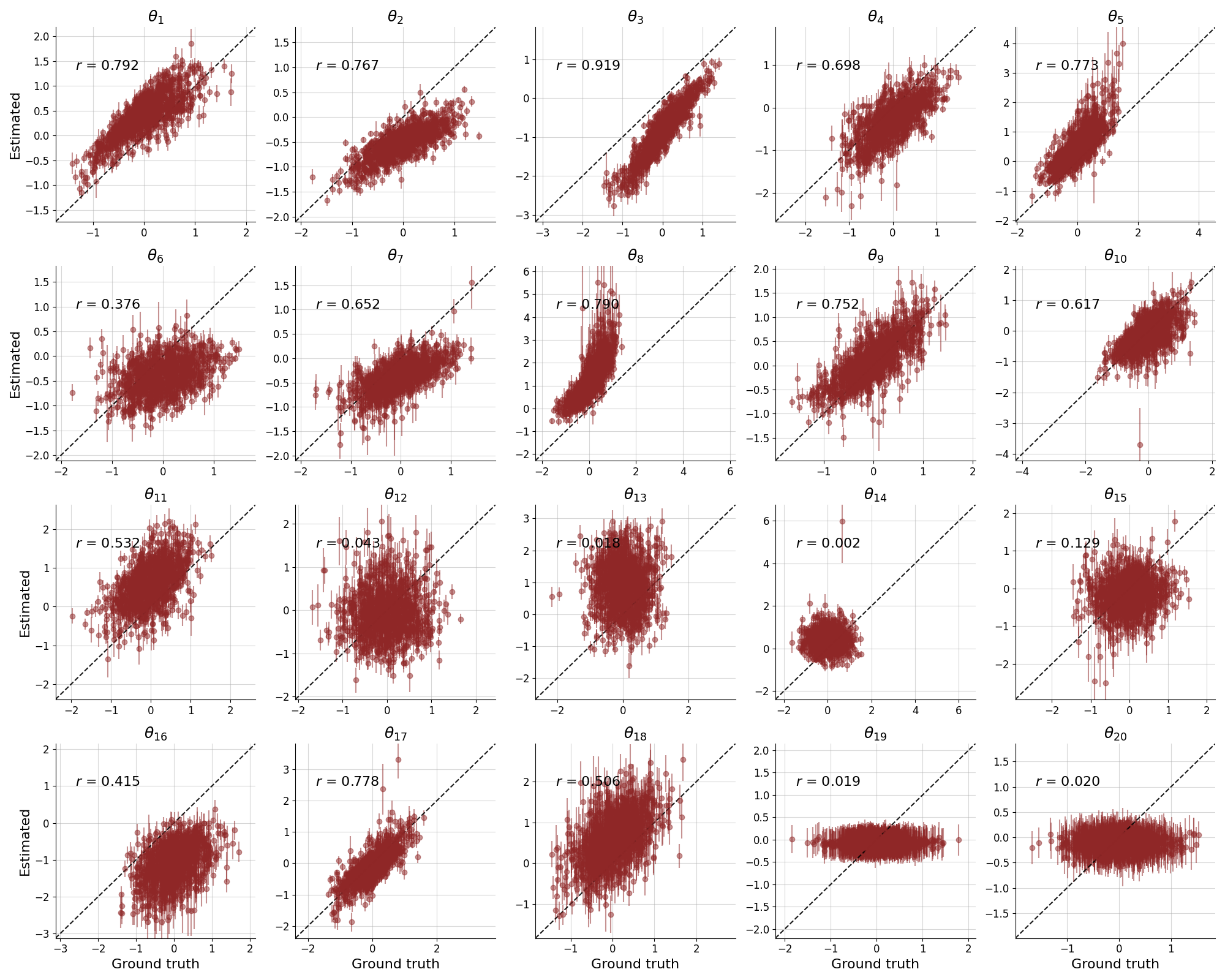}
        \caption{Domain shift (digit 5)}
    \end{subfigure}
    \\
    \begin{subfigure}[b]{0.35\textwidth}
        \centering
        \includegraphics[width=\linewidth]{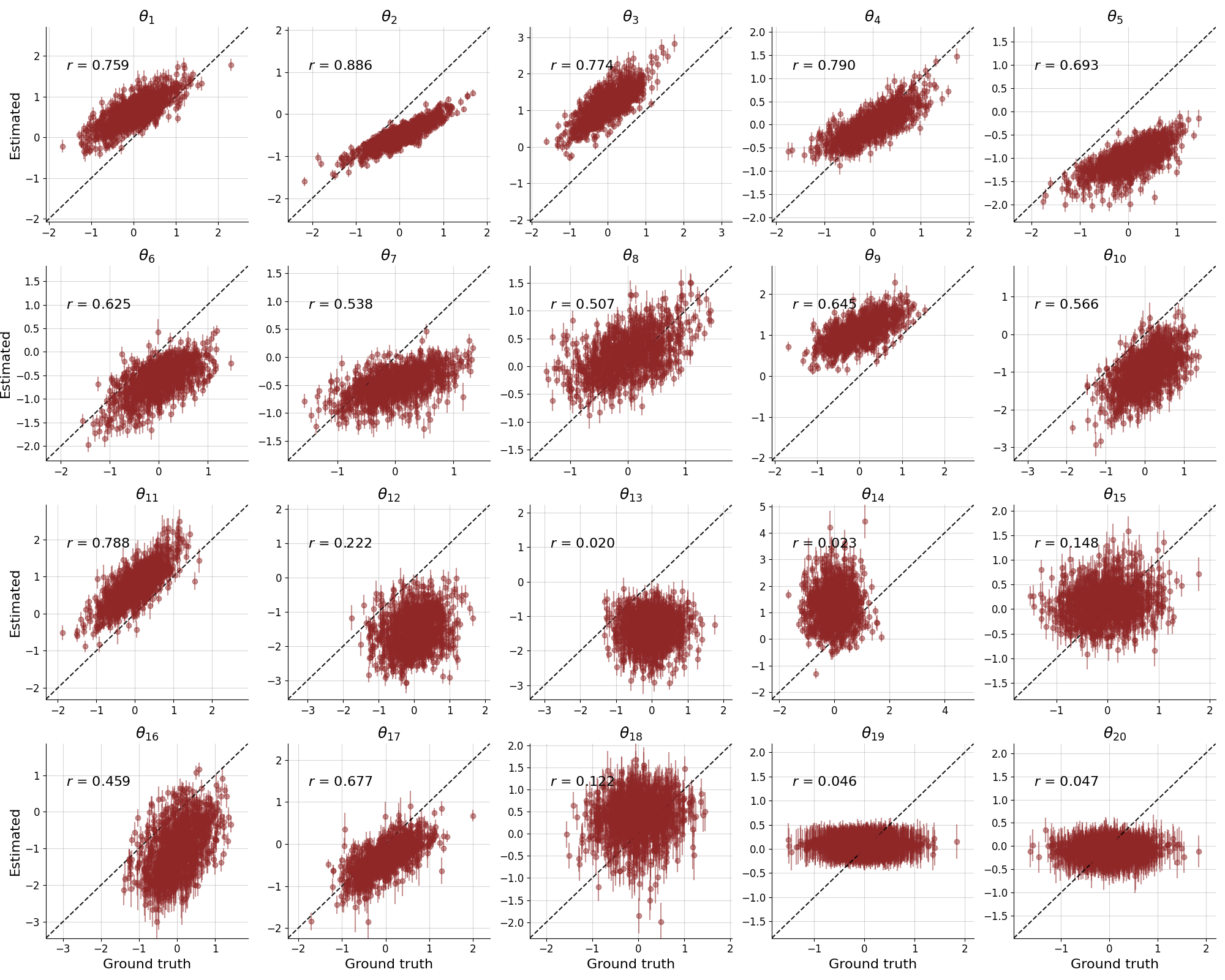}
        \caption{Domain shift (digit 6)}
    \end{subfigure}
    \hspace*{1cm}
    \begin{subfigure}[b]{0.35\textwidth}
        \centering
        \includegraphics[width=\linewidth]{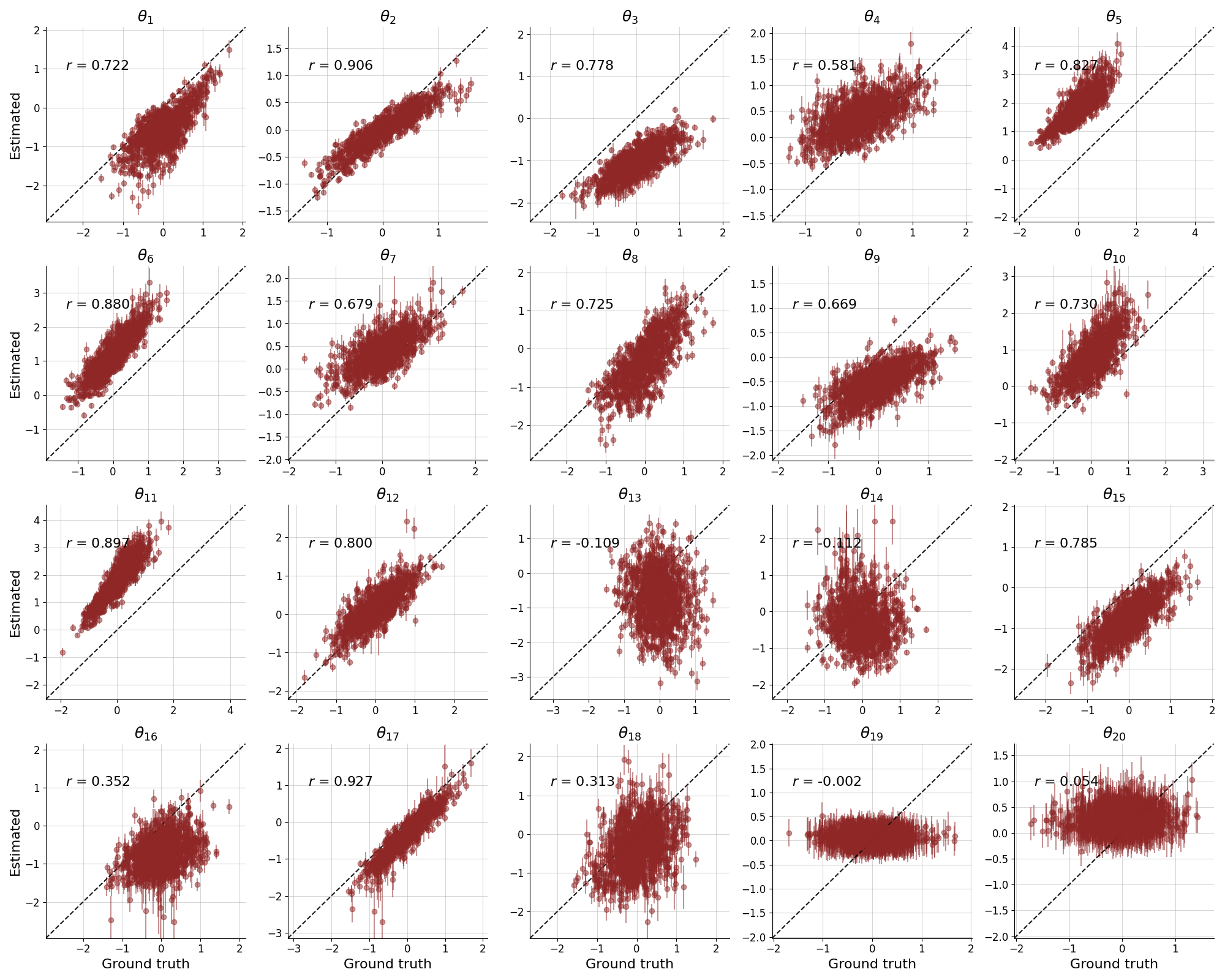}
        \caption{Domain shift (digit 7)}
    \end{subfigure}
    \\
    \begin{subfigure}[b]{0.35\textwidth}
        \centering
        \includegraphics[width=\linewidth]{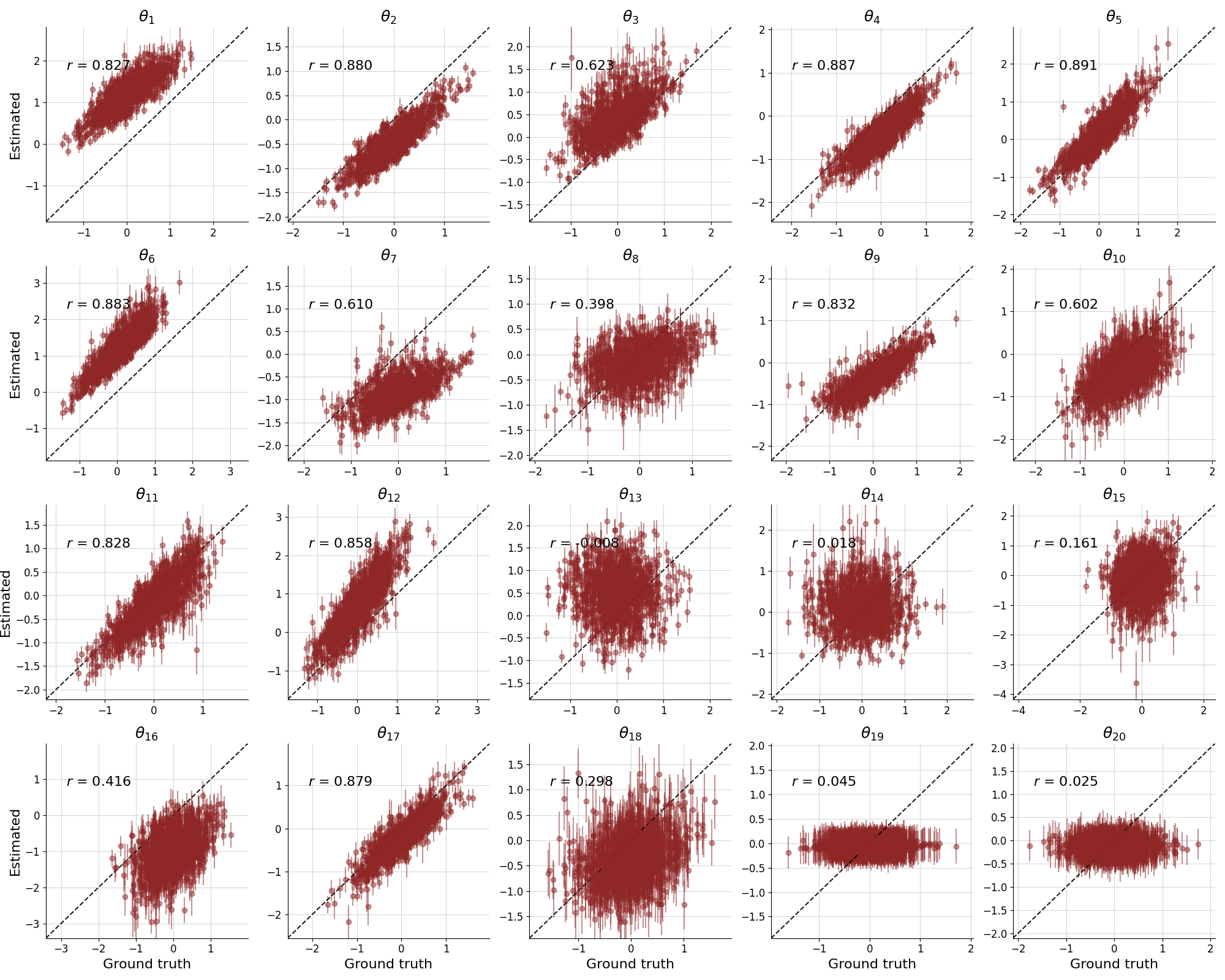}
        \caption{Domain shift (digit 8)}
    \end{subfigure}
    \hspace*{1cm}
    \begin{subfigure}[b]{0.35\textwidth}
        \centering
        \includegraphics[width=\linewidth]{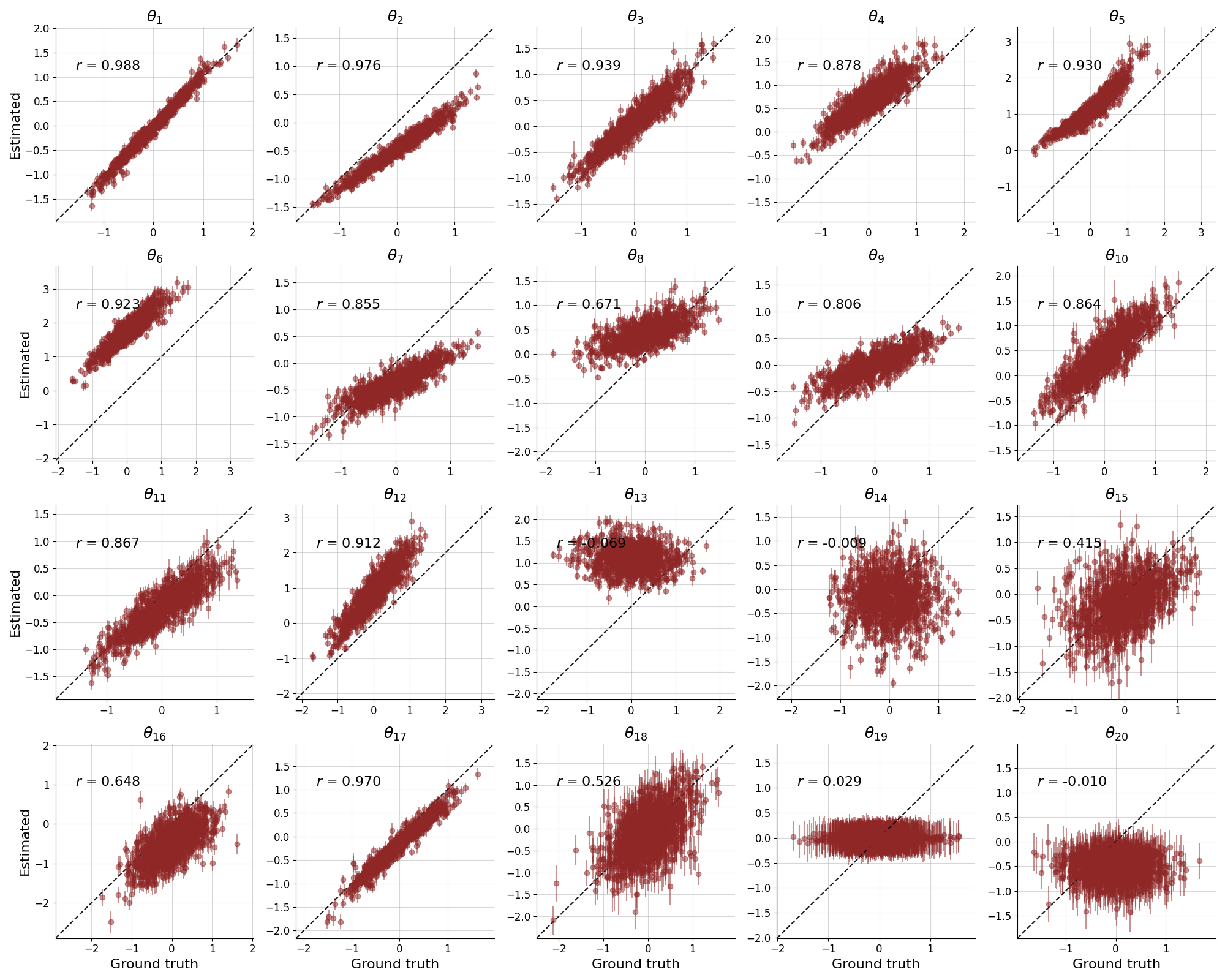}
        \caption{Domain shift (digit 9)}
    \end{subfigure}
    \caption{\textbf{Experiment \numberGIN}, variation 1 (class-agnostic), parameter recovery.}
\end{figure}

\begin{figure}[H]
    \centering
    \begin{subfigure}[b]{0.35\textwidth}
        \centering
        \includegraphics[width=\linewidth]{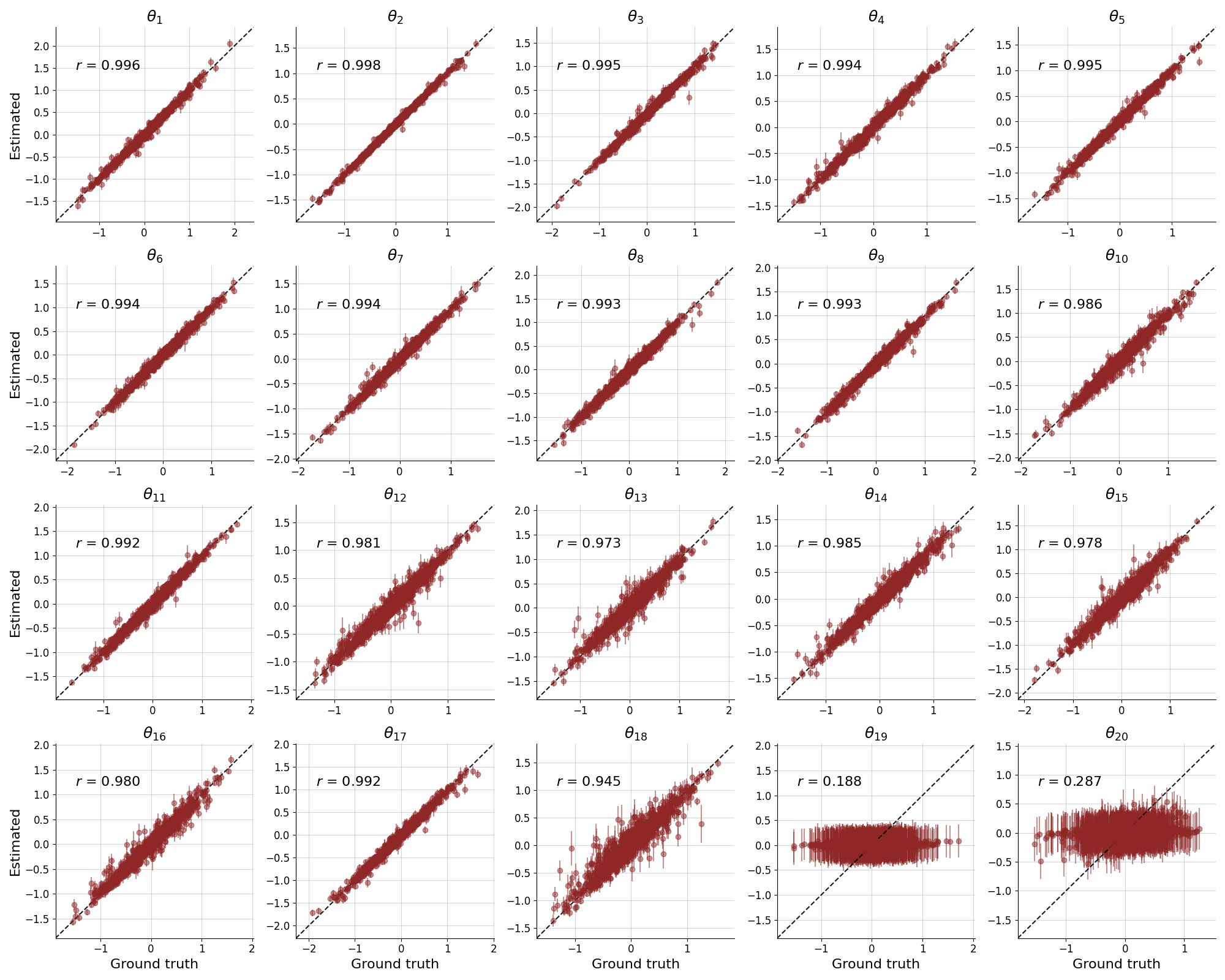}
        \caption{In distribution (digits 0--4)}
    \end{subfigure}
    \\
    \begin{subfigure}[b]{0.35\textwidth}
        \centering
        \includegraphics[width=\linewidth]{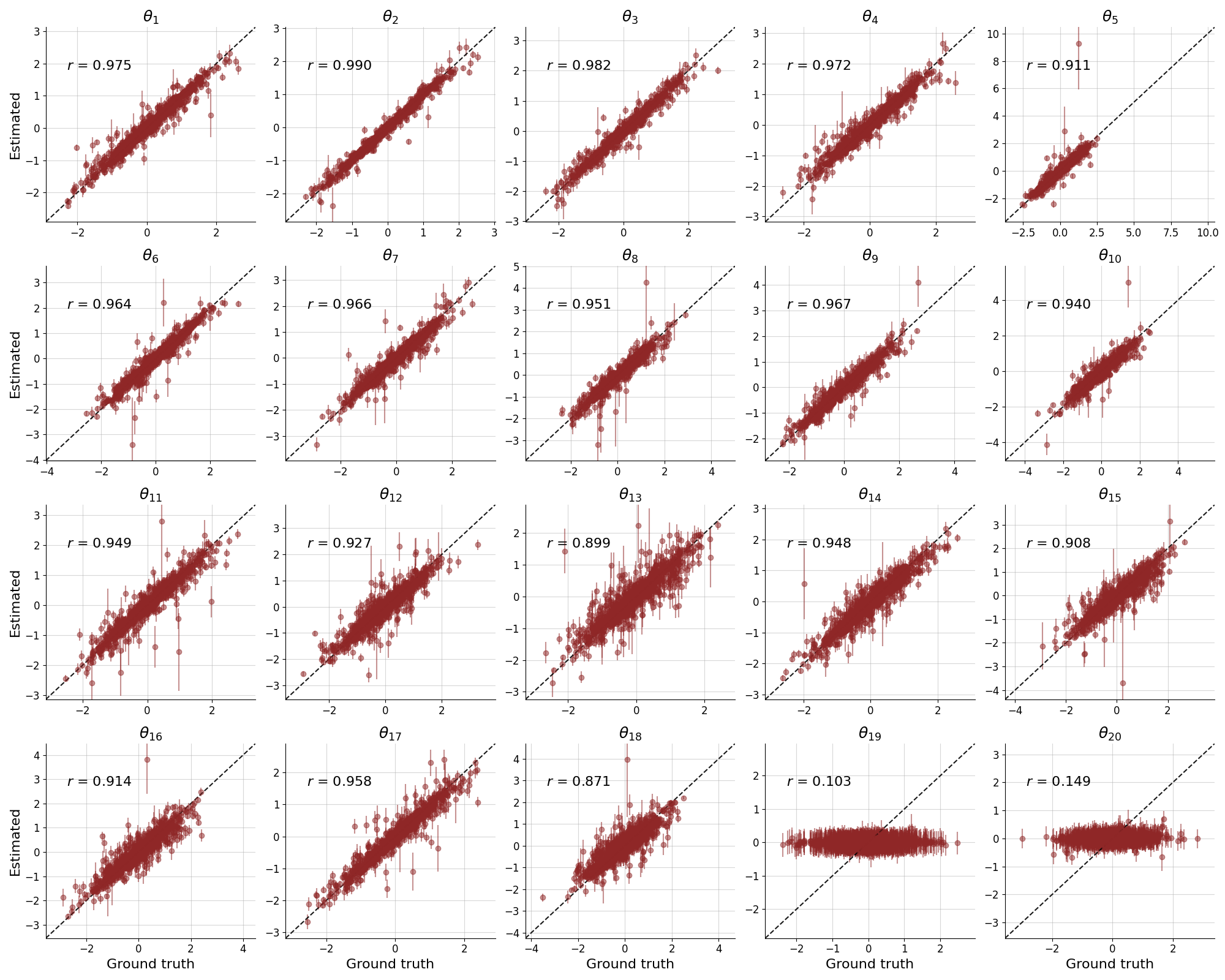}
        \caption{Extreme styles (digits 0--4)}
    \end{subfigure}
    \hspace*{1cm}
    \begin{subfigure}[b]{0.35\textwidth}
        \centering
        \includegraphics[width=\linewidth]{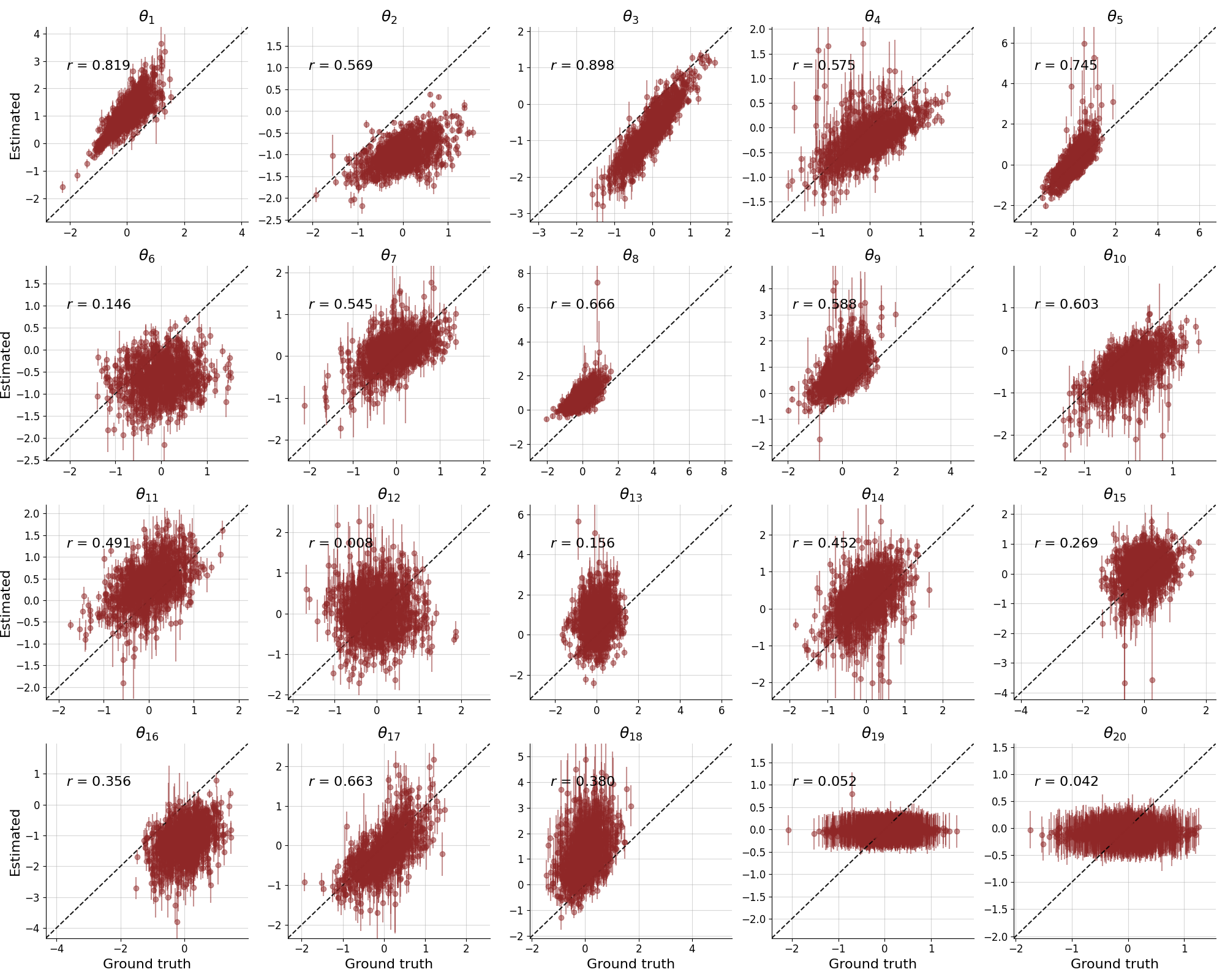}
        \caption{Domain shift (digit 5)}
    \end{subfigure}
    \\
    \begin{subfigure}[b]{0.35\textwidth}
        \centering
        \includegraphics[width=\linewidth]{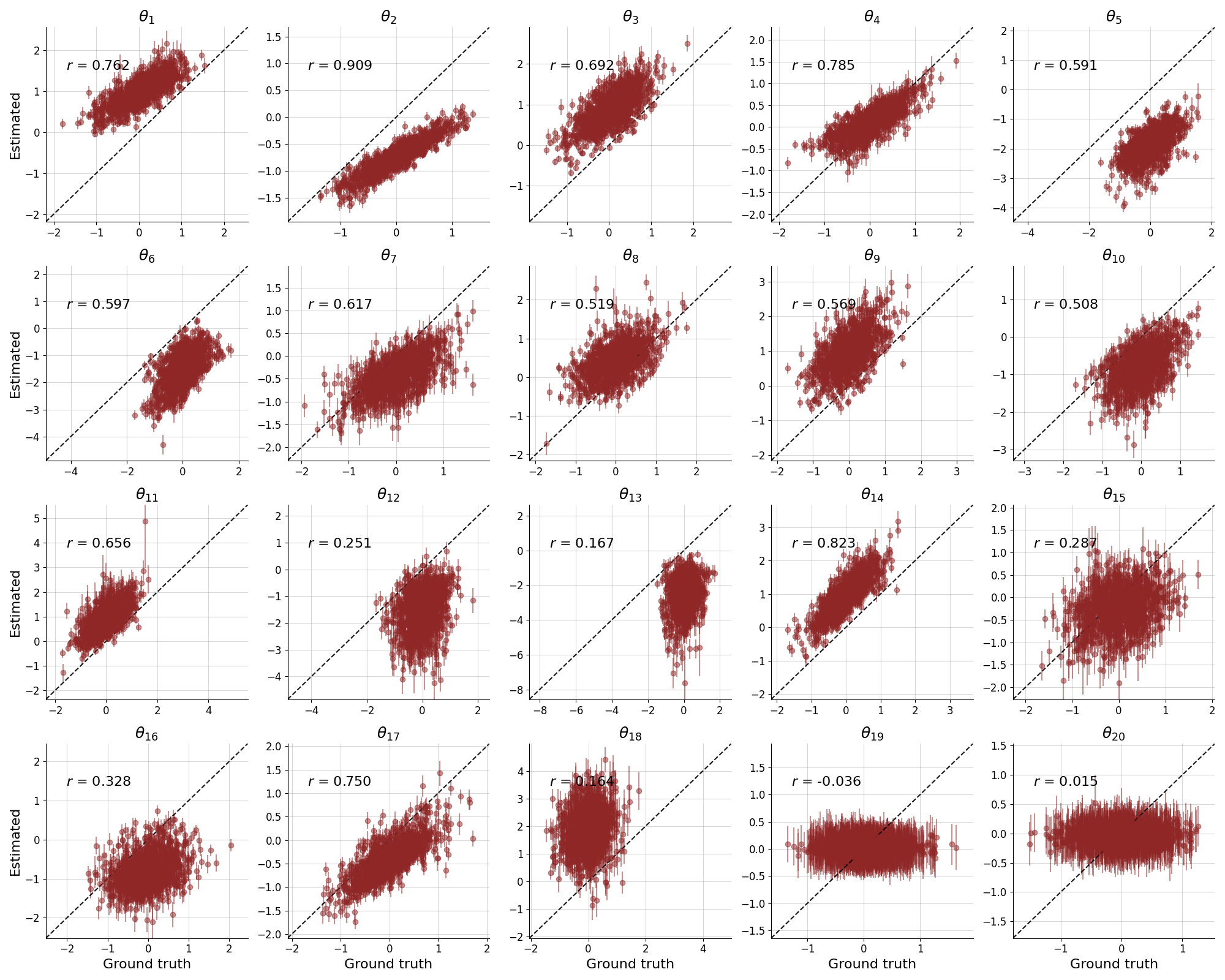}
        \caption{Domain shift (digit 6)}
    \end{subfigure}
    \hspace*{1cm}
    \begin{subfigure}[b]{0.35\textwidth}
        \centering
        \includegraphics[width=\linewidth]{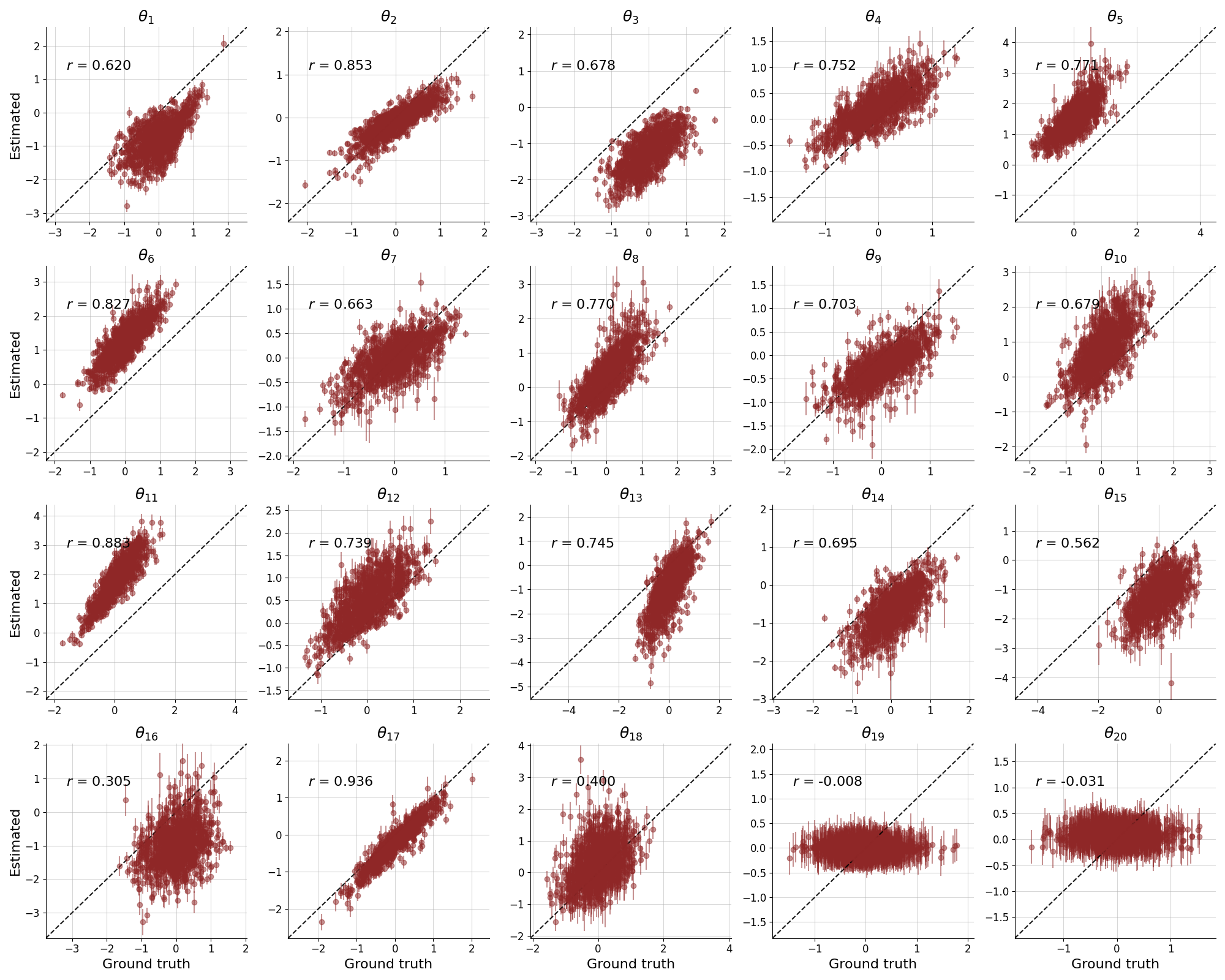}
        \caption{Domain shift (digit 7)}
    \end{subfigure}
    \\
    \begin{subfigure}[b]{0.35\textwidth}
        \centering
        \includegraphics[width=\linewidth]{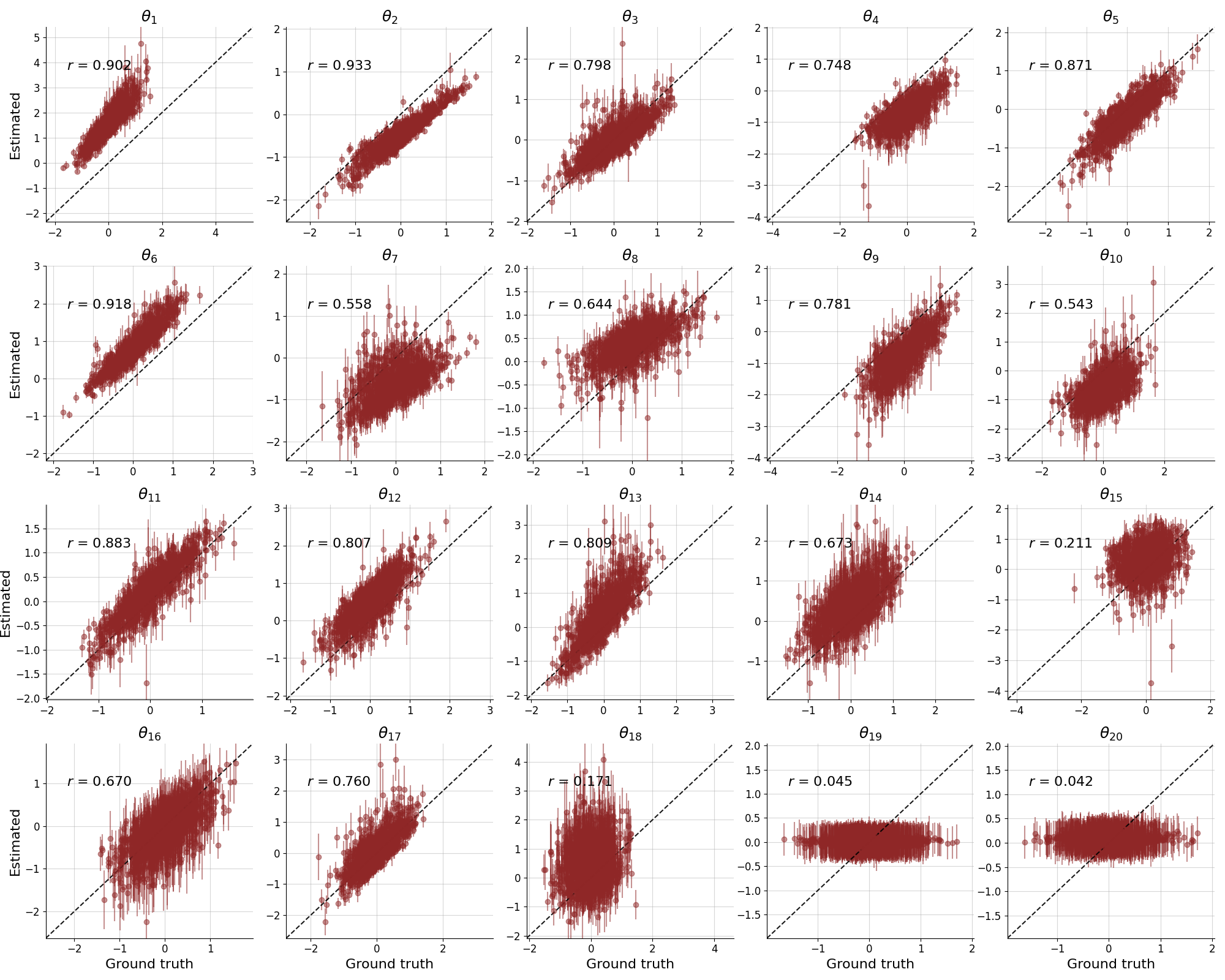}
        \caption{Domain shift (digit 8)}
    \end{subfigure}
    \hspace*{1cm}
    \begin{subfigure}[b]{0.35\textwidth}
        \centering
        \includegraphics[width=\linewidth]{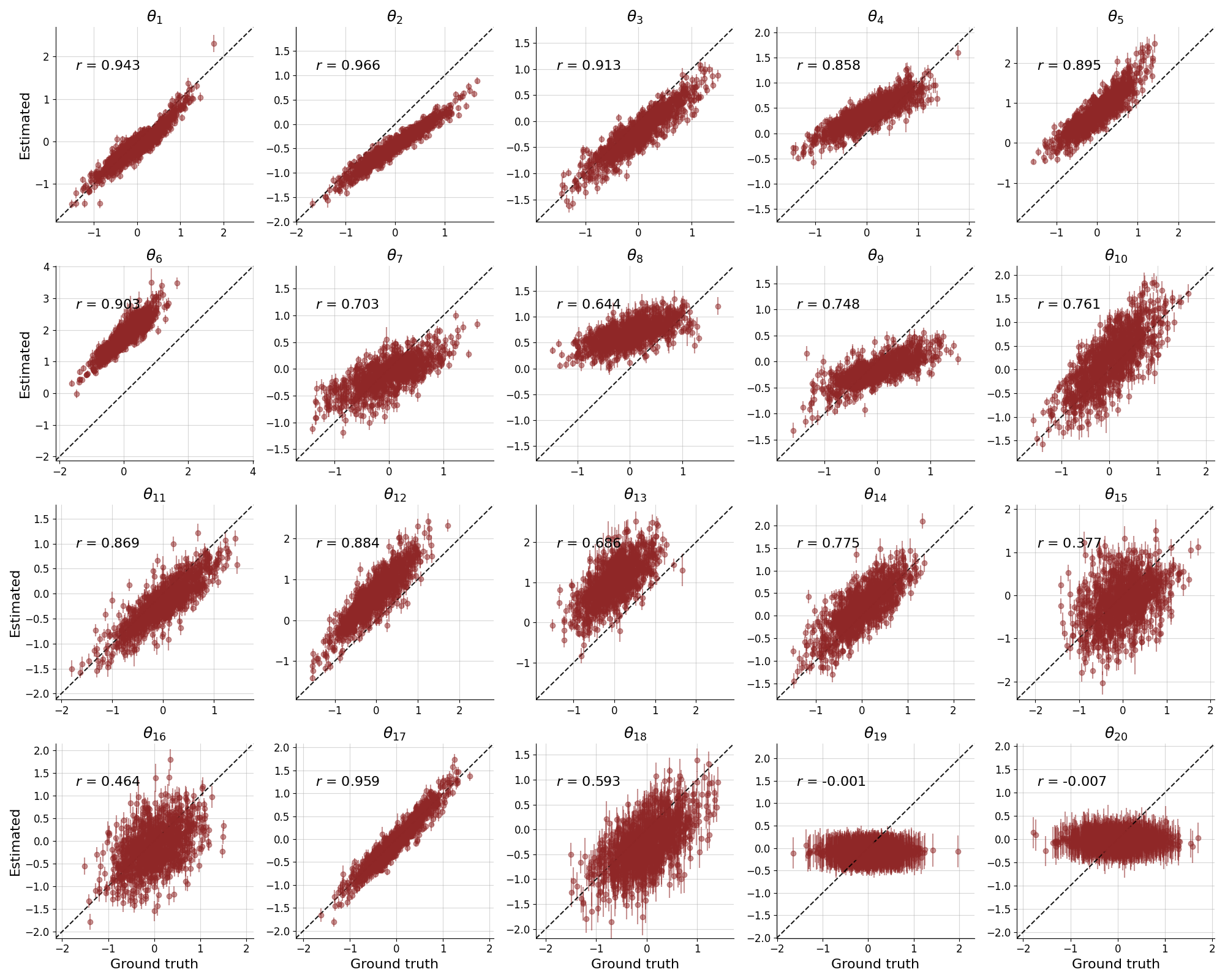}
        \caption{Domain shift (digit 9)}
    \end{subfigure}
    \caption{\textbf{Experiment \numberGIN}, variation 2 (class-informed), parameter recovery.}
\end{figure}

\end{appendices}

\end{document}